\documentclass[Afour,sageh,times]{sagej}

\usepackage{moreverb,url}

\usepackage[colorlinks,bookmarksopen,bookmarksnumbered,citecolor=red,urlcolor=red]{hyperref}

\newcommand\BibTeX{{\rmfamily B\kern-.05em \textsc{i\kern-.025em b}\kern-.08em
T\kern-.1667em\lower.7ex\hbox{E}\kern-.125emX}}

\usepackage{times}
\usepackage{soul}
\usepackage{url}
\usepackage{graphicx}
\usepackage{amsmath,amssymb}
\usepackage{amsthm, nccmath}
\usepackage{booktabs}
\usepackage{algorithm}
\usepackage{algpseudocode}
\usepackage[subrefformat=parens]{subcaption}
\usepackage[dvipsnames]{xcolor}
\usepackage{float}

\usepackage{longtable}
\usepackage{multirow}
\usepackage{makecell}

\definecolor{mydarkgreen}{RGB}{26, 184, 15}

\usepackage{bbold}
\usepackage[export]{adjustbox}
\usepackage{lscape}

\newcommand{\argmax}[2]{\mathrm{arg}\;\underset{#1}{\max}\;#2}
\newcommand{\argmin}[2]{\mathrm{arg}\;\underset{#1}{\min}\;#2}
\newcommand{\norm}[1]{\left\lVert#1\right\rVert}
\newcommand{\Mycomment}[1]{}

\makeatletter
\AfterEndEnvironment{algorithm}{\let\@algcomment\relax}
\AtEndEnvironment{algorithm}{\kern2pt\hrule\relax\vskip3pt\@algcomment}
\let\@algcomment\relax
\newcommand\algcomment[1]{\def\@algcomment{\footnotesize#1}}

\renewcommand\fs@ruled{\def\@fs@cfont{\bfseries}\let\@fs@capt\floatc@ruled
  \def\@fs@pre{\hrule height.8pt depth0pt \kern2pt}%
  \def\@fs@post{}%
  \def\@fs@mid{\kern2pt\hrule\kern2pt}%
  \let\@fs@iftopcapt\iftrue}
\makeatother

\makeatletter
\let\oldlt\longtable
\let\endoldlt\endlongtable
\def\longtable{\@ifnextchar[\longtable@i \longtable@ii}
\def\longtable@i[#1]{\begin{figure}[t]
\onecolumn
\begin{minipage}{0.5\textwidth}
\oldlt[#1]
}
\def\longtable@ii{\begin{figure}[t]
\onecolumn
\begin{minipage}{0.5\textwidth}
\oldlt
}
\def\endlongtable{\endoldlt
\end{minipage}
\twocolumn
\end{figure}}
\makeatother

\theoremstyle{definition}
\newtheorem{definition}{Definition}

\def\volumeyear{2016}

\begin{document}

\runninghead{Ilboudo et al.}

\title{Domains as Objectives: Domain-Uncertainty-Aware Policy Optimization through Explicit Multi-Domain Convex Coverage Set Learning}

\author{Wendyam Eric Lionel Ilboudo\affilnum{1}, Taisuke Kobayashi\affilnum{2} and Takamitsu Matsubara\affilnum{1}}

\affiliation{
\affilnum{1}Nara Institute of Science and Technology (NAIST), Japan\\
\affilnum{2}National Institute of Informatics (NII) / The Graduate University for Advanced Studies (SOKENDAI), Japan
}

\corrauth{Wendyam Eric Lionel Ilboudo, Nara Institute of Science and Technology
Robot Learning Laboratory,
Information Science Division,
Ikoma, Nara,
Japan.}

\email{ilboudo.wendyam\_eric.in1@naist.ac.jp}

\begin{abstract}
The problem of uncertainty is a feature of real world robotics problems and any control framework must contend with it in order to succeed in real applications tasks. Reinforcement Learning is no different, and epistemic uncertainty arising from model uncertainty or misspecification is a challenge well captured by the sim-to-real gap. A simple solution to this issue is domain randomization (DR), which unfortunately can result in conservative agents. As a remedy to this conservativeness, the use of universal policies that take additional information about the randomized domain has risen as an alternative solution, along with recurrent neural network-based controllers. Uncertainty-aware universal policies present a particularly compelling solution able to account for system identification uncertainties during deployment. 
In this paper, we reveal that the challenge of efficiently optimizing uncertainty-aware policies can be fundamentally reframed as solving the convex coverage set (CCS) problem within a multi-objective reinforcement learning (MORL) context. By introducing a novel Markov decision process (MDP) framework where each domain's performance is treated as an independent objective, we unify the training of uncertainty-aware policies with MORL approaches. This connection enables the application of MORL algorithms for domain randomization (DR), allowing for more efficient policy optimization. To illustrate this, we focus on the linear utility function, which aligns with the expectation in DR formulations, and propose a series of algorithms adapted from the MORL literature to solve the CCS, demonstrating their ability to enhance the performance of uncertainty-aware policies.
\end{abstract}


\maketitle

\section{Introduction}

 The very nature of the Reinforcement Learning (RL) framework with its trial-and-error approach motivates the need for policies trained in simulations that can perform optimally on target real robotics systems. Indeed, the exploration required during the training of RL agents combined with the sample inefficiency of RL algorithms often entails a multitude of iterations in order to fine-tune and optimize the policies. However, subjecting physical robots to an extensive number of trials can be impractical, time-consuming, and resource-intensive as the components of the system can wear out or be damaged by haphazard actions.

 Consequently, simulations play a crucial role as they offer a controlled and efficient environment for training RL models. They enable the exploration of diverse scenarios and the refinement of learning algorithms without the constraints imposed by the physical limitations and time constraints associated with real-world trials. Specifically, simulators provide environments where a large amount of data can theoretically be collected in response to the sample inefficiency of RL algorithms.
 This feature also removes the potential damage to the real system that the RL algorithm can cause with its trial-and-error behavior, making it very appealing to the data-driven robotics community.

 However, the accuracy of simulators has not yet caught up with reality, leading to the sim-to-real gap which hinders the deployment of simulation-based optimal policies on real robotic systems. With this problematic and the potential of simulators in mind, researchers have investigated different approaches~\cite{si_perspectives2010, chebotar2019simopt, da_review, rcan2019} in order to produce RL agents trained in simulation and efficient for real world deployment.
 
Among these, \emph{Domain Randomization (DR)} (see~\cite{dr2017Tobin, ruiz2018guidedDR, chen2021understandingdr, weng2019DR}) and \emph{Robust MDP (RMDP)} (see~\cite{rrl2005, bayesian_rrl, action_rrl, mba_rl, mmissspec_rrl, adv_rrl}) have the highest potential for direct and safe policy transfer without policy re-training and without relying on real data collected from (potentially) ill-fitted policies that can damage the system before the proper parameters (and proper policy) have been identified.

 In both methods, a set containing a variety of simulated environments with randomized properties is defined, and the goal is to generate a single policy model that simultaneously maximizes the cumulative reward (or return) of each of the multiple domains from the uncertainty set. Unfortunately, because the agent must be efficient over a wide range of mutually exclusive conditions and depending on the size of the uncertainty set, these approaches can lead to sub-optimal policies that are too conservative once deployed~(\cite{conservative_rrl2019, mozian2020learning}).
 
 To overcome this issue, diverse strategies have been developed by the community, such as dynamic uncertainty sets methods (see~\cite{ramos2019bayessim} and ~\cite{muratore2021data, muratore2022neural}) where a preliminary policy is trained on the full uncertainty set and then deployed to collect (safe) interaction data in order to better tighten the uncertainty set around the target system. The necessity of retraining the policy after each uncertainty update however means that these methods lack flexibility to quickly adapt to different target systems. Another strategy is to employ a recurrent neural network (RNN) such as LSTM for the policy (see~\cite{andrychowicz2020learning, doersch2019sim2real}) which can perform implicit online domain identification. Indeed, RNN-based policies can potentially identify the properties of the current environment and adapt their behavior accordingly. However, the use of RNN comes with its own difficulties, notably with respect to computational cost when combined with off-policy algorithms with replay buffer (~\cite{yang2021recurrent}).
 
In this paper, we focus instead on approaches based on universal policy networks (UPN) with policies conditioned on the domain parameters for explicit online model identification (see~\cite{yu2017uposi, ding2021not}). In contrast to methods based on RNNs, factorizing the controller components into a UPN and a system identifier, means that the overall method can benefit from improvements made independently to each component.

Because uncertainty in the system identification is ubiquitous, methods that train uncertainty-aware UPs capable of adapting to the current uncertainty are necessary.
 Particularly relevant to the present study, \cite{xie2022robust} proposed a mechanism for learning robust policies when given a discrete and finite number of uncertainty sets, based on the idea of identifiability. In their method, one samples randomly a fixed number of uncertainty sets from the full parameter range and then train a policy together with an OSI in order to solve the task. Using the OSI and the notion that some parameters can easily be identified after a few interactions while others may not (or require a very large number of interactions before they are properly identified), their algorithm reduces the undertainty and trains the policy to eventually be robust on the reduced uncertainty set. Although the notion of indentifiability is theoretically appealing to generate less conservative policies, in practice, even if some parameters can easily be identified in simulation, they might not be when deployed on real robot systems, due to many factors such as unmodellable process noise (e.g. air resistance~\citep{semage2022UncAPS}). Indeed, this gives rise to aleatoric uncertainties absent from the simulation model, preventing the perfect identification of the parameters upon deployment.
 
 To overcome this limitation, we tackle the need for algorithms that can generate policies capable of adapting to any of the continuous and infinite number of uncertainties that could arise in practical applications.
The question therefore arises: \emph{what is the best way to learn such uncertainty aware policies?}
 
 In the present work, we show that the answer to this question is tied to the concept of \emph{Convex coverage Set} found in the multi-objective literature, and we investigate various potential algorithms based on the analogy between the problem of learning uncertainty-aware policies in the Multi-Domain RL (MDRL) setting and the problem of learning preference-based policies in Multi-Objective RL (MORL).
 In particular, we propose and evaluate five algorithms, each with various ties to an underlying MORL principle, and all falling under the common goal of solving the Convex Coverage Set (CCS). We also propose a new framework for learning the OSI based on variational inference in order to avoid the effect of redundant parameter representations which could artificially prevent point-like estimates of the true parameters (see~\cite{ding2021not}).
 
 After evaluating all algorithms in simulation on the new metric based on the CCS both on 2d and high-dimensional parameter spaces, we evaluate all proposals on a real robot task using the D'Claw robot (see Figure~\ref{fig:dclaw} and/or \cite{ahn2020robel}).

\section{Background}

 Since our goal is to tie up DR and MORL in order to derive new algorithms for uncertainty-aware policy optimization, we start by recalling the details of each framework.

\subsection{Domain Randomization}

\subsubsection{Reinforcement Learning:}

 Reinforcement learning (RL) is usually formulated as a Markov decision process (MDP) represented by the tuple $\langle \mathcal{S}, \mathcal{A}, \mathcal{P}_{\kappa}, \mu_{\kappa}, r_{\kappa}, \gamma \rangle$, where $\gamma\in (0,1]$ and where we consider a discrete time dynamical system with transition probability $s_{t+1} \sim \mathcal{P}_{\kappa}(s_{t+1} | s_{t}, a_{t})$ for each states $s_t\in\mathcal{S}$ and actions $a_t\in\mathcal{A}$ at time step $t\in\mathbb{N}$. The environment, also known as the domain, is characterized by a set of parameters $\kappa\in\mathcal{U}\subset\mathbb{R}^{n}$ (e.g. masses, link sizes, friction coefficients, etc.) which are considered fixed for each definition of the MDP (i.e. for $\kappa_{1}\neq\kappa_{2}$, $\langle \mathcal{S}, \mathcal{A}, \mathcal{P}_{\kappa_{1}}, \mu_{\kappa_{1}}, r_{\kappa_{1}}, \gamma \rangle$ and $\langle \mathcal{S}, \mathcal{A}, \mathcal{P}_{\kappa_{2}}, \mu_{\kappa_{2}}, r_{\kappa_{2}}, \gamma \rangle$ are two different MDP problems). The initial state $s_0$ is sampled from the initial state distribution $s_0\sim\mu_{\kappa}(s_0)$.
 
 The goal of a RL agent is to maximize the expected discounted sum of rewards $r_{\kappa}(s_t, a_t)$\footnote{in general, the reward is also dependent on the next state and is therefore also a random variable. A common simplification, however, is to consider a deterministic function of the current state and action} which measures the policy's performance and is also often referred to as the \emph{return}. Specifically, for a stochastic policy $\pi_{\theta}(a_t|s_t)$ with parameters $\theta$ acting in the domain with parameters $\kappa$, the objective is to maximize the expected discounted return given by:
\begin{align}
\begin{split}
&J_{\kappa}(\theta) = \mathbb{E}_{s_0\sim\mu_{\kappa}(s_0)} \bigg[\\
& \mathbb{E}_{a_{t}\sim\pi_{\theta}(a_t|s_t), s_{t+1}\sim\mathcal{P}_{\kappa}(s_{t+1} | s_{t}, a_{t})} \big[ \sum\limits_{t=0}^{T-1} \gamma^t r_{\kappa}(s_t, a_t) | s_{0} \big]  \bigg]
\end{split}
\label{eq:expected_disc_return}
\end{align}

 The agent then learns $\pi$ by updating its policy parameters $\theta$ (using for example a SGD-based optimizer) based on experience collected from the environment. The states and actions resulting from the interaction from time step $t=0$ to $t=T-1$ are collected in trajectories $\tau=\{ s_0, a_0, s_1, a_2, \cdots, s_{T-1}, a_{T-1}, s_{T} \}$ and each trajectory is associated with a specific return value.

\subsubsection{Domain Randomization:}

 From equation~\eqref{eq:expected_disc_return}, it is evident that the parameters of the domain, $\kappa$, are assumed to be known and fixed. However, in practice, this is usually not the case. Indeed, when one wishes to deploy a RL agent trained on simulation onto the real world, uncertainties due to unknown differences between the simulator and the real domain need to be accounted for. In such a case, $\kappa$ is no longer a fixed and deterministic event, but a random variable with some distribution $\varpi(\kappa)$ that reflects the prior uncertainty that we have on the real conditions of the target domain (i.e. the real system). In that case, the uncertainty over $\kappa$ must be taken into account and marginalized out, leading to (with $\tau$ a trajectory induced by the policy): 
\begin{align*}
&\mathbb{E}_{\kappa\sim\varpi(\kappa)}\left[ D_{\mathrm{KL}}\left[ q_{\pi_{\theta}}(\tau|\kappa) || p_{\mathrm{opt}}(\tau|\kappa) \right] \right] \\
&= \mathbb{E}_{\kappa\sim\varpi(\kappa)}\left[ \int q_{\pi_{\theta}}(\tau|\kappa) \ln \frac{q_{\pi_{\theta}}(\tau|\kappa)}{p_{\mathrm{opt}}(\tau|\kappa)} d\tau \right] \\
&= \mathbb{E}_{\kappa\sim\varpi(\kappa)}\left[ \mathbb{E}_{\tau\sim q_{\pi_{\theta}}}\left[ \sum\limits_{t=0}^{T-1} \left\{ \gamma^t r_{\kappa}(s_t, a_t) - \ln \pi_{\theta}(a_t|s_t) \right\} \right]  \right]
\end{align*}

\begin{figure}[h]
    \centering
    \includegraphics[keepaspectratio=true,width=\linewidth]{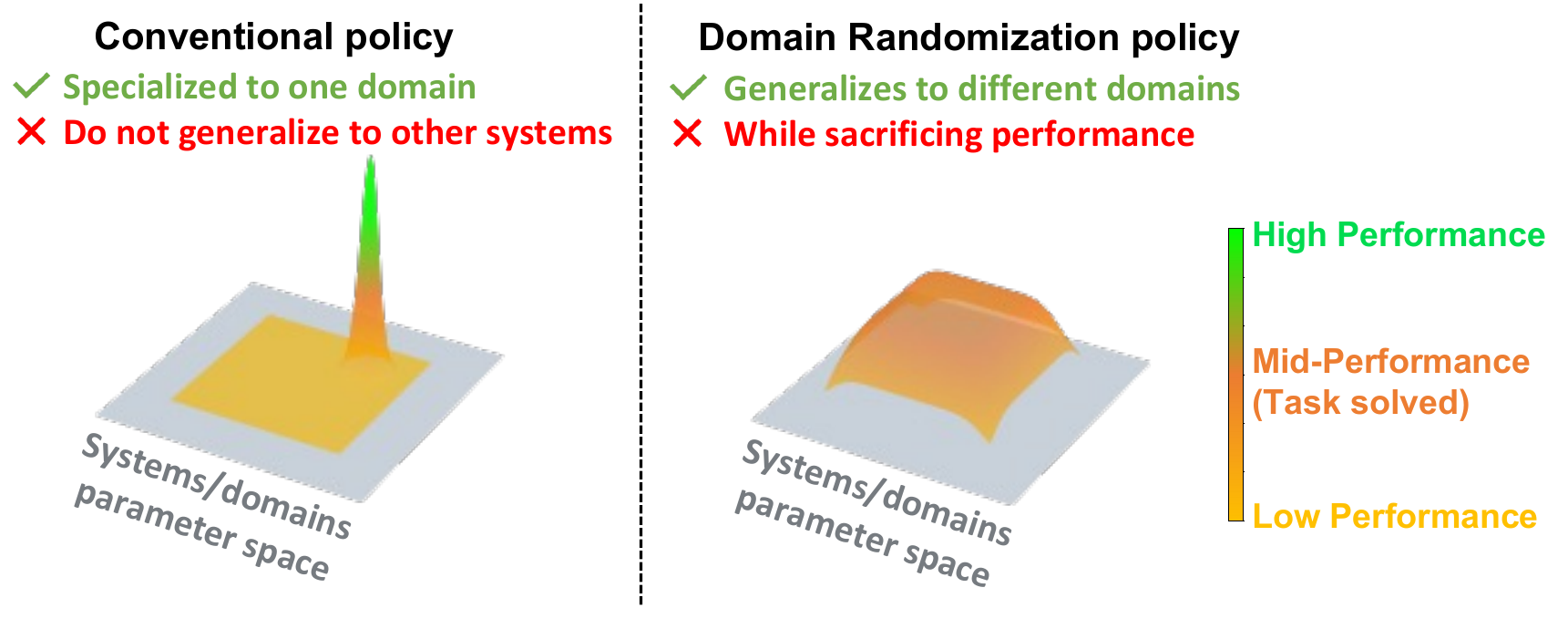}
    \caption{Domain randomization sacrifices the overall performance in order to produce policies which generalize to a wide range of conditions, leading to conservative behaviors. The bigger the randomized domain parameter space is, the more conservative the policy needs to be.}
    \label{fig:dr_conservativeness}
\end{figure}

 This, with the fact that $\mathbb{E}_{\tau\sim q_{\pi_{\theta}}(\tau|\kappa)}[\cdot] = \mathbb{E}_{s_0\sim\mu_{\kappa}(s_0), a_{t}\sim\pi_{\theta}(a_t|s_t), s_{t+1}\sim\mathcal{P}_{\kappa}(s_{t+1} | s_{t}, a_{t})}[\cdot]$, implies that the optimization problem becomes:
\begin{align}
\pi^{*} &= \argmax{\pi_{\theta}}{ \mathbb{E}_{\kappa\sim\varpi(\kappa)}\left[ \mathcal{J}_{\kappa}(\pi_{\theta}) \right] } = \argmax{\pi_{\theta}}{ \mathcal{J}_{\varpi(\kappa)}(\pi_{\theta}) }
\label{eq:dr_objective}
\end{align}
which corresponds to the \emph{Domain Randomization} (DR) objective. $\varpi(\kappa)$ expresses the prior uncertainty over the target domain. The quality of the policy obtained from DR on the target environment will therefore be influenced by how large this uncertainty is. In typical DR applications, one seeks to obtain a policy that can generalize to a wide range of conditions, since knowledge on what kind of environment and in what conditions the agent will have to interact with the system is usually limited. Hence, large uncertainties typically lead to conservative policies that are sub-optimal when deployed, as illustrated in Figure~\ref{fig:dr_conservativeness}.

\subsection{Multi-Objective Reinforcement Learning and Convex Coverage Sets}

\subsubsection{Multi-Objective Reinforcement Learning:}

In the context of multi-objective scenarios, the primary aim is to find solution(s) that strike a favorable balance among various competing objectives. This balance is typically assessed using Pareto dominance, a mechanism for comparing two solutions, as illustrated in Figure~\ref{fig:ccs1}. One solution is said to dominate another if it outperforms it in at least one objective while being at least as good in all other objectives. When two solutions are deemed incomparable, it indicates that each excels in at least one objective compared to the other. Solutions that are dominated hold limited value, as the dominating solution is evidently the preferable choice. Consequently, the top solutions within a set can be pinpointed by retaining only those that either dominate or are incomparable to every other member of the set. When this procedure is applied to the complete set of solutions, the resulting assortment of non-dominated solutions is termed the Pareto optimal front (or the Pareto Coverage Set). This front represents the globally optimal set of compromised solutions.

Seeking a set of compromised solutions offers several advantages over the pursuit of a single ``optimal" solution. Indeed, methods focused on a single solution necessitate predefined user decisions regarding the desired characteristics of that solution. This often requires the users to possess domain knowledge, and minor adjustments in these preferences can lead to significant variations in the resulting solution, potentially accepting a suboptimal outcome. For instance, assigning a slightly higher preference for one objective may hinder the discovery of a solution that significantly improves all other objectives.

The most direct and frequently employed approach to adapt conventional RL algorithms for multi-objective scenarios is to transform these problems into single-objective tasks. This gives rise to Multi-Objective Reinforcement Learning (MORL), which is a framework for multiple criteria decision-making. MORL focuses on the discovery of one or more optimal control policies that can concurrently optimize multiple objective (or reward) functions. The key distinction between single-objective and MORL lies in the nature of the reward. In the former, the reward is a single value (scalar), while in the latter, it is represented as a vector, with each element corresponding to a specific objective.

 Consequently, multi-objective tasks can be converted into single-objective tasks using a process called \emph{scalarization}, where a function is applied to the reward vector to produce a single scalar reward. The most common form of scalarization is a linear combination of the individual objective rewards (see~\cite{natarajan2005dynamicpreference}), allowing the users to control the policy's characteristics by adjusting the weights (or \emph{preferences}) assigned to each objective. However, linear scalarization has a fundamental limitation in that it cannot identify solutions located in concave or linear regions of the Pareto front. Therefore, more complex, non-linear functions tailored to the problem domain are instead occasionally employed (see~\cite{tesauro2007nonlinearscalarization}). Linear scalarization is favored for its simplicity, as it can be seamlessly integrated into existing RL algorithms with minimal modifications.

Mathematically, a Linear-scalarization-based Multi-Objective Markov Decision Process (MOMDP) can be represented by the tuple $\langle \mathcal{S}, \mathcal{A}, \mathcal{P}, \textbf{r}, \Omega, f_{\Omega} \rangle$ where $\mathcal{S}$, $\mathcal{A}$ and $\mathcal{P}(s'|s, a)$ are respectively the state space, action space and transition distribution, $\textbf{r}(s, a)$ is a vector reward function whose components $\textbf{r}_{i}(s, a)$ are the different criteria or objectives with $i \in \{1, \cdots, n\}$, $\Omega$ the space of preference vectors $\boldsymbol{\omega} = \left[ \boldsymbol{\omega}_1, \cdots, \boldsymbol{\omega}_n \right]$ and $f_{\Omega}$ is the space of all linear scalar functions $f_{\boldsymbol{\omega}}(\textbf{r}) = \boldsymbol{\omega}^{\top}\textbf{r} = \sum_{i} \boldsymbol{\omega}_{i}\textbf{r}_{i}$ for which $\boldsymbol{\omega} \in \Omega$. The functions $f_{\boldsymbol{\omega}}(\textbf{r})$ are also called \emph{utility functions} and, as stated before, these are not always necessarily linear functions.

\subsubsection{Convex Coverage Sets:}

\begin{figure*}[tb]
    \centering
    \captionsetup{justification=centering}
    \begin{minipage}[c]{0.5\linewidth}
        \centering
        \includegraphics[keepaspectratio=true,width=\linewidth]{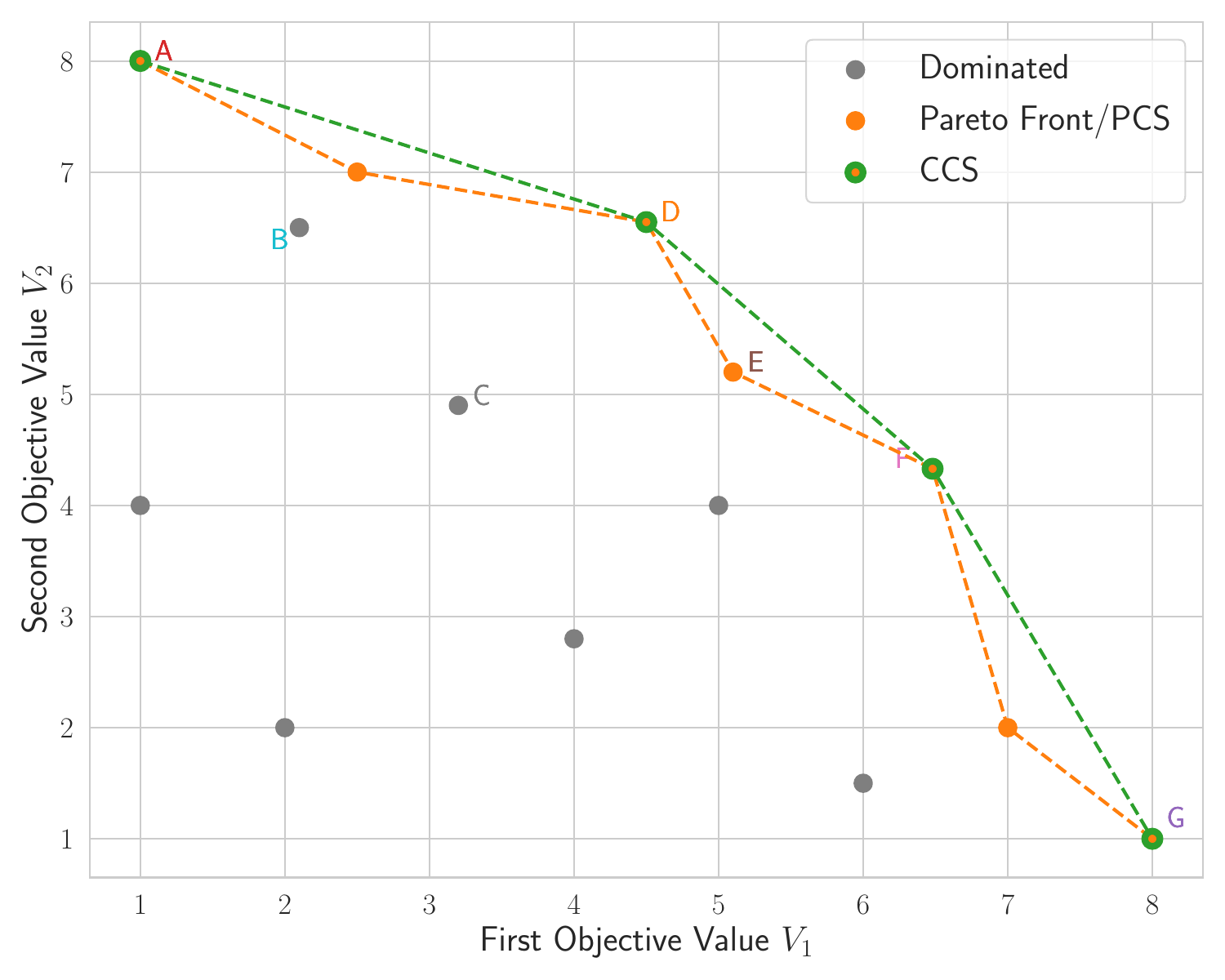}
        \subcaption{Visualization of a Pareto Coverage Set (PCS) and its Convex Coverage Set (CCS)}
        \label{fig:ccs1}
    \end{minipage}%
    \begin{minipage}[c]{0.5\linewidth}
        \centering
        \includegraphics[keepaspectratio=true,width=\linewidth]{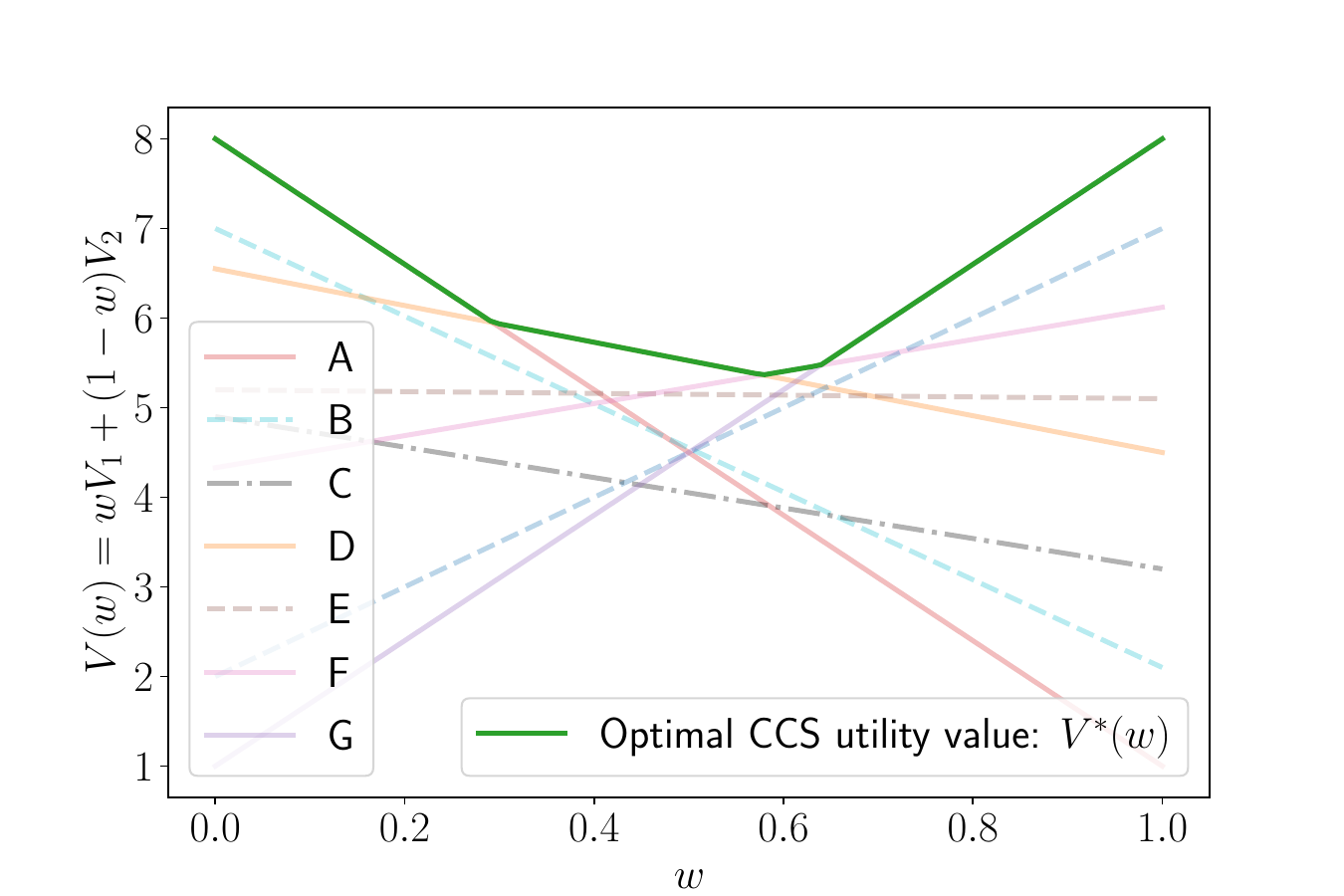}
        \subcaption{Visualization of the piecewise-linear and convex (PWLC) nature of $V^{*}(\boldsymbol{\omega})$}
        \label{fig:ccs2}
    \end{minipage}
    \caption{Visualization of the value functions of different policies $\left\{ A, B, C, D, \cdots \right\}$ on a MORL with two objectives (note that for domain randomization, one domain would correspond to one objective).}
    \label{fig:morl_ccs_visual}
\end{figure*}

 Given a MORL problem, one can summarize all optimal (or non-dominated) cumulative reward (or returns) using a Pareto Coverage Set (PCS):
\begin{align*}
\mathrm{PCS} = \left\{ \textbf{V} \;|\; \nexists \textbf{V}' s.t. \textbf{V}' \succ \textbf{V} \right\}
\end{align*}
where $\textbf{V} = \mathbb{E}_{\pi}\left[ \sum_{t} \gamma^{t} \textbf{r}(s_t, a_t) \right]$ is the expected return and the notation $\textbf{V}' \succ \textbf{V}$ is used here as a short hand for $\textbf{V}_{i}' > \textbf{V}_{i}, \forall i \in \{1, \cdots, n\}$. As explained before, this Pareto front defines the set of undominated expected return vectors (i.e. the set of compromised optimal solutions).
 Based on the utility function, the set of solutions can be further restricted to only include expected return values that provide the maximum cumulative utility. In the case of a linear utility function, this new set, which is always a subset of the PCS, is known as the Convex Coverage Set (CCS) and is defined as:
\begin{align}
\begin{split}
\mathrm{CCS} = \{ \textbf{V} \in \mathrm{PCS} \;|\; \exists \boldsymbol{\omega}\in\Omega \;\mathrm{s.t.}\; &\boldsymbol{\omega}^{\top}\textbf{V} \geq  \boldsymbol{\omega}^{\top}\textbf{V}', \\
& \forall \textbf{V}' \in \mathrm{PCS} \}
\end{split}
\label{eq:val_ccs}
\end{align}
 The CCS is a set of solutions which is optimal for each possible weight vector $\boldsymbol{\omega}$ of a linear scalarization function $f_{\boldsymbol{\omega}}$. It serves as a sufficient set for determining the optimal scalarized value for each weight vector $\boldsymbol{\omega}$: $V^{*}(\boldsymbol{\omega}) = \underset{\textbf{V} \in \mathrm{CCS}}{\max} \boldsymbol{\omega}^{\top}\textbf{V}$. This optimal scalarized value function $V^{*}(\boldsymbol{\omega})$ exhibits a piecewise-linear and convex (PWLC) behavior with respect to the scalarization weights (see Figure~\ref{fig:ccs2}). The task of finding $V^{*}(\boldsymbol{\omega})$ and, consequently, identifying the CCS, amounts to solving the multi-objective decision problem.
A visualization of a PCS and its corresponding CCS in the case of two objectives is given in Figure~\ref{fig:ccs1}.

\section{Pseudo-Multi-Objective Problem for Convex Coverage Set Optimization}

 In this section, we make the connection between DR as a MDRL problem and MORL by proposing a new Markov Decision Process (MDP) framework (named Pseudo-Multi-Objective MDP - PMOMDP) for MDRL, which formalizes the similarities and differences between the two settings. Then, under this framework, we show how three algorithms used for solving the CCS in MORL can be adapted to solve the CCS in MDRL.

\subsection{Multi-Domain Reinforcement Learning as a Pseudo-Multi-Objective Problem}

 Consider a scenario where we have a target environment $E_{\mathrm{target}}$ in which we aim to discover the optimal policy $\pi$. We define an uncertainty set $\mathcal{U}$ by considering various instances of the simulation parameters $\kappa$. This uncertainty set characterizes the ambiguity we have regarding the actual characteristics of our target environment $E_{\mathrm{target}}$. Specifically, we assume that the uncertainty set or randomized domain $\mathcal{U}$ lies within a compact space and includes parameters $\kappa_{\mathrm{target}}$ or a subset that can effectively replicate the characteristics of the true target environment $E_{\mathrm{target}}$.

 This work primarily focuses on reinforcement learning challenges where we have uncertainty about the true transition distribution or dynamics of the environment, i.e. dynamics randomization (\cite{peng2018sim2real}). This type of randomization is relevant in robotics applications in which parameters such as mass, friction, damping, etc., are uncertain and is distinct from scenarios involving visual randomization (\cite{dehban2019impact, volpi2021continual}).

To demonstrate the parallels between this problem setting and MORL, let's begin with the scenario of a discrete uncertainty set $\mathcal{U} = \left\{\kappa_{1}, \kappa_{2}, \cdots, \kappa_{n} \right\}$ for simplicity. We will later extend this to continuous sets, which are the primary focus of domain randomization methods.

In this simplified context, it is apparent that for any given state $s$ and action $a$, each environment $E(\kappa_i)$ produces a unique reward, $r_{\kappa_i}(s, a)$, based on its transition distribution or dynamics denoted as $\mathcal{P}_{\kappa_i}(s'|s, a)$. Specifically, as illustrated in Fig.~\ref{fig:mdrl_morl}, starting from an initial state $s_0$ and executing a sequence of actions $a_0, \cdots, a_{T}$, each environment generates a distinct reward sequence $r_{\kappa_i}(s^{i}_{t}, a_{t})$ resulting in different expected returns $V_{i}(s_{0}) = \mathbb{E}_{s_0\sim\mu(s_0)}\left[ \sum\limits_{t=0}^{T} \gamma^{t} r_{\kappa_i}(s^{i}_{t}, a_{t}) \right]$. Here, each $s^{i}_{t}$ is conditioned on the transition dynamics of the corresponding environment, i.e., $s^{i}_{t} \sim \mathcal{P}_{\kappa_i}(s'|s^{i}_{t-1}, a_{t-1})$, given the shared initial condition $s^{i}_{0} = s^{j}_{0} \sim \mu(s_0)$ for all $i, j \in \{1, \cdots, n \}$.
\Mycomment{ 
\begin{figure}[h]
    \centering
    \includegraphics[keepaspectratio=true,width=0.8\linewidth]{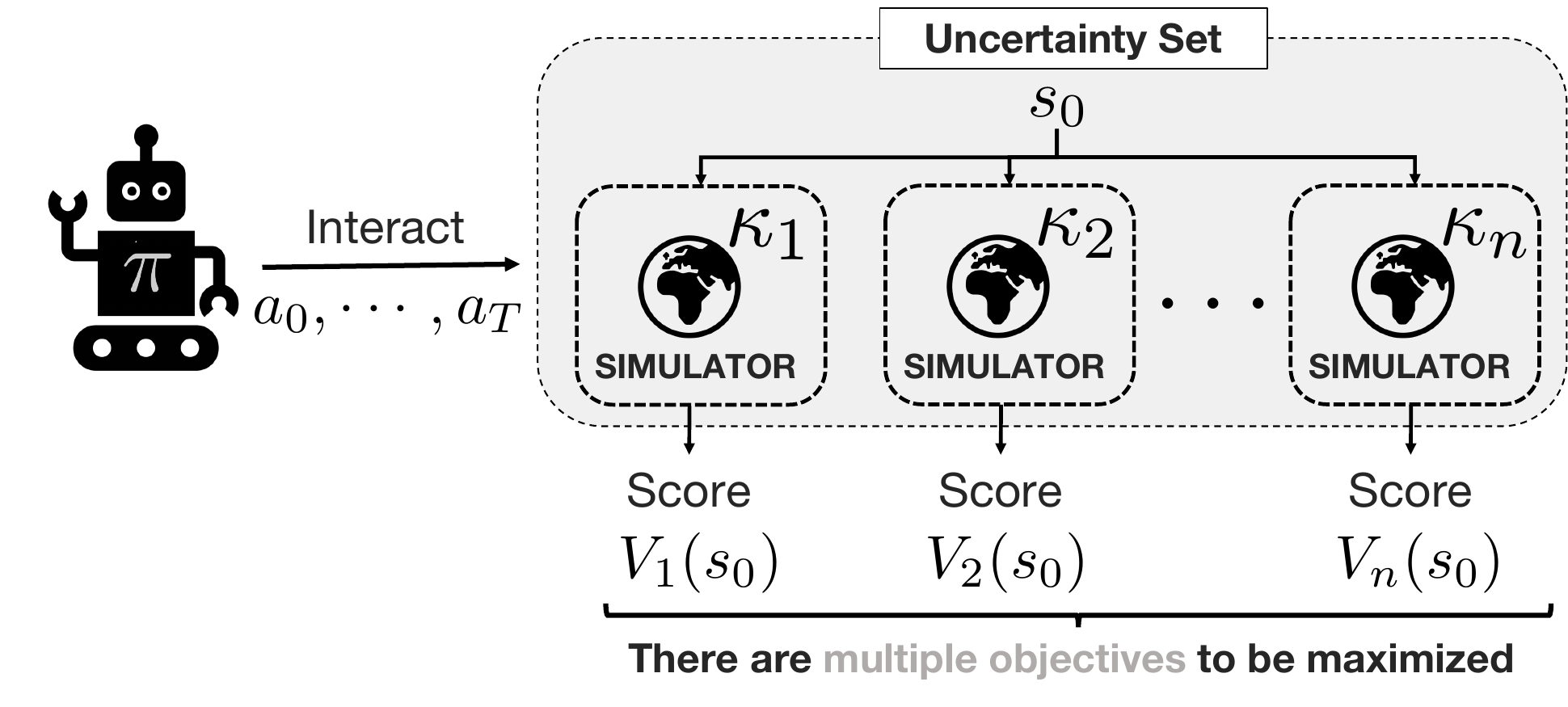}
    \caption{Analogy between multi-objective RL and domain randomization as a multi-domain RL problem.}
    \label{fig:mdrl_morl}
\end{figure}
} 
\begin{figure*}[tb]
    \centering
    \captionsetup{justification=centering}
    \begin{minipage}[c]{0.35\linewidth}
        \centering
        \includegraphics[keepaspectratio=true,width=\linewidth]{mdrl_figures/mdrl_morl}
    \end{minipage}%
    \begin{minipage}[c]{0.7\linewidth}
        \centering
        \includegraphics[keepaspectratio=true,width=\linewidth]{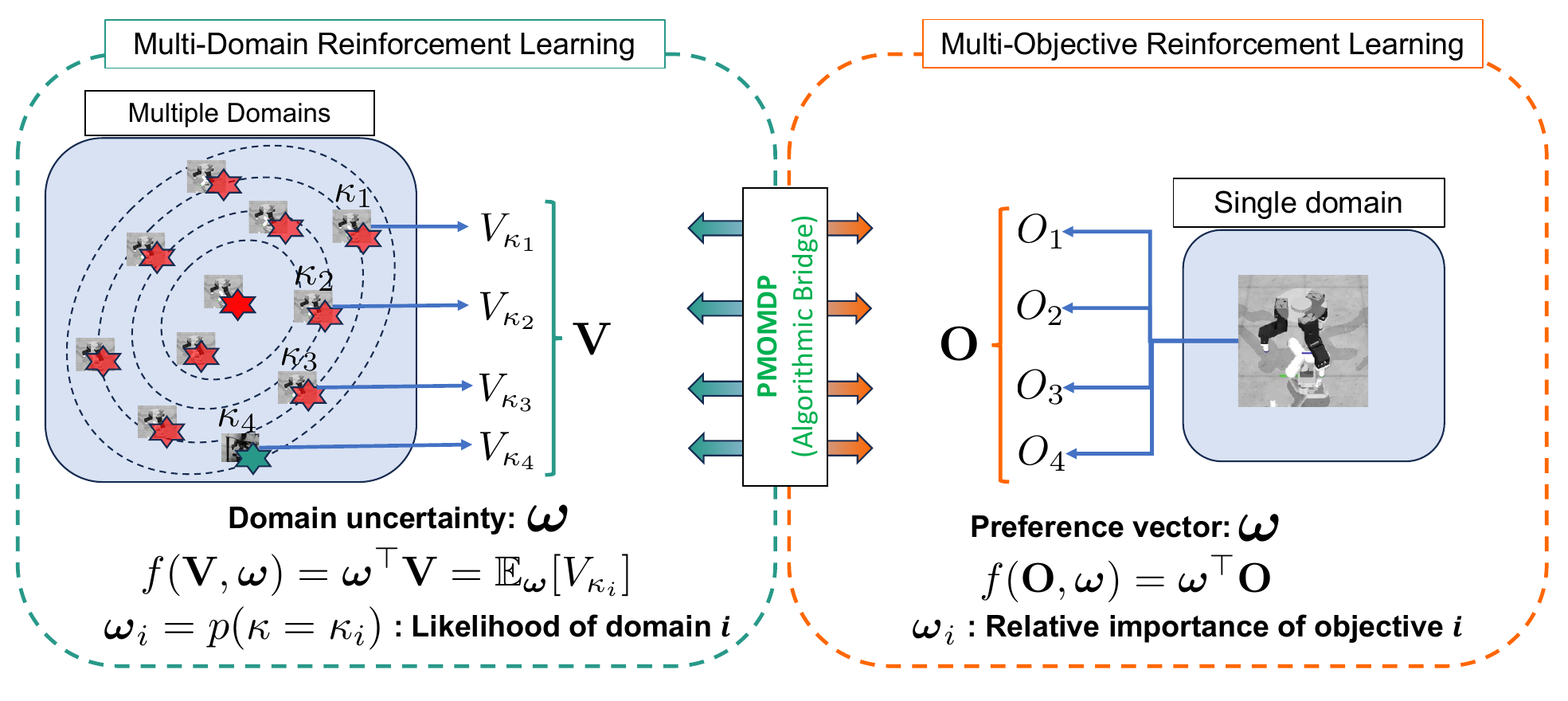}
    \end{minipage}
    \caption{Analogy between multi-objective RL and domain randomization as a multi-domain RL problem.}
    \label{fig:mdrl_morl}
\end{figure*}

The expected returns $V_i$ can be organized in a vector, $\textbf{V} = \left[ V_1, \cdots, V_n \right]$, and following the analogy to MORL, we can express our preferences for each environment by using a preference vector, $\boldsymbol{\omega}$, which assigns a relative importance $\boldsymbol{\omega}_{i}$ to each $V_{i}$. Consequently, we can also define a scalar linear utility function $f_{\boldsymbol{\omega}}(\textbf{V}) = \boldsymbol{\omega}^{\top}\textbf{V}$.

Without loss of generality, let's assume that $\boldsymbol{\omega}$ is structured such that $\sum_{i} \boldsymbol{\omega}_{i} = 1$. In this context, the preference vector $\boldsymbol{\omega}$ can be regarded as a discrete probability distribution. Each element $\boldsymbol{\omega}_{i}$ then represents the probability that our target environment corresponds to environment $E(\kappa_{i})$, specifically, $\boldsymbol{\omega}_{i} \doteq p(\kappa_{\mathrm{target}} = \kappa_{i})$. Furthermore, the utility function essentially transforms into an expected value, i.e. $f_{\boldsymbol{\omega}}(\textbf{V}) = \boldsymbol{\omega}^{\top}\textbf{V} = \sum\limits_{i=1}^{n} w_{i}V_{i} = \mathbb{E}_{\kappa_i\sim\boldsymbol{\omega}}[V_{i}]$.

 Since we are interested in optimizing over control policies $\pi$, we can make the value functions explicitly dependent on the policy $\pi$ and rewrite the utility function as:
\begin{align*}
f_{\boldsymbol{\omega}}(\textbf{V}^{\pi}(s_{0})) &= \mathbb{E}_{\kappa_{i}\sim \boldsymbol{\omega}}[V^{\pi}(s_{0}, \kappa_{i})] \\
&\mathrm{with} \; \textbf{V}^{\pi}(s_{0}) = \left[ V^{\pi}(s_{0}, \kappa_{1}), \cdots, V^{\pi}(s_{0}, \kappa_{n}) \right] \\
V^{\pi}(s_{0}, \kappa_{i}) &= \mathbb{E}_{s_0\sim\mu(s_0), a_{t}\sim\pi(a_{t}|s_{t})}\left[ \sum\limits_{t=0}^{T} \gamma^{t} r_{\kappa_i}(s^{i}_{t}, a_{t}) \right]
\end{align*}
By analogy to equation~\eqref{eq:val_ccs}, the PCS and the CCS can then be defined over the set of all policies $\Pi$ as:
\begin{align}
\begin{split}
&\mathrm{PCS} \doteq \bigl\{ \pi^{*} |\; V^{\pi^{*}}(s) \succ V^{\pi}(s),\; \forall \pi \in \Pi, \forall s \in \mathcal{S} \bigr\}
\end{split}\\
\begin{split}
&\mathrm{CCS} \doteq \bigl\{ \pi^{*} \in \mathrm{PCS} \;|\; \exists \boldsymbol{\omega} \in \Omega \; \mathrm{s.t.} \bigr.\\
&\bigl. \mathbb{E}_{\kappa \sim \boldsymbol{\omega}}[V^{\pi^{*}}(s, \kappa)] \geq \mathbb{E}_{\kappa\sim \boldsymbol{\omega}}[V^{\pi}(s, \kappa)],\; \forall \pi \in \mathrm{PCS}, \forall s \in \mathcal{S} \bigr\}
\end{split}
\label{eq:policy_convexcov}
\end{align}
from which the optimal policy corresponding to the optimal value $V^{*}(\boldsymbol{\omega}) = \underset{\textbf{V} \in \mathrm{CCS}}{\max} f_{\boldsymbol{\omega}}(\textbf{V})$ for a given parameters uncertainty $\boldsymbol{\omega}$ can explicitly be defined as:
\begin{align}
\pi^{*}_{\boldsymbol{\omega}} &= \argmax{\pi}{\mathbb{E}_{\kappa\sim \boldsymbol{\omega}}[V^{\pi}(s, \kappa)]} \\
&= \argmax{\pi_{\theta}}{ \mathbb{E}_{\kappa\sim\varpi(\kappa)}\left[ \mathcal{J}_{\kappa}(\pi_{\theta}) \right] } = \argmax{\pi_{\theta}}{ \mathcal{J}_{\varpi(\kappa)}(\pi_{\theta}) }
\label{eq:opt_policy}
\end{align}

 A quick look back at equation~\eqref{eq:dr_objective} lets us notice that we have again arrived at the domain randomization objective, \emph{from a MORL perspective}.
 The rewriting of the utility function as an expectation also allows a direct generalization to continuous uncertainty sets, e.g. when considering $d$-dimensional parameters $\kappa$ that can take any real value within some boundaries, i.e. $\mathcal{U} \subset \mathbb{R}^{d}$. Indeed, in such case, the preference distribution $\boldsymbol{\omega}$ then corresponds to a continuous distribution and the utility function is defined as $f_{\boldsymbol{\omega}}(\textbf{V}^{\pi}(s)) = \mathbb{E}_{\kappa\sim \boldsymbol{\omega}}[V^{\pi}(s, \kappa)] = \int_{\Omega} \boldsymbol{\omega}(\kappa)V^{\pi}(s, \kappa)d\kappa$, where the sum of the inner product is replaced by an integral.

 As noticed, if $\boldsymbol{\omega}$ (or $\varpi(\kappa)$ for the continuous case) is given and fixed, then \eqref{eq:opt_policy} is simply the domain randomization with prior uncertainty $\boldsymbol{\omega}$ (respectively $\varpi(\kappa)$). However, MORL is concerned with solving the full CCS, i.e. $\boldsymbol{\omega}$ can take any value within the set of preferences $\Omega$, allowing for the solution to change when the preference vector is updated.
 This feature is of particular interest to domain randomization where the uncertainty set usually corresponds to the prior knowledge. Indeed, after interaction with the environment, it is usually possible to apply some system identification (SI) algorithms in order to get a better estimate of the target environment. The ability to handle various adaptive posterior uncertainty is captured through the CCS of \eqref{eq:opt_policy}.

 Hence, similarly to the MORL framework, we are interested in this work in learning the full CCS by finding a policy that can generalize across the entire space of uncertainties. Using the conceptualization discussed above, we can mathematically cast this MDRL problem into a Pseudo-Multi-Objective Markov Decision Process:
\begin{definition}[Pseudo-MOMDP or PMOMDP]
 Let $\mathcal{U}$ be an uncertainty set, and let $\boldsymbol{\omega}$ be a (continuous or discrete) distribution over $\mathcal{U}$. Also define $\kappa$ to be the parameters associated with the different environments $E$ suct that $\kappa \in \mathcal{U}$. Then, we define the PMOMDP by the tuple $\langle \mathcal{S}, \mathcal{A}, \mathcal{P}_{\kappa}, \mu_{\kappa}, r_{\kappa}, \gamma, \Omega, f_{\Omega} \rangle$, where $\mathcal{S}$ and $\mathcal{A}$ are respectively the state and action spaces, $\gamma$ the discount factor, $\mu_{\kappa}(s_0)$ and $\mathcal{P}_{\kappa}(\cdot | s, a)$ are the parameterized initial state and transition distributions with parameters $\kappa$, $r_{\kappa}(s, a)$ is a reward function that can depend on the transition $\mathcal{P}_{\kappa}$, $\Omega$ is the space of preference distributions $\boldsymbol{\omega}$ and $f_{\Omega}$ is the space of scalar utility functions defined by $f_{\boldsymbol{\omega}}(\cdot) \in \mathbb{R}$.
\end{definition}

 Typically, $\Omega$ will be defined by the full randomized domain, i.e. the prior uncertainty of the DR sub-problem, and $\boldsymbol{\omega}$ will be all distributions whose support lie within this full randomized domain. In the next sections, we describe how the PMOMDP can be used to solve the MDRL problem expressed above, through the adaptation of MORL algorithms.

\subsection{Multi-Domain Uncertainty-Aware CCS Optimization}

 In this section, for simplicity, we will use the discrete formulation of MDRL (i.e. we have a discrete and finite number of environments $\mathcal{U} = \left\{ \kappa_{1}, \cdots, \kappa_{n} \right\}$, $n\in\mathbb{N}$). Hence, the bold letters and symbols are vectors, e.g. $\textbf{r}$ and $\boldsymbol{\omega}$ are respectively the vectors $[r_1, \cdots, r_n]$ and $[\omega_1, \cdots, \omega_n]$ and $\textbf{Q}(s, a)$ is a vector of Q-values $[Q_1(s,a), \cdots, Q_n(s,a)]$.

The real valued uncertainty set case, i.e. $\mathcal{U} \subset \mathbb{R}^{d}$, is easily obtained by replacing the inner product by an expectation and replacing the vector representation $\boldsymbol{\omega}$ with a continuous distribution $\boldsymbol{\omega}(\kappa)$, $\kappa\in\mathcal{U}$.
 We also recall the relation between the state value function $\textbf{V}(s)$ and state-action value function $\textbf{Q}(s, a)$ which is: $\textbf{V}(s) = \mathbb{E}_{a\sim\pi}\left[ \textbf{Q}(s, a) \right]$. This relation implies that $\textbf{V}(s)$ and $\textbf{Q}(s, a)$ both represent the expected return and the RL objective can be written using either of them.
 
 Remember that our goal is to solve for the CCS optimal policy $\pi^{*}$ corresponding to the optimal scalarized value function, for each uncertainty $\boldsymbol{\omega}$, i.e. $V^{*}(s, \boldsymbol{\omega}) = \underset{\textbf{V} \in \mathrm{CCS}}{\max} \boldsymbol{\omega}^{\top}\textbf{V}(s, \boldsymbol{\omega})$, or in terms of the Q-function, $Q^{*}(s, a, \boldsymbol{\omega}) = \underset{\textbf{Q} \in \mathrm{CCS}}{\max} \boldsymbol{\omega}^{\top}\textbf{Q}(s, a, \boldsymbol{\omega})$.
 
\subsubsection{\textbf{Conditioned MDRL:}}

 One of the easiest and most naive approach is to acquire the coverage set by learning a set of policies, with each policy represented by a neural network. This would be done by evaluating a series of scalarization problems, each with its own associated value vector (representing the values for each domain covered by the uncertainty vector), for every conceivable uncertainty vector that might arise in practice. Such an approach would be exactly similar to the scalarized MORL algorithm from~\cite{mossalam2016sMORL}.

 Since, for each problem, $\boldsymbol{\omega}$ is fixed and given, this scalarized MDRL approach is simply equivalent to solving a sequence of Domain Randomization (DR) problems with different uncertainty sets, in which we typically learn a scalar value function $Q(s, a)$ which is related to the vector $\textbf{Q}$ by the following relation $Q(s, a) = \boldsymbol{\omega}^{\top}\textbf{Q}(s, a)$. For each problem solved within the sequence, the key difference from typical DR would therefore reside in the explicit vectorization of the state-action value $\textbf{Q}(s, a)$ before applying the scalarization, instead of learning directly the scalarized value function $Q(s, a)$.
 
 Although the scalarized MDRL algorithm described in the previous paragraphs is an offline setting where the CCS is fully learned before the deployment, it is also possible to use it in an online setting, i.e. if the weight vector $\boldsymbol{\omega}$ is allowed to change over time. Indeed, in that case, instead of pre-building the CCS set $\left\{ \pi^{*}_{\boldsymbol{\omega}^{i}}(s) \right\}_{i\in\mathbb{N}}$, it suffices to simply retrain the policy and Q-value each time $\boldsymbol{\omega}$ changes. This corresponds to applications of DR where a system identifier is used to reduce the domain uncertainty and then the new uncertainty set is used to retrain the policy (as done by~\cite{chebotar2019simopt, ramos2019bayessim, muratore2022neuraldr}).

\begin{figure}[h]
    \centering
    \includegraphics[keepaspectratio=true,width=\linewidth]{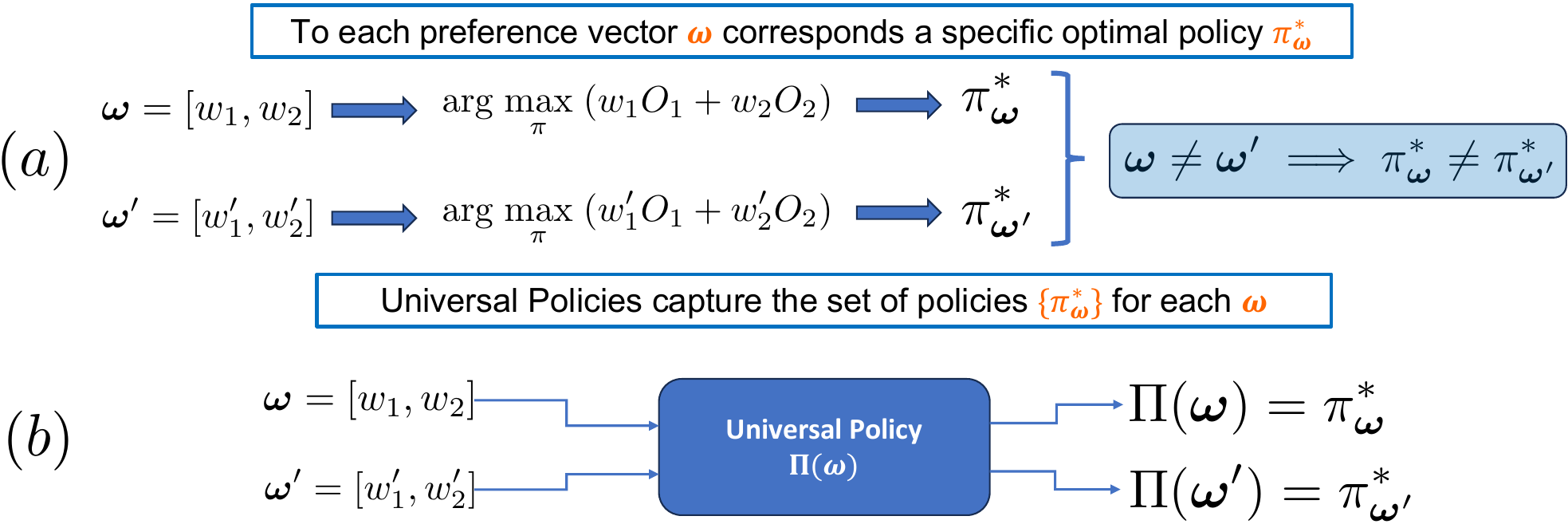}
    \caption{The options available to us for learning the CCS here illustrated in the case where we have two discrete uncertainties or preferences $\boldsymbol{\varpi}$ and $\boldsymbol{\varpi}'$: (a) Learn two different policies, one for each uncertainty vector, as performed in the sMORL approach. (b) Learn a single function of the uncertainty vector that takes the vector as input and generates the corresponding optimal policy. This function of $\boldsymbol{\varpi}$ is known as a Universal Policy (UP).}
    \label{fig:up_morl connections}
\end{figure}

 Unfortunately, just like in MORL, it is often not feasible to repeatedly solve the decision problem each time the prior uncertainty $\boldsymbol{\omega}$ changes. For example, when an agent is deployed in the real environment and left to interact with uncertain or shifting environmental conditions.
Hence, algorithms that do not need to retrain the policy each time the uncertainty is updated by the (online) system identifier are required.
Furthermore, instead of building a finite coverage set by learning multiple NNs for a finite number of uncertainties $\boldsymbol{\omega}$, we wish to learn a single policy that can continuously adapt to different uncertainty levels $\boldsymbol{\omega}$.
 Naturally, these requirements translate to the use of Universal Policy Networks (UPN) (see Fig.~\ref{fig:up_morl connections}).

 The most straightforward universal network approach to solving the CCS MORL is to learn a policy and a value function conditioned on the uncertainty vector such that they are able to differentiate between different $\boldsymbol{\omega}$. For example, if we have two different weights $\boldsymbol{\omega}_1$ and $\boldsymbol{\omega}_2$, then we can learn $\left( \pi^{*}(\cdot|s_{t}, \boldsymbol{\omega}), \textbf{Q}^{*}(s_t, a_t, \boldsymbol{\omega}) \right)$ such that $\left( \pi^{*}(\cdot|s_{t}, \boldsymbol{\omega}_{1}), \textbf{Q}^{*}(s_t, a_t, \boldsymbol{\omega}_1) \right)$ will theoretically correspond to the policy and vector-valued Q-function trained using a scalarized DR (sDR) algorithm with a fixed $\boldsymbol{\omega}=\boldsymbol{\omega}_1$ and $\left( \pi^{*}(\cdot|s_{t}, \boldsymbol{\omega}_{2}), \textbf{Q}^{*}(s_t, a_t, \boldsymbol{\omega}_2) \right)$ will theoretically correspond to the policy and vector-valued Q-function trained using sDR with $\boldsymbol{\omega}=\boldsymbol{\omega}_2$.

 Specifically, the Conditioned MDRL (cMDRL)'s Bellman equation (which can be associated to the conditioned MORL by \cite{abels2019cMORL}) under the PMOMDP is then simply given by:
\begin{align}
\begin{split}
\mathcal{T}^{\pi}Q_{i}(s_t, &a_t, \boldsymbol{\omega}) = r_{i}(s_t, a_t) \\
&+ \gamma \mathbb{E}_{a'\sim\pi(\cdot|s_{t+1}, \boldsymbol{\omega})}\left[ Q_i\left(s^{i}_{t+1}, a', \boldsymbol{\omega} \right) \right]
\end{split}
\label{eq:cmdrl}
\end{align}
 As can be noticed, $\textbf{Q}(s, a', \boldsymbol{\omega})$ represents the expected vectorized return under the stochastic policy $\pi(\cdot|s, \boldsymbol{\omega}) = \pi_{\boldsymbol{\omega}}(\cdot|s)$ after taking action $a'$ in state $s$. Hence, $\textbf{Q}(s, a', \boldsymbol{\omega})$ can be rewritten as $\textbf{Q}^{\pi_{\boldsymbol{\omega}}}(s, a')$.
 
Note that if instead of the linear utility function, we defined the utility function as $f(v^{\pi}) = \underset{\kappa}{\min}\;{v^{\pi}(s, \kappa)}$, we would derive the following optimal policy $\pi^{*}_{\varpi} = \argmax{\pi}{ \underset{\kappa\sim\boldsymbol{\omega}}{\min}\;{v^{\pi}(s, \kappa, \boldsymbol{\omega})} }$ from the convex set, which corresponds exactly to the Robust Reinforcement Learning (RRL) objective, solving for the worst environment from the uncertainty set defined by $\boldsymbol{\omega}$. 

 As mentioned before, in this work we mainly focus only on linear utility functions of the form $f_{\boldsymbol{\omega}}(\cdot) = \mathbb{E}_{\kappa\sim\boldsymbol{\omega}}[\cdot] \in \mathbb{R}$.

\subsubsection{\textbf{MDRL with CCS optimality filter (Envelope MDRL):}}

 In the MORL literature, the CCS optimality filter or envelope MORL algorithm is an algorithm developed by~\cite{yang2019generalizedMORL} which takes the concepts related to universal networks and conditioned MORL a step further. It includes an explicit CCS optimality filter to the Bellman equation in order for the algorithm to be able to quickly identify Q-values $\textbf{Q}(s_{t+1}, a', \boldsymbol{\omega}^{*})$ that are more optimal under the weight $\boldsymbol{\omega}$. Indeed, recall that if $\boldsymbol{\omega} \neq \boldsymbol{\omega}^{*}$, then $\textbf{Q}(s_{t+1}, a', \boldsymbol{\omega}) \neq \textbf{Q}(s_{t+1}, a', \boldsymbol{\omega}^{*})$ and from the definition of the CCS in~\eqref{eq:policy_convexcov} rewritten with respect to the state-action value:
\begin{align}
\begin{split}
&\mathrm{CCS} \doteq \bigl\{ \pi \in \mathrm{PCS} \;|\; \exists \boldsymbol{\omega} \in \Omega \; \mathrm{s.t.}\\
&\boldsymbol{\omega}^{\top}\textbf{Q}^{\pi_{\boldsymbol{\omega}}}(s_{t+1}, a') \geq \boldsymbol{\omega}^{\top}\textbf{Q}^{\pi_{\boldsymbol{\omega}'}}(s_{t+1}, a'),\; \forall \boldsymbol{\omega}' \in \Omega, \forall s \in \mathcal{S} \bigr\}
\end{split}
\label{eq:policy_convexcov}
\end{align}
 we see that if $\boldsymbol{\omega} \neq \boldsymbol{\omega}^{*}$ and $\boldsymbol{\omega}^{\top}\textbf{Q}^{\pi_{\boldsymbol{\omega}^{*}}}(s_{t+1}, a') \geq \boldsymbol{\omega}^{\top}\textbf{Q}^{\pi_{\boldsymbol{\omega}}}(s_{t+1}, a')$, then $\pi_{\boldsymbol{\omega}} \notin \mathrm{CCS}$ and the CCS optimal value under $\boldsymbol{\omega}$ should be set as $\textbf{Q}^{*}(s_{t+1}, a') = \underset{\pi_{\boldsymbol{\omega}'}}{\max}\; \boldsymbol{\omega}^{\top}\textbf{Q}^{\pi_{\boldsymbol{\omega}'}}(s_{t+1}, a') = \textbf{Q}^{\pi_{\boldsymbol{\omega}^{*}}}(s_{t+1}, a')$ and $\pi_{\boldsymbol{\omega}}$ should be updated as $\pi_{\boldsymbol{\omega}} \leftarrow \pi_{\boldsymbol{\omega}^{*}}$.
 
 The Envelope MORL's Bellman equation that achieves this is defined as:
\begin{align}
\mathcal{T}Q_{i}(s_t, a_t, \boldsymbol{\omega}) &= r_{i}(s_t, a_t) + \gamma Q_i\left(s_{t+1}, a^{*}, \boldsymbol{\omega}^{*}\right)
\label{eq:emorl}\\
a^{*}, \boldsymbol{\omega}^{*} &= \argmax{a', \boldsymbol{\omega}'}{\boldsymbol{\omega}^{\top}\textbf{Q}(s_{t+1}, a', \boldsymbol{\omega}')}
\end{align}
where the second equation is referred to as the \emph{optimality filter}.

 The Bellman equation of the corresponding Envelope MDRL (noted eMDRL) under the PMOMDP is then simply obtained by:
\begin{align}
\begin{split}
&\mathcal{T}^{\pi}Q_{i}(s_t, a_t, \boldsymbol{\omega}) = \\
&\;\;\;r_{i}(s_t, a_t) + \gamma \mathbb{E}_{a'\sim\pi(\cdot|s_{t+1}, \boldsymbol{\omega}^{*})}\left[ Q_i\left(s^{i}_{t+1}, a', \boldsymbol{\omega}^{*} \right) \right]
\end{split}
\label{eq:emdrl}\\
&\boldsymbol{\omega}^{*} = \argmax{\boldsymbol{\omega}'}{\mathbb{E}_{a'\sim\pi(\cdot|s_{t+1}, \boldsymbol{\omega}')}\left[ \boldsymbol{\omega}^{\top}\textbf{Q}(s^{i}_{t+1}, a', \boldsymbol{\omega}') \right]}
\label{eq:optim_filter}
\end{align}
where we take care to emphasize that each environment (or domain) indexed by $i$ will produce its own next state $s^{i}_{t+1}$ which will typically be different from the other environments' next states. This is the main difference between the PMOMDP and the MOMDP frameworks, since in MORL problems, the next state is always the same for all objectives.

\begin{figure}[h]
    \centering
    \includegraphics[keepaspectratio=true,width=0.9\linewidth]{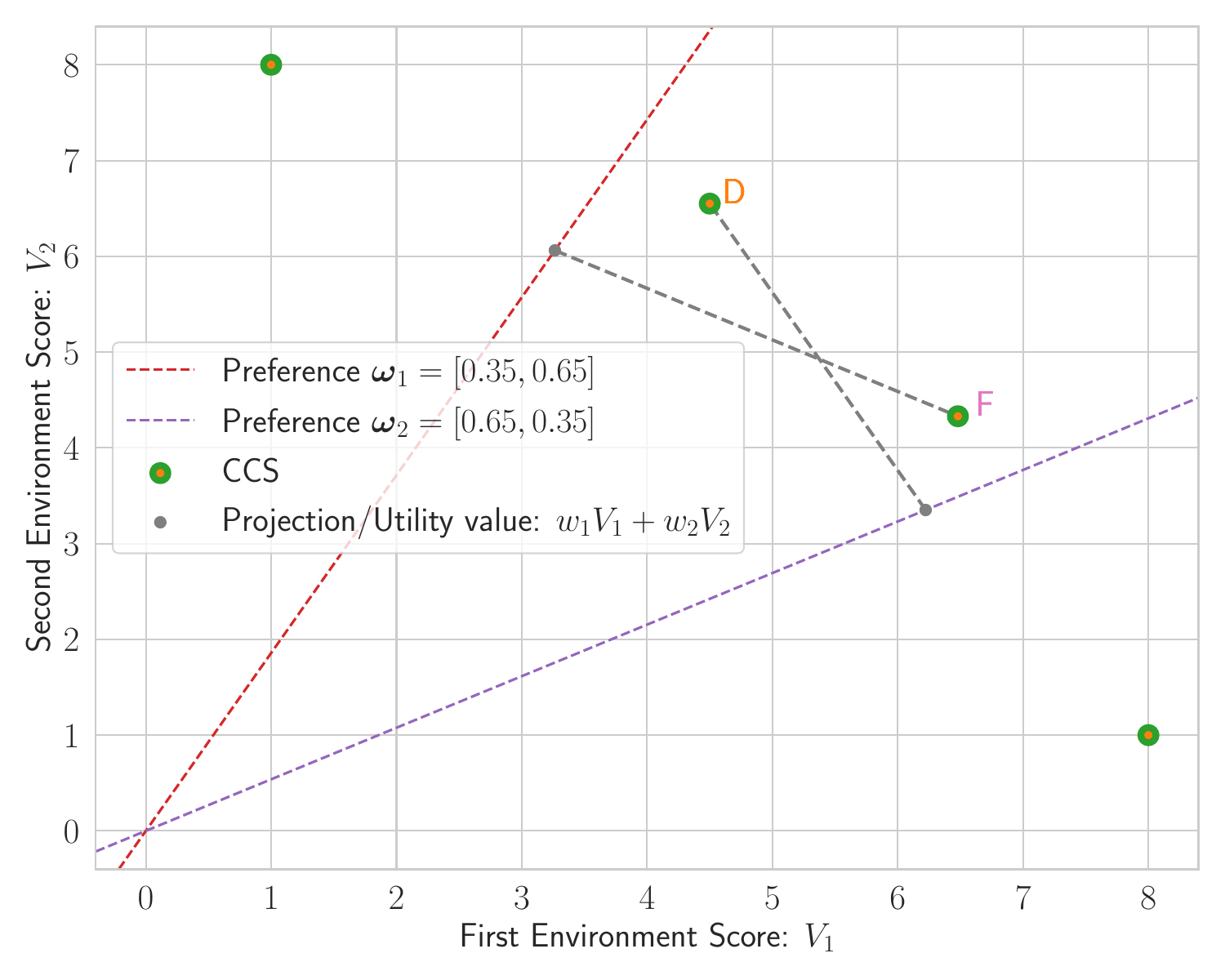}
    \caption{Visualization of a missaligned CCS solution example.}
    \label{fig:ccs_missaligned}
\end{figure}

 The advantage of the optimality filter of equation~\eqref{eq:optim_filter} is to avoid or alleviate potential misalignment of the CCS solutions during the learning process and allow the algorithm to quickly find the optimal value $\textbf{Q}^{*}(s_{t+1}, a')$ corresponding to each $\boldsymbol{\omega}$. Indeed, as illustrated in Figure~\ref{fig:ccs_missaligned}, the sample inefficiency of RL can cause the algorithm to assign non-optimal policies to different $\boldsymbol{\omega}$ (the policy $D$ is assigned to $\boldsymbol{\omega}_{2}$ instead of $\boldsymbol{\omega}_{1}$ and the policy $F$ is assigned to $\boldsymbol{\omega}_{1}$ instead of $\boldsymbol{\omega}_{2}$). The correction of such a misaligned solution may require many iterations of the algorithm in the typical conditioned MORL/MDRL case, even if the policy $D$ has been experienced under $\boldsymbol{\omega}_{1}$. The optimality filter introduced in the envelope algorithm seeks to avoid such a local optimum by explicitly solving for the best solution within the CCS for each $\boldsymbol{\omega}$. For example, as illustrated in Figure~\ref{fig:ccs_filter}, in the case of $\boldsymbol{\omega}_{1}$, equation~\eqref{eq:optim_filter} basically solves the equation $\underset{\pi\in\mathrm{CCS}}{\min} \norm{\boldsymbol{\omega}_{1}^{\top}\left(\textbf{V}_{\pi} - \textbf{V}_D\right)}$. The algorithm can therefore quickly move to the optimal value corresponding to each uncertainty.

 Unfortunately, in practice, the use of approximate neural network functions and more critically, the use of a target network in certain Deep RL algorithms can introduce some bias when the maximum is taken in order to solve the CCS optimality filter (for example in the Soft Actor-Critic algorithm\footnote{For the comparison using the SAC algorithm, we shall make modifications to the theoretical definition in order to avoid the application of the envelope filter on the target networks}. Such a bias could defeat the theoretical advantage of the envelope filter and lead the algorithm astray.
Hence, it is desirable to have a strategy that serves the same goal as the envelope filter solution above, but without the need to explicitly take the maximum of the value function.
 
 To fulfill such requirement, we again exploit the PMOMDP and the analogy between MORL and MDRL, as described in the next section.

\subsubsection{\textbf{Utopia-based MDRL:}}

 In the MORL litterature (see ~\cite{van2013scalarized}), there exists an algorithm based on a particular loss function that introduces an utopian point $\textbf{z}^{*}$:
\begin{align}
L &= \sqrt{ \boldsymbol{\omega}^{\top}(\textbf{z}^{*} - \textbf{Q}(s_t, a_t))^{2} } = \norm{\textbf{z}^{*} - \textbf{Q}(s_t, a_t)}_{\boldsymbol{\omega}}
\end{align}
where the utopian point $\textbf{z}^{*}$ is a parameter that is being constantly adjusted during the learning process by recording the best value so far for each objective, augmented by a small positive constant $\epsilon$ (or a negative constant if  the problem is to be minimized). This point $\textbf{z}^{*}$ serves as a point of attraction in order to steer the search process towards high-quality solutions.

 We can notice that the envelope filter in eMORL/eMDRL actually serves as a way to transform the TD target into an utopian point, i.e. in eMORL/eMDRL, $\textbf{z}^{*} = r_{i}(s_t, a_t) + \gamma \mathbb{E}_{a'\sim\pi(\cdot|s_{t+1}, \boldsymbol{\omega})}\left[ Q_i\left(s^{i}_{t+1}, a', \boldsymbol{\omega}^{*} \right) \right]$.
 The difference is that the utopian point in that case is not theoretically ``utopian", since $\textbf{z}^{*}$ lies within the optimization space, i.e. it is a point of the CCS and is therefore reachable (there is a realistic policy corresponding to $\textbf{z}^{*}$).
 
\begin{figure*}[tb]
    \centering
    \captionsetup{justification=centering}
    \begin{minipage}[c]{0.5\linewidth}
        \centering
        \includegraphics[keepaspectratio=true,width=\linewidth]{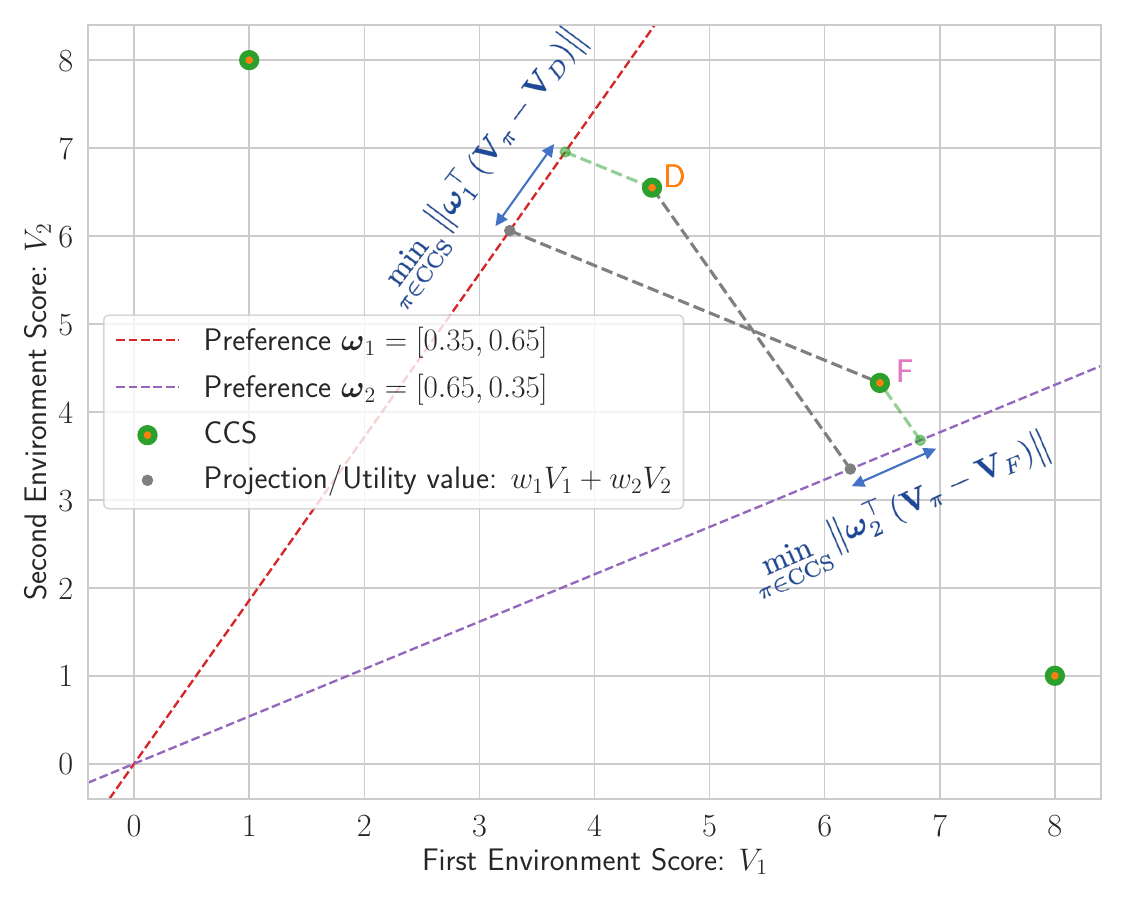}
        \subcaption{Envelope-based Filter}
        \label{fig:ccs_filter}
    \end{minipage}%
    \begin{minipage}[c]{0.5\linewidth}
        \centering
        \includegraphics[keepaspectratio=true,width=\linewidth]{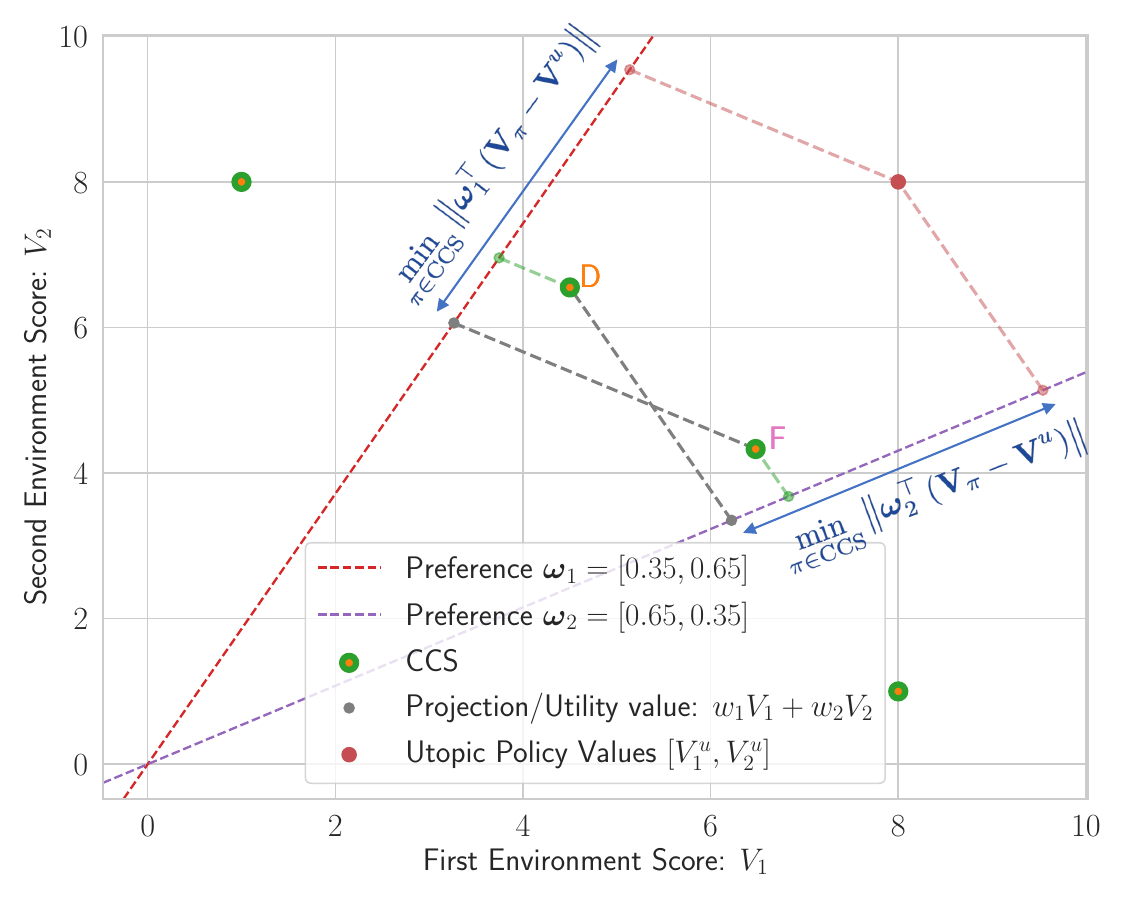}
        \subcaption{Utopia-based Filter}
        \label{fig:ccs_utopia}
    \end{minipage}
    \caption{Visualization of the difference between the \emph{Envelope-based Filter} and the \emph{Utopia-based Filter}. The red dot on the right figure represents an utopian policy capable of reaching the highest score on each environment simultaneously.}
    \label{fig:morl_ccs_optimfilt}
\end{figure*}

 However, in practice, $\textbf{z}^{*}$ need not be reachable and instead of using the same utopian point as in the eMDRL algorithm, we can use a different point defined as $\textbf{z}^{*} = r_{i}(s_t, a_t) + \gamma \mathbb{E}_{a'\sim\pi(\cdot|s_{t+1}, \boldsymbol{\omega})}\left[ Q_i\left(s^{i}_{t+1}, a', \delta_i \right) \right]$ as illustrated in Figure~\ref{fig:ccs_utopia} and where $\delta_i = \delta(\kappa - \kappa_{i})$ is the Dirac delta function. 
In this case, because $Q_i\left(s, a', \delta_i \right)$ represents the expected return that the policy $\pi(\cdot|s, \delta_i)$ can achieve in the $i$-th environment, it represents the highest possible state-action value for that environment. Therefore:
\begin{align}
\boldsymbol{\omega}^{\top}\textbf{Q}(s, a', \boldsymbol{\omega}) \leq \boldsymbol{\omega}^{\top}\textbf{Q}(s, a', \boldsymbol{\omega}^{*}) < \boldsymbol{\omega}^{\top}\textbf{Q}(s, a', \delta) = \boldsymbol{\omega}^{\top}\textbf{z}^{*}
\end{align}
where $\textbf{Q}(s, a', \delta) = [Q_{1}(s, a', \delta_{1}), \cdots, Q_{n}(s, a', \delta_{n})]$.

 With this, instead of minimizing $\norm{\boldsymbol{\omega}^{\top}\left( \textbf{Q}(s, a', \boldsymbol{\omega}) - \textbf{Q}(s, a', \boldsymbol{\omega}^{*}) \right)}$, the utopia-based MDRL (in short uMDRL) minimizes $\norm{\boldsymbol{\omega}^{\top}\left( \textbf{Q}(s, a', \boldsymbol{\omega}) - \textbf{Q}(s, a', \delta) \right)}$. We shall see that this utopia-based approach typically simplifies the input to the critic from $\textbf{Q}(s, a, \boldsymbol{\omega})$ to $\textbf{Q}(s, a)$.

\subsection{Qualitative Analysis of the proposed algorithms}

\subsubsection{\textbf{The influence of the linear utility function:}}

 Due to the scalarization function $f_{\varpi}(\cdot) = \mathbb{E}_{\kappa\sim\varpi}\left[ \cdot \right]$, the optimal policy $\pi^{*}_{\varpi}$ corresponding to an uncertainty $\varpi$ is not always assured to be the best policy for each individual environment $\kappa\sim\varpi$. For example, consider the discrete case with two environments $\mathcal{U}=\{\kappa_1, \kappa_2\}$ and a discrete uncertainty distribution $\varpi(\kappa\in\mathcal{U})=\boldsymbol{\omega}=[0.5, 0.5]$. Then, for two policies $\pi$ and $\pi'$ with Q-vector-values $\textbf{Q}^{\pi} = [5., 2.]$ and $\textbf{Q}^{\pi'} = [3., 3.]$, we see that $f_{\varpi}(\textbf{Q}^{\pi}) = 3.5 \geq f_{\varpi}(\textbf{Q}^{\pi'}) = 3$. Hence, although $\pi$ is the most optimal policy under the CCS, $\pi'$ is still a better policy on the second environment.
 
 Note that this property of the linear utility CCS makes it irrelevant to compare CCS-based policies on single environments. Indeed, the efficiency of such policies can only be evaluated over the entire support of the uncertainty distribution $\varpi$. This also holds true for the DR policies, which is obviously not the most optimal on the nominal environment, but is optimal with respect to the scalarization $f_{\varpi}(\cdot)$ over the entire randomized parameters.

\subsubsection{\textbf{Theoretical classification of the algorithms:}}

 In the sections above, we have considered three algorithms based on the PMOMDP and the analogy between MDRL and MORL as potential candidates for solving the CCS problem and therefore solving the problem of efficiently learning uncertainty-aware policies in MDRL. 

 Although each of the three types of algorithms possesses its own characteristics, we can theoretically expect a specific hierarchy between them based on MORL. That is, for a given uncertainty level $\varpi$, we expect to see the following dynamic:
\begin{align}
\begin{split}
&\mathrm{DR}_{\varpi\subset\varpi'} \leq \mathrm{cMDRL}_{\varpi} \leq \mathrm{eMDRL}_{\varpi} \\
&\mathrm{eMDRL}_{\varpi} \leq \mathrm{uMDRL}_{\varpi} \leq \mathrm{DR}_{\varpi}
\end{split}
\end{align}
where the notation $\mathrm{DR}_{\varpi\subset\varpi'}$ describes a DR algorithm trained with a larger fixed prior uncertainty $\varpi'$ than the uncertainty $\varpi$ and $\cdot \mathrm{MDRL}_{\varpi}$ is the performance of the uncertainty-aware policy $\pi: \mathcal{S}\times\Omega\to\mathcal{A}$ generated from the $\cdot \mathrm{MDRL}$ algorithm and evaluated on $\varpi\in\Omega$.
 If the true uncertainty level is explicitly given and fixed, then the DR algorithms are sufficient. However, since in practice the true level of uncertainty is typically unknown and can vary from one instance of a system to another, the uncertainty-aware algorithms $\{$c,e,u$\}$MDRL are needed.
 
 In the experiments below, we design experiments based on the CCS in order to verify if this hierarchy holds within the MDRL setting and in order to verify which algorithm is the best in practice for generating uncertainty-aware policies. For this purpose, we start by integrating the ideas developed above to a specific value-base RL algorithm.

\section{Application to Soft Actor-Critic} \label{sec:mdsac}

 The previous equations are valid for and can be integrated to arbitrary RL algorithms which are based on value functions. In order to evaluate and compare their performances, we shall however need a specific RL algorithm. In this work, we choose to apply them to the Soft-Actor Critic algorithm.

\subsection{Soft Actor-Critic}

 Soft-Actor-Critic (SAC by~\cite{haarnoja2018soft}) is a maximum entropy online RL algorithm that learns a soft Q-value function $Q(s, a)$ by repeatedly applying the Bellman evaluation operator $\mathcal{T}^{\pi}$, for a fixed policy $\pi$, defined by:
\begin{align}
\begin{split}
\mathcal{T}^{\pi}Q(s_{t}, a_{t}) &:= r(s_{t}, a_{t}) \\
&+ \gamma \mathbb{E}_{s_{t+1}\sim\mathcal{P}(s_{t+1}|s_{t}, a_{t}), a_{t+1}\sim\pi} \left[ Q(s_{t+1}, a_{t+1}) \right. \\
&\left. - \alpha \ln \pi(a_{t+1}|s_{t+1}) \right]
\end{split}
\label{eq:sac_BellmanOpe}
\end{align}
where $\alpha$ is a scalar called the ``temperature parameter" that controls the relative importance of the entropy term against the return and thus determines the stochasticity of the optimal policy. For a given Q-function $Q(s, a)$, the optimal policy is defined as
\begin{align}
\pi^{*}(a|s) = \frac{\exp\left(\frac{1}{\alpha} Q(s, a) \right)}{\int \exp\left(\frac{1}{\alpha} Q(s, a) \right) da} \propto \exp\left(\frac{1}{\alpha} Q(s, a) \right)
\label{eq:sac_optPolicy}
\end{align}

 The authors of the SAC algorithm were able to show (see theorem 1 in~\cite{haarnoja2018soft}) that the repeated application of $\mathcal{T}^{\pi}$ from equation~\eqref{eq:sac_BellmanOpe} (Soft Policy Evaluation) and the constant update of the current policy towards the optimal policy of equation~\eqref{eq:sac_optPolicy} (Soft Policy Improvement) eventually converges to a policy $\pi^{*}$ such that $Q^{\pi^{*}}(s, a) \geq Q^{\pi}(s, a)$, $\forall \pi\in\Pi$ and $\forall (s, a) \in \mathcal{S}\times\mathcal{A}$ with $|\mathcal{A}|<\infty$.

\subsection{Conditioned Multi-Domain Soft Actor-Critic}

 As stated before, the scalarization of the MDRL problem allows for the use of a vanilla RL algorithm as a solver with minor modifications. Indeed, the optimal policy, for a given uncertainty $\varpi$ corresponding to the single environment's equation~\eqref{eq:sac_optPolicy} is readily obtained with the scalarized state-action value $Q(s, a) = \mathbb{E}_{\kappa\sim\varpi}\left[ Q(s, a, \kappa) \right]$ as:
\begin{align}
\pi^{*}(a|s) = \frac{\exp\left(\frac{1}{\alpha} \mathbb{E}_{\kappa\sim\varpi}\left[ Q(s, a, \kappa) \right] \right)}{Z(s)}
\label{eq:drsac_optPolicy}
\end{align}
If $\varpi$ is allowed to change dynamically (as in cMDRL, eMDRL and uMDRL), then the adaptive optimal policy can be rewritten as a function of $Q(s, a, \kappa, \varpi)$:
\begin{align}
\pi^{*}(a|s, \varpi) = \frac{\exp\left(\frac{1}{\alpha} \mathbb{E}_{\kappa\sim\varpi}\left[ Q(s, a, \kappa, \varpi) \right] \right)}{Z(s, \varpi)}
\label{eq:mdsac_optPolicy}
\end{align}

Hence, the policy improvement step corresponding to the CCS is given by the Kullback-Leibler divergence optimization, $\forall (s, \varpi)\in\mathcal{S}\times\Omega$:
\begin{align}
\begin{split}
\pi_{\mathrm{new}} &= \argmin{\pi}{D_{\mathrm{KL}}} \bigg( \pi(a|s, \varpi)) \| \\
&\frac{\exp\left(\frac{1}{\alpha} \mathbb{E}_{\kappa\sim\varpi}\left[ Q^{\pi_{\mathrm{old}}}(s, a, \kappa, \varpi) \right] \right)}{Z^{\pi_{\mathrm{old}}}(s, \varpi)} \bigg)
\end{split}
\label{eq:mdsac_pol_imp_cont}
\end{align}
\Mycomment{
In the discrete case, this simplifies to:
\begin{align}
\begin{split}
\pi_{\mathrm{new}} &= \argmin{\pi}{D_{\mathrm{KL}}} \bigg( \pi(a|s_t, \boldsymbol{\omega}) \| \\
& \vphantom{\frac{\exp(\frac{1}{\alpha} Q_{i}^{\pi_{\mathrm{old}}}(s_t, \cdot, \boldsymbol{\omega}))}{Z_{i}^{\pi_{\mathrm{old}}}(s_t, \boldsymbol{\omega})}} \frac{ \prod\limits_{i} \exp(\frac{\omega_{i}}{\alpha} Q_{i}^{\pi_{\mathrm{old}}}(s_t, a, \boldsymbol{\omega})) }{Z^{\pi_{\mathrm{old}}}(s_t, \boldsymbol{\omega})} \bigg)
\end{split}
\label{eq:mdsac_pol_imp_disc}
\end{align}
}

 If we then take the Bellman operator to be the same as in equation~\eqref{eq:sac_BellmanOpe}, but applied to the state-action value function $Q(s, a, \kappa, \varpi)$, then we have defined the conditioned multi-domain version of the SAC algorithm, which we call \emph{cMDSAC}.

\subsection{Envelope Multi-Domain Soft Actor-Critic}

 By its very name, cMDSAC is straightforward to derive from SAC by simply making the conditioning explicit. On the contrary, the envelope version of the multi-domain Soft Actor-Critic algorithm (named \emph{eMDSAC} from now on) is not as straightforward.
 As described above, the particularity of the envelope algorithm comes from the use of the CCS optimality filter of equation~\eqref{eq:optim_filter}. We must therefore derive the soft maximum-entropy algorithm by using it from the start.
 
 For a given uncertainty set captured by $\varpi$, let the augmented reward function be given by:
\begin{align*}
\tilde{r}(s, a, \kappa) &= r(s, a, \kappa) + \alpha \mathcal{H}(\pi(\cdot|s, \varpi))
\end{align*}
where $\pi(\cdot|s, \varpi)$ ($= \pi_{\varpi}(\cdot|s)$) is the universal policy network such that $a\sim\pi(\cdot|s, \varpi)$ and $\mathcal{H}(\cdot)$ is the entropy. Assuming this reward function, the CCS optimal state value function satisfies the following equation:
\begin{align}
\begin{split}
V^{*}&(s, \varpi) = \underset{\pi_{\varpi}}{\max}\; \mathbb{E}_{a\sim\pi_{\varpi}}\bigg[ \underset{\varpi'}{\max}\; \mathbb{E}_{\kappa\sim\varpi}\big[ r(s, a, \kappa) - \alpha \ln \pi_{\varpi}(a|s) \\
&+ \gamma \mathbb{E}_{s'\sim\mathcal{P}_{\kappa}}\left[ V(s', \kappa, \varpi') \right] \big] \bigg]
\end{split} \nonumber \\
\begin{split}
&= \underset{\pi_{\varpi}}{\max}\; \mathbb{E}_{a\sim\pi_{\varpi}}\bigg[ \mathbb{E}_{\kappa\sim\varpi}\big[ Q(s, a, \kappa, \varpi^{*}) \big] - \alpha \ln \pi_{\varpi}(a|s) \bigg]
\end{split}
\label{eq:emd_OptimValueFunc}
\end{align}
where we have employed the Bellman optimality equation of equation~\eqref{eq:emdrl} in the first equality, and employed equation~\eqref{eq:optim_filter} for the deterministic optimality filter $\varpi^{*} = \argmax{\varpi'}{\mathbb{E}_{\kappa\sim\varpi}\big[ \mathbb{E}_{s'}\left[ V(s', \kappa, \varpi') \right] \big]}$.

The right-hand-side of equation~\eqref{eq:emd_OptimValueFunc} is a constrained optimization problem with the constraint given by $\int \pi(a | s, \varpi) da = 1$. We can therefore solve it using the Lagrange Multipliers method (with $\lambda$ the Lagrange multiplier) to obtain:
\begin{align}
\begin{split}
\pi^{*}&(a | s, \varpi) = \exp\bigg( \frac{1}{\alpha} \underset{\varpi'}{\max}\; \mathbb{E}_{\kappa\sim\varpi}\big[ r(s, a, \kappa) \\
& + \gamma \mathbb{E}_{s'}\left[ V(s', \kappa, \varpi') \right] \big] \bigg) \exp\left( \frac{1}{\alpha} \lambda \right)
\end{split} \\
\begin{split}
\pi^{*}&(a | s, \varpi) \propto \exp\bigg( \frac{1}{\alpha} \mathbb{E}_{\kappa\sim\varpi}\big[ Q(s, a, \kappa, \varpi^{*}) \big] \bigg)
\end{split}
\end{align}

 Hence, the policy improvement in the envelope algorithm is given by:
\begin{align*}
\begin{split}
\pi_{\mathrm{new}} &= \argmin{\pi}{D_{\mathrm{KL}}} \bigg( \pi(a|s, \varpi)) \| \\
&\frac{\exp\big( \frac{1}{\alpha} \mathbb{E}_{\kappa\sim\varpi}\big[ Q^{\pi_{\mathrm{old}}}(s, a, \kappa, \varpi^{*}) \big] \big)}{Z^{\pi_{\mathrm{old}}}(s, \varpi)} \bigg)
\end{split}
\end{align*}
 with the policy evaluation as:
\begin{align*}
Q^{\pi_{\mathrm{old}}}(s, a, \kappa, \varpi^{*}) &= \underset{\varpi'}{\max}\; \mathbb{E}_{\kappa\sim\varpi}\big[ r(s, a, \kappa)  \\
& + \gamma \mathbb{E}_{s'}\left[ V(s', \kappa, \varpi') \right] \big]
\end{align*}
 As stated before, this objective can be unstable due to the max operator being applied, in SAC, on the target network and we found that in practice, it tends to overestimate the value function and cause the algorithm to fail to learn effectively.
 Since our objective is to align the policy corresponding to $\varpi$ with the policy corresponding to $\varpi^{*}$, we can assure this by updating the critic corresponding to $\varpi$ with that of the critic corresponding to $\varpi^{*}$. This means replacing $Q^{\pi_{\mathrm{old}}}(s, a, \kappa, \varpi^{*})$ by $Q^{\pi_{\mathrm{old}}}(s, a, \kappa, \varpi)$ in both policy evaluation and improvement steps:
\begin{align}
\begin{split}
&\pi_{\mathrm{new}} = \argmin{\pi}{D_{\mathrm{KL}}} \bigg( \pi(a|s, \varpi)) \| \\
&\;\;\;\;\;\frac{\exp\big( \frac{1}{\alpha} \mathbb{E}_{\kappa\sim\varpi}\big[ Q^{\pi_{\mathrm{old}}}(s, a, \kappa, \varpi) \big] \big)}{Z^{\pi_{\mathrm{old}}}(s, \varpi)} \bigg)
\end{split}
\label{eq:emdsac_pol_imp_cont} \\
\begin{split}
&Q^{\pi_{\mathrm{old}}}(s, a, \kappa, \varpi) = \mathbb{E}_{\kappa\sim\varpi}\big[ r(s, a, \kappa) \\
&\;\;\; + \gamma \mathbb{E}_{s'}\left[ \mathbb{E}_{a'\sim\pi(\cdot|s', \chi^{*}(s, \varpi))}\left[ Q(s', a', \kappa, \chi^{*}(s', \varpi)) \right. \right. \\
&\;\;\;\left. \left. - \alpha \ln \pi(a'|s', \chi^{*}(s', \varpi)) \right] \right] \big]
\end{split}
\end{align}

where
\begin{align}
\begin{split}
\chi^{*}(s, \varpi) = \varpi^{*} &= \argmax{\varpi'}{ \mathbb{E}_{a\sim\pi_{\varpi'}, \kappa\sim\varpi}\left[ Q(s, a, \kappa, \varpi') \right. \\
&\left. - \alpha \ln \pi(a|s, \varpi') \right] }
\end{split}
\end{align}

$\chi^{*}(s, a, \varpi)$ is the optimality filter which solves explicitly for the CCS in order to find the best policy indexed by $\varpi^{*}$, given the uncertainty $\varpi$.

\subsection{Utopia-based Multi-Domain Soft Actor-Critic}

 As discussed previously, the utopia-based approach to the multi-domain soft actor-critic algorithm is obtained by replacing the CCS optimality filter solution of the envelope algorithm by an utopian point, given, in this paper, by the expected value over the uncertainty $\varpi$ of the state-action value function $Q(s, a, \kappa, \delta_{\kappa})$ generated from environment $\kappa$ by taking action $a$ in state $s$ and then following the policy $\pi_{\delta_{\kappa}}$. Hence, the utopia-based optimal state value function satisfies the following equation:
\begin{align}
\begin{split}
V^{*}&(s, \varpi) = \underset{\pi_{\varpi}}{\max}\; \mathbb{E}_{a\sim\pi_{\varpi}}\bigg[ \mathbb{E}_{\kappa\sim\varpi}\big[ r(s, a, \kappa) - \alpha \ln \pi_{\varpi}(a|s) \\
&+ \gamma \mathbb{E}_{s'\sim\mathcal{P}_{\kappa}}\left[ V(s', \kappa, \delta_{\kappa}) \right] \big] \bigg]
\end{split}
\label{eq:umd_OptimValueFunc}
\end{align}

 By following the Lagrange multipliers method, we can derive the optimal policy as:
\begin{align}
\begin{split}
\pi^{*}(a | s, \varpi) &\propto \exp\bigg( \frac{1}{\alpha} \mathbb{E}_{\kappa\sim\varpi}\big[ r(s, a, \kappa) + \gamma \mathbb{E}_{s'}\left[ V(s', \kappa, \delta_{\kappa}) \right] \big] \bigg)
\end{split} \\
&\propto \exp\bigg( \frac{1}{\alpha} \mathbb{E}_{\kappa\sim\varpi}\big[ Q(s, a, \kappa) \big] \bigg)
\end{align}
where, because the tuple $(\kappa, \delta_{\kappa})$ does not carry more information than simply $\kappa$, we have made the simplification $Q(s, a, \kappa, \delta_{\kappa}) = Q(s, a, \kappa)$. The policy improvement step of the utopia-based algorithm is therefore given by:
\begin{align}
\begin{split}
\pi_{\mathrm{new}} &= \argmin{\pi}{D_{\mathrm{KL}}} \bigg( \pi(a|s, \varpi)) \| \\
&\frac{\exp\big( \frac{1}{\alpha} \mathbb{E}_{\kappa\sim\varpi}\big[ Q^{\pi_{\mathrm{old}}}( s, a, \kappa) \big] \big)}{Z^{\pi_{\mathrm{old}}}(s, \varpi)} \bigg)
\end{split}
\label{eq:umdsac_pol_imp_cont}
\end{align}
with the policy evaluation applied only with $Q(s, a, \kappa)$ (instead of $Q(s, a, \kappa, \varpi)$), i.e.:
\begin{align}
\begin{split}
&\mathcal{T}^{\pi}Q(s, a, \kappa) := r(s, a, \kappa) \\
&+ \gamma \mathbb{E}_{s'\sim\mathcal{P}_{\kappa}(s'|s, a), a'\sim\pi_{\delta_{\kappa}}} \left[ Q(s', a', \kappa) - \alpha \ln \pi(a'|s', \delta_{\kappa}) \right]
\end{split}
\label{eq:umdsac_pol_eval}
\end{align}

 These are the same update rules that were derived in our previous work (described in~\cite{ilboudo2023domains}) where it was wrongly named as \emph{sMDSAC} (for scalarized MDSAC). In the present paper, we investigate another version of this update rule as described below.

 Indeed, we start by noticing that during training, $\pi(a'|s', \delta_{\kappa})$ is not guaranteed to be the most optimal policy for the environment $\kappa$, which means that the utopian point may not steer the policy search towards a better value function. This can be easily solved by reintroducing an optimality solver similar to the one found in the envelope algorithm:
\begin{align}
\begin{split}
\chi^{*}(s, \kappa) = \varpi^{*} &= \argmax{\varpi'}{ \mathbb{E}_{a\sim\pi_{\varpi'}}\left[ Q(s, a, \kappa) \right. \\
&\left. - \alpha \ln \pi(a|s, \varpi') \right] }
\end{split}
\end{align}

Here, in contrast to the solver of eMDSAC, $\chi^{*}(s, \kappa)$ solves for the best policy  indexed by $\varpi^{*}$ for the specific environment $\kappa$. The policy evaluation is then:
\begin{align}
\begin{split}
&\mathcal{T}^{\pi}Q(s, a, \kappa) := r(s, a, \kappa) \\
&+ \gamma \mathbb{E}_{s'\sim\mathcal{P}_{\kappa}(s'|s, a), a'\sim\pi(\cdot|s', \chi^{*}(s', \kappa))} \left[ Q(s', a', \kappa) \right. \\
&\left. - \alpha \ln \pi(a'|s', \chi^{*}(s', \kappa)) \right]
\end{split}
\label{eq:umdsacv2_pol_eval}
\end{align}

 Both versions of the utopia algorithm are evaluated in the experiments under the names, \emph{uMDSAC-v1} and \emph{uMDSAC-v2} respectively.

\section{Online System Identifier}

 An uncertainty aware policy's performance is also correlated to the performance of the system identifier. Although \cite{xie2022robust} proposed an OSI for their uncertainty-aware algorithm (i.e. SIRSA), they overlooked an important issue pertaining to the identification of dynamics parameters, i.e. the fact that there may be redundant parameter representations causing the presence of multiple distinct parameters within the uncertainty set that produce the same dynamics. To avoid the effect of these redundant parameter representations, \cite{ding2021not} proposed to learn an embedding of the parameters and then employ a Bayesian optimization to identify the embedded representation.

 In this work, we seek for an Online System Identifier that can be trained simultaneously with the uncertainty-aware policy and which is able to perform its prediction within the original parameter space without the need for an intermediary embedding space, while still avoiding the effect of redundant parameter representations on the uncertainty prediction. For this purpose, we reformulate the OSI under a variational inference setting as described in the next paragraphs.

\subsection{System identification as variational inference}

 Let $h_{t} = (s_0, a_0, \cdots, s_{t-1}, a_{t-1}, s_{t})$ be the full history up to time step $t$ and let $\kappa$ be the simulation randomized parameters. Then, we can define:
\begin{itemize}
\item the OSI as: $\varpi_{t+1}(\kappa) = p(\kappa | h_{t}, a_{t}, s_{t+1})$ and
\item the Dynamics model as: $D(s_{t+1}) = p(s_{t+1} | h_{t}, a_{t}, \kappa)$.
\end{itemize}
 To reformulate the problem as a variational inference objective, we make the following identification:
\begin{itemize}
\item Conditional encoder (i.e. $p(z|x, c)$) $\equiv$ OSI with $\kappa$ the latent variable, $x$ the next state $s_{t+1}$ and the context $c$ the history and action $(h_t, a_t)$.
\item Conditional decoder (i.e. $p(x|z, c)$) $\equiv$ Dynamics model.
\end{itemize}

 We also write the target optimal OSI as:
\begin{align*}
q(\kappa | h_{t}, a_{t}, s_{t+1}) = \frac{p(s_{t+1} | h_{t}, a_{t}, \kappa) \varpi_{t}(\kappa)}{p(s_{t+1} | h_{t}, a_{t})} = \frac{D(s_{t+1})\varpi_{t}(\kappa)}{Z}
\end{align*}
where we set the prior $p(\kappa) = \varpi_{t}(\kappa) = p(\kappa | h_{t-1}, a_{t-1}, s_{t})$. Then, the objective is:
\begin{align*}
(D^{*}, \varpi^{*}_{t+1}) &= \argmin{D, \varpi_{t+1}}{\mathrm{KL}\left[ \varpi_{t+1}(\kappa) || q(\kappa | h_{t}, a_{t}, s_{t+1}) \right]} \\
\begin{split}
&= \argmin{D, \varpi_{t+1}}{ \mathbb{E}_{\kappa\sim\varpi_{t+1}}\left[ \ln \varpi_{t+1}(\kappa) \right] \\
&\;\;- \mathbb{E}_{\kappa\sim\varpi_{t+1}}\left[ \ln \frac{D(s_{t+1})\varpi_{t}(\kappa)}{Z} \right] }
\end{split} \\
\begin{split}
&= \argmin{D, \varpi_{t+1}}{ \mathbb{E}_{\kappa\sim\varpi_{t+1}}\left[ -\ln D \right] \\
&\;\;+ \mathrm{KL}\left[ \varpi_{t+1} || \varpi_{t} \right] } + \ln Z
\end{split}
\end{align*}
Since the normalization constant $Z$ is independent of both $D$ and $\varpi$, the Variational Auto-Encoder (VAE) objective corresponding to the variational inference of the system identification is given by:
\begin{align}
(D^{*}, \varpi^{*}_{t+1}) = \argmin{D, \varpi_{t+1}}{ \mathbb{E}_{\kappa\sim\varpi_{t+1}}\left[ -\ln D \right] + \mathrm{KL}\left[ \varpi_{t+1} || \varpi_{t} \right] }
\label{eq:vae}
\end{align}

\subsection{Practical algorithm}

 In practice, using the full history is hard, as the input dimension of the neural network grows fast. Hence, the partial history used by \cite{yu2017uposi} for their OSI. Furthermore, because of the reliance on history data, it is not easy to train the OSI simultaneously with the actor using the replay buffer. Therefore, the definition given above is modified in order to only rely on the immediate transition $s-a-s$ information available at each step.

For that, we simply replace the past full history information by the previously estimated parameter uncertainty in order to simplify the OSI and we use the Markovian feature in order to simplify the Dynamics model:
\begin{itemize}
\item The OSI is given by: $\varpi_{t+1}(\kappa) = p(\kappa | h_{t}, a_{t}, s_{t+1})$
\end{itemize}
\begin{align*}
p(\kappa | h_{t}, a_{t}, s_{t+1}) &= p(\kappa | s_0, a_0, \cdots, s_{t-1}, a_{t-1}, s_{t}, a_{t}, s_{t+1})\\
&= p(\kappa | \varpi_{t}, s_{t}, a_{t}, s_{t+1}) = \varpi_{t+1}(\kappa)
\end{align*}
\begin{itemize}
\item As for the Dynamics model, it is given by: $D(s_{t+1}) = p(s_{t+1} | h_{t}, a_{t}, \kappa)$.
\end{itemize}
\begin{align*}
p(s_{t+1} | h_{t}, a_{t}, \kappa) &= p(s_{t+1} | s_0, a_0, \cdots, s_{t-1}, a_{t-1}, s_{t}, a_{t}, \kappa) \\
&= p(s_{t+1} | s_{t}, a_{t}, \kappa) = D(s_{t+1})
\end{align*}

 In our implementation, following \cite{xie2022robust}, we employ a neural network ensemble for the OSI, where each network $\mathrm{NN}^{(i)}$ takes in the input $(\varpi_{t}, s_{t}, a_{t}, s_{t+1})$ and predicts the parameter $\kappa^{(i)}_{t}$. The uncertainty $\varpi_{t}(\kappa)$ is then given by taking the mean and standard deviation of all the predictions made by the ensemble, i.e.:
\begin{align}
\varpi_{t+1} &= [\mu_{t+1}, \sigma_{t+1}] \\
\begin{split}
&= \bigg[ \underset{i=1,\cdots, k}{\mathrm{mean}}\left( \mathrm{NN}^{(i)}(s_{t}, a_{t}, s_{t+1}, \varpi_{t}) \right),\\
&\;\;\;\;\;\;\underset{i=1,\cdots, k}{\mathrm{std}}\left( \mathrm{NN}^{(i)}(s_{t}, a_{t}, s_{t+1}, \varpi_{t}) \right) \bigg]
\end{split}
\end{align}
where $k$ is the ensemble size (set to $4$ in the experiments below).

\section{Comparative evaluations} \label{sec:evaluations}

 In this section, we evaluate the performance of the proposed algorithms first in simulation and then on a real robot. For each environment and each method, we train 8 models with different random seeds. For all MDSAC algorithms, in order to match the uniform DRSAC (Domain Randomization SAC), the space $\Omega$ is chosen to be the space of continuous uniform distributions with support lying on and within the boundaries of the full domain. To represent one distribution $\varpi\in\Omega$ as an input to the MDSAC agents, we use the mean and standard deviation such that $\varpi = [\mu, \sigma]$ where $\mu, \sigma \in \mathbb{R}^d$, $d$ being the dimension of the randomized parameters.

 Furthermore, to compute the expectation necessary for the scalarization of the proposed algorithms (cMD, eMD and uMD-SAC), we make use of the unscented transform (UT - see \cite{uhlmann1995dynamic}) applied to the uniform distribution so as to discretize $\varpi$. Specifically, we numerically solve for the sigma-points (using a simple mirror descent algorithm) in order to match the first $N$ centered moments of the uniform distribution, where $N = \max{(10, d)}$ (see Appendix~\ref{apdx:UT_details} for the obtained sigma-points for the 2 dimensional case).

 The neural networks for the critic, actor and CCS-optimality-filter solver are made of two hidden layers, each with 256 neurons for all tasks. The outputs of the actor networks are the mean and standard deviation of a Gaussian distribution (only the mean is used during the evaluation, as deterministic policy) and the optimization algorithm used for both actor and critic training is the t-Adam robust optimizer (i.e. the t-momentum version of the Adam algorithm by \cite{ilboudo2020robust}).

 We evaluate all four of our proposed algorithms (cMDSAC, eMDSAC, uMDSAC-v1 and uMDSAC-v2) along with DRSAC and SIRSA (by \cite{xie2022robust}, focusing on the SystemID version of their algorithm which uses the linear utility function of interest in this paper, instead of the worst-case version). In addition, we also test a variant of the utopia algorithm based on SIRSA and named uMD-SIRSA, where the solver is replaced by the predictions from the OSI. Table~\ref{tab:algo_summary} gives the summary of the algorithms evaluated in this section\footnote{The implementations of all algorithms can be found on https://github.com/Mahoumaru/MDRL}. In all evaluations, the SIRSA algorithm is always evaluated using its OSI for parameter uncertainty reduction. For all results, we report the interquartile mean and the interquartile range (in parentheses) over all trials and random seeds (i.e. over a total number of data points $n = n_{\mathrm{trials}} \times n_{\mathrm{seeds}}$).

\subsection{Sim-to-sim transfer}

\subsubsection{Setup:}

 The simulation environments employed are the Hopper, Walker2D and DClawTurnFixed (see~\cite{ahn2020robel}) in the Mujoco~(\cite{todorov2012mujoco}) simulator. 
 For each environment, we consider both a 2-dimensional and a high dimensional domain randomization case.
\begin{itemize}
\item[•] In the 2D case: For the Hopper and Walker2d tasks, we have randomized the mass of the bodies and the joint's armature. However, rather than randomizing each component of the mass and armature independently, we instead consider the relative mass $\textbf{m}_{\mathrm{rel}} = \alpha_{m} \textbf{m}_{\mathrm{nominal}}$ and relative armature $\textbf{arm}_{\mathrm{rel}} = \alpha_{j} \textbf{arm}_{\mathrm{nominal}}$ and randomized the scalar coefficients $\alpha_{m}$ and $\alpha_{j}$ each in the range $[0.5, 2]$. Similarly, for the D'Claw environment, we randomize the proportional gain of the actuators using a scalar coefficient $\alpha_{g}$ within range $[0.2, 1.8]$, and the joints' damping using a coefficient $\alpha_{d} \in [0.05, 1.95]$.
\item[•] In the high-dimensional case: For the Hopper and Walker2d tasks, we randomized the body mass, the joints armature and damping, the gravity coefficient and the simulation timestep for a total of 12 dimensions for Hopper and 21 dimensions for Walker. For the D'Claw environment, we randomize the proportional gain of the actuators, the joints damping, the simulation timestep and the kinematic position of the base of each fingers for a total of 22 dimensions.
\end{itemize}

%
 As mentioned above in the qualitative analysis, for a given uncertainty, a comparison of the performance of the different algorithms should be performed over the CCS score. In the evaluation, we use a different set of sigma-points than those used during training in order to compute the linear-utility CCS score (the highest score is the best).
\begin{table*}[t]
  \captionsetup{width=\textwidth, justification=centering}
  \caption{DClawTurnFixed-v0 2D Sim-to-Sim evaluations CCS scores. The highest score is highlighted in red while the second and third highest scores are highlighted in green and blue respectively (\textcolor{red}{Red}$>$\textcolor{mydarkgreen}{Green}$>$\textcolor{blue}{Blue}). Hence, the column with the most red is the best.}
  \label{tab:dclaw_2dim_ccs}
  \centering
  \begin{adjustbox}{width=1\textwidth}
  \begin{tabular}{p{4.5cm}|c|c|c|c|c|c|c}
		\hline\hline
		$\varpi$ & DRSAC & SIRSA & cMDSAC (Ours) & eMDSAC (Ours) & uMD-SIRSA (Ours) & uMDSAC-v1 (Ours) & uMDSAC-v2 (Ours) \\
		\hline
		$\varpi_{1} = {[1.0, 1.0, 0.46, 0.55]}$ & $412.5745$  & $-382.8779$  & $737.9066$  & $688.9708$  & $\textcolor{blue}{\textbf{744.0013}}$  & $\textcolor{red}{\textbf{780.2303}}$  & $\textcolor{mydarkgreen}{\textbf{765.4302}}$  \\ & $(± 108.6551)$  & $(± 454.9917)$  & $(± 181.6842)$  & $(± 60.4077)$  & $(± 58.5287)$  & $(± 49.1149)$  & $(± 38.1599)$  \\

		\hline
		$\varpi_{2} = {[0.46, 1.82, 0.14, 0.06]}$ & $-653.0271$  & $-605.41$  & $\textcolor{blue}{\textbf{-563.2355}}$  & $-588.1312$  & $-598.1766$  & $\textcolor{red}{\textbf{-552.2157}}$  & $\textcolor{mydarkgreen}{\textbf{-563.1066}}$  \\ & $(± 26.3997)$  & $(± 47.3944)$  & $(± 84.1147)$  & $(± 41.7332)$  & $(± 30.8033)$  & $(± 21.2865)$  & $(± 68.2546)$  \\

		\hline
		$\varpi_{3} = {[1.58, 1.27, 0.12, 0.37]}$ & $706.8552$  & $27.5476$  & $\textcolor{blue}{\textbf{1133.5153}}$  & $1044.8462$  & $1131.4172$  & $\textcolor{mydarkgreen}{\textbf{1147.6322}}$  & $\textcolor{red}{\textbf{1156.5735}}$  \\ & $(± 84.7283)$  & $(± 780.7982)$  & $(± 72.8041)$  & $(± 95.687)$  & $(± 52.02)$  & $(± 45.5395)$  & $(± 29.2382)$  \\

		\hline
		$\varpi_{4} = {[1.55, 1.05, 0.12, 0.39]}$ & $793.2925$  & $-160.4021$  & $\textcolor{blue}{\textbf{1227.2184}}$  & $1126.4471$  & $1199.9169$  & $\textcolor{mydarkgreen}{\textbf{1227.2561}}$  & $\textcolor{red}{\textbf{1238.5501}}$  \\ & $(± 131.5844)$  & $(± 663.5729)$  & $(± 74.9352)$  & $(± 89.6816)$  & $(± 46.1242)$  & $(± 41.9951)$  & $(± 35.5167)$  \\

		\hline
		$\varpi_{5} = {[1.38, 1.31, 0.16, 0.14]}$ & $506.2797$  & $-32.5911$  & $\textcolor{blue}{\textbf{1029.8591}}$  & $938.8148$  & $1021.1108$  & $\textcolor{mydarkgreen}{\textbf{1042.2673}}$  & $\textcolor{red}{\textbf{1066.6963}}$  \\ & $(± 153.9456)$  & $(± 732.9099)$  & $(± 83.2279)$  & $(± 108.7355)$  & $(± 56.0276)$  & $(± 55.2997)$  & $(± 41.6255)$  \\

		\hline
		$\varpi_{6} = {[1.11, 0.96, 0.05, 0.03]}$ & $627.1696$  & $35.1069$  & $\textcolor{mydarkgreen}{\textbf{1141.7047}}$  & $1045.9166$  & $1124.9842$  & $\textcolor{blue}{\textbf{1128.8596}}$  & $\textcolor{red}{\textbf{1157.9025}}$  \\ & $(± 168.9836)$  & $(± 732.1397)$  & $(± 68.6193)$  & $(± 88.6227)$  & $(± 47.3608)$  & $(± 48.0837)$  & $(± 45.6069)$  \\

		\hline
		$\varpi_{7} = {[0.87, 1.45, 0.28, 0.1]}$ & $-174.9208$  & $-188.0297$  & $\textcolor{blue}{\textbf{359.6933}}$  & $264.8421$  & $336.1526$  & $\textcolor{mydarkgreen}{\textbf{390.3469}}$  & $\textcolor{red}{\textbf{406.5549}}$  \\ & $(± 99.9844)$  & $(± 453.0017)$  & $(± 133.253)$  & $(± 136.6209)$  & $(± 89.4407)$  & $(± 56.7042)$  & $(± 29.9702)$  \\

		\hline
		$\varpi_{8} = {[1.66, 0.5, 0.06, 0.14]}$ & $1170.2303$  & $-529.9982$  & $\textcolor{mydarkgreen}{\textbf{1440.0005}}$  & $1369.2283$  & $\textcolor{red}{\textbf{1445.9927}}$  & $\textcolor{blue}{\textbf{1426.344}}$  & $1422.259$  \\ & $(± 126.7956)$  & $(± 509.0167)$  & $(± 86.2901)$  & $(± 58.7093)$  & $(± 68.841)$  & $(± 97.7713)$  & $(± 95.5274)$  \\

		\hline
		$\varpi_{9} = {[0.33, 1.63, 0.06, 0.03]}$ & $-671.0422$  & $-617.8119$  & $-624.189$  & $-619.2468$  & $\textcolor{mydarkgreen}{\textbf{-615.4759}}$  & $\textcolor{red}{\textbf{-602.5084}}$  & $\textcolor{blue}{\textbf{-615.803}}$  \\ & $(± 35.9674)$  & $(± 28.5436)$  & $(± 56.4238)$  & $(± 24.5671)$  & $(± 16.0105)$  & $(± 8.936)$  & $(± 43.0569)$  \\

		\hline
		$\varpi_{10} = {[1.69, 0.36, 0.01, 0.13]}$ & $1180.6446$  & $-586.8184$  & $1286.6837$  & $1310.2872$  & $\textcolor{mydarkgreen}{\textbf{1444.3799}}$  & $\textcolor{red}{\textbf{1450.4834}}$  & $\textcolor{blue}{\textbf{1432.4596}}$  \\ & $(± 146.1964)$  & $(± 528.2031)$  & $(± 178.358)$  & $(± 138.2064)$  & $(± 55.4455)$  & $(± 82.7463)$  & $(± 58.7266)$  \\

		\hline
		$\varpi_{11} = {[1.45, 0.48, 0.08, 0.04]}$ & $1186.0367$  & $-392.3437$  & $1461.0842$  & $1422.5814$  & $\textcolor{blue}{\textbf{1472.2984}}$  & $\textcolor{red}{\textbf{1480.0586}}$  & $\textcolor{mydarkgreen}{\textbf{1475.4533}}$  \\ & $(± 142.6936)$  & $(± 709.3881)$  & $(± 70.2527)$  & $(± 63.3701)$  & $(± 37.3318)$  & $(± 45.9576)$  & $(± 60.1973)$  \\

		\hline
	\end{tabular}
	\end{adjustbox}
\end{table*}
\begin{table*}[t]
  \captionsetup{width=\textwidth, justification=centering}
  \caption{Hopper-v3 2D Sim-to-Sim evaluations CCS scores. The highest score is highlighted in red while the second and third highest scores are highlighted in green and blue respectively (\textcolor{red}{Red}$>$\textcolor{mydarkgreen}{Green}$>$\textcolor{blue}{Blue}). Hence, the column with the most red is the best.}
  \label{tab:hopper_2dim_ccs}
  \centering
  \begin{adjustbox}{width=1\textwidth}
  \begin{tabular}{p{4.5cm}|c|c|c|c|c|c|c}
		\hline\hline
		$\varpi$ & DRSAC & SIRSA & cMDSAC (Ours) & eMDSAC (Ours) & uMD-SIRSA (Ours) & uMDSAC-v1 (Ours) & uMDSAC-v2 (Ours) \\
		\hline
		$\varpi_{1} = {[1.25, 1.25, 0.43, 0.43]}$ & $1492.3732$  & $1219.4104$  & $1718.0671$  & $2263.0833$  & $\textcolor{mydarkgreen}{\textbf{2516.3093}}$  & $\textcolor{red}{\textbf{2562.1918}}$  & $\textcolor{blue}{\textbf{2418.4629}}$  \\ & $(± 276.6411)$  & $(± 715.3257)$  & $(± 514.2659)$  & $(± 246.3545)$  & $(± 307.3695)$  & $(± 487.0147)$  & $(± 437.6187)$  \\

		\hline
		$\varpi_{2} = {[0.74, 1.9, 0.13, 0.05]}$ & $1284.7206$  & $1390.6931$  & $1735.364$  & $\textcolor{blue}{\textbf{2109.9239}}$  & $1953.0556$  & $\textcolor{red}{\textbf{2629.877}}$  & $\textcolor{mydarkgreen}{\textbf{2491.6399}}$  \\ & $(± 309.0561)$  & $(± 1014.6125)$  & $(± 780.715)$  & $(± 839.3669)$  & $(± 726.34)$  & $(± 501.8131)$  & $(± 717.0078)$  \\

		\hline
		$\varpi_{3} = {[1.79, 1.46, 0.11, 0.29]}$ & $851.3529$  & $955.9403$  & $2075.1336$  & $2073.5169$  & $\textcolor{mydarkgreen}{\textbf{2315.8069}}$  & $\textcolor{blue}{\textbf{2179.2577}}$  & $\textcolor{red}{\textbf{2604.1004}}$  \\ & $(± 188.9776)$  & $(± 987.741)$  & $(± 564.7649)$  & $(± 679.7576)$  & $(± 269.3111)$  & $(± 771.4855)$  & $(± 235.3481)$  \\

		\hline
		$\varpi_{4} = {[1.76, 1.29, 0.11, 0.31]}$ & $916.7127$  & $982.0879$  & $2102.0393$  & $2298.0319$  & $\textcolor{mydarkgreen}{\textbf{2494.7063}}$  & $\textcolor{blue}{\textbf{2412.5296}}$  & $\textcolor{red}{\textbf{2653.9077}}$  \\ & $(± 113.6536)$  & $(± 1062.384)$  & $(± 618.8013)$  & $(± 546.7736)$  & $(± 227.6545)$  & $(± 676.9878)$  & $(± 261.583)$  \\

		\hline
		$\varpi_{5} = {[1.61, 1.49, 0.15, 0.11]}$ & $962.9463$  & $1174.4931$  & $\textcolor{blue}{\textbf{2493.8533}}$  & $2482.364$  & $\textcolor{mydarkgreen}{\textbf{2746.7966}}$  & $2381.8399$  & $\textcolor{red}{\textbf{2777.873}}$  \\ & $(± 206.0835)$  & $(± 1010.4909)$  & $(± 591.7436)$  & $(± 554.5615)$  & $(± 237.6881)$  & $(± 846.818)$  & $(± 192.4108)$  \\

		\hline
		$\varpi_{6} = {[1.36, 1.22, 0.05, 0.03]}$ & $1257.475$  & $1354.3132$  & $2792.7189$  & $\textcolor{blue}{\textbf{3027.5931}}$  & $\textcolor{mydarkgreen}{\textbf{3096.8693}}$  & $\textcolor{red}{\textbf{3125.1515}}$  & $2915.3662$  \\ & $(± 306.7293)$  & $(± 1065.6874)$  & $(± 610.6632)$  & $(± 401.813)$  & $(± 98.7901)$  & $(± 100.6202)$  & $(± 465.862)$  \\

		\hline
		$\varpi_{7} = {[1.13, 1.61, 0.26, 0.08]}$ & $1405.2391$  & $1922.6331$  & $2307.404$  & $2519.5467$  & $\textcolor{mydarkgreen}{\textbf{2770.1154}}$  & $\textcolor{red}{\textbf{2877.234}}$  & $\textcolor{blue}{\textbf{2764.6516}}$  \\ & $(± 416.5206)$  & $(± 879.2666)$  & $(± 752.9563)$  & $(± 488.5601)$  & $(± 243.1551)$  & $(± 315.0719)$  & $(± 363.0927)$  \\

		\hline
		$\varpi_{8} = {[1.87, 0.86, 0.06, 0.11]}$ & $972.7031$  & $820.315$  & $2247.9948$  & $2585.9703$  & $\textcolor{mydarkgreen}{\textbf{2794.2752}}$  & $\textcolor{blue}{\textbf{2747.4839}}$  & $\textcolor{red}{\textbf{3030.092}}$  \\ & $(± 147.8778)$  & $(± 920.6933)$  & $(± 980.5274)$  & $(± 735.6066)$  & $(± 588.4541)$  & $(± 805.0848)$  & $(± 157.7919)$  \\

		\hline
		$\varpi_{9} = {[0.62, 1.75, 0.06, 0.02]}$ & $\textcolor{blue}{\textbf{1888.9021}}$  & $1496.3557$  & $1595.3573$  & $\textcolor{mydarkgreen}{\textbf{2067.315}}$  & $1546.0998$  & $\textcolor{red}{\textbf{2081.7853}}$  & $1883.3591$  \\ & $(± 662.7498)$  & $(± 983.3814)$  & $(± 753.0129)$  & $(± 837.2501)$  & $(± 520.6149)$  & $(± 704.9607)$  & $(± 860.598)$  \\

		\hline
		$\varpi_{10} = {[1.9, 0.74, 0.01, 0.1]}$ & $987.615$  & $819.3686$  & $2268.2685$  & $2634.7086$  & $\textcolor{mydarkgreen}{\textbf{2793.5664}}$  & $\textcolor{blue}{\textbf{2785.9304}}$  & $\textcolor{red}{\textbf{3073.8208}}$  \\ & $(± 195.1827)$  & $(± 877.4964)$  & $(± 1017.4728)$  & $(± 757.0456)$  & $(± 746.6672)$  & $(± 870.7391)$  & $(± 219.8464)$  \\

		\hline
		$\varpi_{11} = {[1.67, 0.84, 0.08, 0.03]}$ & $1270.632$  & $1113.325$  & $2837.5326$  & $\textcolor{blue}{\textbf{3050.8883}}$  & $\textcolor{mydarkgreen}{\textbf{3087.1737}}$  & $2969.5175$  & $\textcolor{red}{\textbf{3134.7945}}$  \\ & $(± 153.1137)$  & $(± 1183.8638)$  & $(± 614.2109)$  & $(± 307.5426)$  & $(± 203.319)$  & $(± 607.8562)$  & $(± 145.066)$  \\

		\hline
	\end{tabular}
	\end{adjustbox}
\end{table*}

\begin{table*}[t]
  \captionsetup{width=\textwidth, justification=centering}
  \caption{Walker2d-v3 2D Sim-to-Sim evaluations CCS scores. The highest score is highlighted in red while the second and third highest scores are highlighted in green and blue respectively (\textcolor{red}{Red}$>$\textcolor{mydarkgreen}{Green}$>$\textcolor{blue}{Blue}). Hence, the column with the most red is the best.}
  \label{tab:walker_2dim_ccs}
  \centering
  \begin{adjustbox}{width=1\textwidth}
  \begin{tabular}{p{4.5cm}|c|c|c|c|c|c|c}
		\hline\hline
		$\varpi$ & DRSAC & SIRSA & cMDSAC (Ours) & eMDSAC (Ours) & uMD-SIRSA (Ours) & uMDSAC-v1 (Ours) & uMDSAC-v2 (Ours) \\
		\hline
		$\varpi_{1} = {[1.25, 1.25, 0.43, 0.43]}$ & $\textcolor{red}{\textbf{3518.2846}}$  & $1801.2659$  & $2734.0351$  & $3151.4706$  & $3073.8049$  & $\textcolor{blue}{\textbf{3259.8602}}$  & $\textcolor{mydarkgreen}{\textbf{3479.0658}}$  \\ & $(± 477.3515)$  & $(± 1336.2896)$  & $(± 960.2202)$  & $(± 452.0426)$  & $(± 426.7793)$  & $(± 507.8301)$  & $(± 739.1661)$  \\

		\hline
		$\varpi_{2} = {[0.74, 1.9, 0.13, 0.05]}$ & $3829.4187$  & $3550.9469$  & $3998.2533$  & $4504.066$  & $\textcolor{red}{\textbf{4647.7484}}$  & $\textcolor{blue}{\textbf{4544.0306}}$  & $\textcolor{mydarkgreen}{\textbf{4609.8294}}$  \\ & $(± 430.7219)$  & $(± 1422.9357)$  & $(± 1163.5906)$  & $(± 685.7721)$  & $(± 410.4905)$  & $(± 605.5297)$  & $(± 872.3166)$  \\

		\hline
		$\varpi_{3} = {[1.79, 1.46, 0.11, 0.29]}$ & $3048.3426$  & $1474.8984$  & $2939.1742$  & $3083.8538$  & $\textcolor{blue}{\textbf{3148.2214}}$  & $\textcolor{mydarkgreen}{\textbf{3190.8751}}$  & $\textcolor{red}{\textbf{3473.4067}}$  \\ & $(± 784.6853)$  & $(± 1256.4357)$  & $(± 904.0469)$  & $(± 905.9124)$  & $(± 401.6332)$  & $(± 537.3574)$  & $(± 515.0468)$  \\

		\hline
		$\varpi_{4} = {[1.76, 1.29, 0.11, 0.31]}$ & $3024.7044$  & $1337.8146$  & $3013.8434$  & $3175.687$  & $\textcolor{mydarkgreen}{\textbf{3245.9872}}$  & $\textcolor{blue}{\textbf{3220.3075}}$  & $\textcolor{red}{\textbf{3512.7577}}$  \\ & $(± 771.1141)$  & $(± 1312.3557)$  & $(± 870.7554)$  & $(± 759.828)$  & $(± 286.158)$  & $(± 548.4791)$  & $(± 510.0029)$  \\

		\hline
		$\varpi_{5} = {[1.61, 1.49, 0.15, 0.11]}$ & $3359.1542$  & $1693.0532$  & $3243.6889$  & $\textcolor{mydarkgreen}{\textbf{3633.8666}}$  & $\textcolor{blue}{\textbf{3366.2831}}$  & $3331.16$  & $\textcolor{red}{\textbf{3682.6468}}$  \\ & $(± 566.4236)$  & $(± 1280.2122)$  & $(± 766.7566)$  & $(± 325.1646)$  & $(± 341.6366)$  & $(± 576.4641)$  & $(± 524.3876)$  \\

		\hline
		$\varpi_{6} = {[1.36, 1.22, 0.05, 0.03]}$ & $3560.2712$  & $1717.8403$  & $3621.024$  & $\textcolor{red}{\textbf{4047.5151}}$  & $\textcolor{blue}{\textbf{3886.8498}}$  & $3874.004$  & $\textcolor{mydarkgreen}{\textbf{3999.5931}}$  \\ & $(± 659.4754)$  & $(± 1544.3769)$  & $(± 677.7429)$  & $(± 263.2184)$  & $(± 199.4342)$  & $(± 441.4766)$  & $(± 584.4008)$  \\

		\hline
		$\varpi_{7} = {[1.13, 1.61, 0.26, 0.08]}$ & $3747.3185$  & $2830.3946$  & $3421.5563$  & $\textcolor{mydarkgreen}{\textbf{3904.8028}}$  & $3801.1642$  & $\textcolor{blue}{\textbf{3834.9544}}$  & $\textcolor{red}{\textbf{4135.7003}}$  \\ & $(± 484.7109)$  & $(± 1320.959)$  & $(± 1163.193)$  & $(± 421.5108)$  & $(± 343.2847)$  & $(± 629.0879)$  & $(± 718.0712)$  \\

		\hline
		$\varpi_{8} = {[1.87, 0.86, 0.06, 0.11]}$ & $2720.5911$  & $972.174$  & $2881.0773$  & $\textcolor{mydarkgreen}{\textbf{3264.3783}}$  & $\textcolor{blue}{\textbf{3164.1366}}$  & $3160.3911$  & $\textcolor{red}{\textbf{3393.4378}}$  \\ & $(± 965.2928)$  & $(± 1053.3264)$  & $(± 914.6501)$  & $(± 431.0031)$  & $(± 214.3686)$  & $(± 617.9108)$  & $(± 506.1744)$  \\

		\hline
		$\varpi_{9} = {[0.62, 1.75, 0.06, 0.02]}$ & $3563.7993$  & $3618.0555$  & $4214.8283$  & $\textcolor{blue}{\textbf{4757.5229}}$  & $\textcolor{red}{\textbf{4996.4446}}$  & $\textcolor{mydarkgreen}{\textbf{4978.7969}}$  & $4486.0906$  \\ & $(± 688.6667)$  & $(± 1365.8451)$  & $(± 1565.3818)$  & $(± 736.5625)$  & $(± 483.0907)$  & $(± 579.7944)$  & $(± 1354.7107)$  \\

		\hline
		$\varpi_{10} = {[1.9, 0.74, 0.01, 0.1]}$ & $2654.9133$  & $823.7221$  & $2847.406$  & $\textcolor{mydarkgreen}{\textbf{3232.4713}}$  & $\textcolor{blue}{\textbf{3160.8182}}$  & $3120.3731$  & $\textcolor{red}{\textbf{3369.3018}}$  \\ & $(± 1054.0004)$  & $(± 961.7541)$  & $(± 913.565)$  & $(± 497.4788)$  & $(± 247.5882)$  & $(± 673.2414)$  & $(± 515.5197)$  \\

		\hline
		$\varpi_{11} = {[1.67, 0.84, 0.08, 0.03]}$ & $3255.0745$  & $1209.6049$  & $3125.5473$  & $\textcolor{mydarkgreen}{\textbf{3610.3818}}$  & $\textcolor{blue}{\textbf{3433.276}}$  & $3371.1694$  & $\textcolor{red}{\textbf{3628.1878}}$  \\ & $(± 689.7449)$  & $(± 1293.7924)$  & $(± 829.2888)$  & $(± 326.1735)$  & $(± 219.0386)$  & $(± 581.277)$  & $(± 551.3985)$ \\

		\hline
	\end{tabular}
	\end{adjustbox}
\end{table*}

\subsubsection{Results:}

 The tables~\ref{tab:dclaw_2dim_ccs}, \ref{tab:hopper_2dim_ccs}, \ref{tab:walker_2dim_ccs}, \ref{tab:dclaw_multidim_ccs}, \ref{tab:hopper_multidim_ccs}, \ref{tab:walker_multidim_ccs} show the CCS score obtained after performing 50 trials for each of 8 different random seed models in both 2-dimensional and high-dimensional cases.

 As can be seen, in the 2-dimensional case (corresponding to the tables~\ref{tab:dclaw_2dim_ccs}, \ref{tab:hopper_2dim_ccs}, \ref{tab:walker_2dim_ccs}), \emph{uMDSAC-v2} is overall the best algorithm. Indeed, while \emph{uMDSAC-v1} is almost as good in the D'Claw environment, it fails to land properly on the podium in the Walker2d environment. \emph{uMD-SIRSA}, the hybrid algorithm between \emph{SIRSA} and \emph{uMDSAC}, analogously performs competitively on the Hopper environment, being consistently second best when both \emph{uMDSAC} sometimes takes the third place, but fails to consistently make it to the top three on the D'Claw environment.
 
 Although \emph{uMDSAC-v2} clearly tops the other algorithms in the 2-dimensional case, in higher dimensions (corresponding to the tables~\ref{tab:dclaw_multidim_ccs}, \ref{tab:hopper_multidim_ccs}, \ref{tab:walker_multidim_ccs}), this trend is no longer observed. Worst yet, none of the algorithms appear to decisively outperform the others. Indeed, while \emph{eMDSAC} is the best algorithm in the D'Claw environment, in the Hopper environment it is superseded by the \emph{cMDSAC} algorithm. \emph{cMDSAC} is then in turn outperformed by the \emph{uMDSAC-v1} algorithm in the Walker2d environment.

 Despite that, we can still find out which algorithm is, if not always the best, always guaranteed to be in the top three. Indeed, from Table~\ref{tab:dclaw_multidim_ccs}, we can observe the following hierarchy: \emph{eMDSAC} $>$ \emph{uMDSAC-v2} $>$ \emph{DRSAC} $>$ \emph{uMDSAC-v1} (indeed, after \emph{eMDSAC}, \emph{uMDSAC-v2} is the most colorful with three ``green" and four ``blue". \Mycomment{ Although \emph{DRSAC} has one more ``red", \emph{uMDSAC-v1} has more ``green" }). Similarly, from Table~\ref{tab:hopper_multidim_ccs}, we have \emph{cMDSAC} $>$ \emph{uMDSAC-v2} $>$ \emph{uMD-SIRSA} $>$ \emph{uMDSAC-v1} and from Table~\ref{tab:walker_multidim_ccs}, \emph{uMDSAC-v1} $>$ \emph{uMD-SIRSA} $>$ \emph{uMDSAC-v2} $>$ \emph{cMDSAC}. Hence, \emph{uMDSAC-v2} still offers the best comprise among all of the algorithms for the environments considered.
 
 In both of the 2-dimensional and high-dimensional cases, we can also notice that SIRSA performs hardly better than DRSAC. This is explained by the fact that in the SIRSA algorithm, one must first sample a fixed number of uncertainty sets (here we use the same number $100$ as the one in the original paper) and then train the policy on these uncertainty sets. Hence, the performance of the algorithm heavily depends on the uncertainty sets sampled and therefore also on the random seed. In our experiments, for fairness, we did not consider any specific tuning for the fixed uncertainty sets, hence its poor performance.
 
 Overall, \emph{uMDSAC-v2} therefore appears to be the best algorithm for efficiently learning uncertainty-aware policies that approximate the CCS.
\begin{table*}[t]
  \captionsetup{width=\textwidth, justification=centering}
  \caption{DClawTurnFixed-v0 22D Sim-to-Sim evaluations CCS scores. The highest score is highlighted in red while the second and third highest scores are highlighted in green and blue respectively (\textcolor{red}{Red}$>$\textcolor{mydarkgreen}{Green}$>$\textcolor{blue}{Blue}). Hence, the column with the most red is the best.}
  \label{tab:dclaw_multidim_ccs}
  \centering
  \begin{adjustbox}{width=1\textwidth}
  \begin{tabular}{p{2cm}|c|c|c|c|c|c|c}
		\hline\hline
		$\varpi$ & DRSAC & SIRSA & cMDSAC (Ours) & eMDSAC (Ours) & uMD-SIRSA (Ours) & uMDSAC-v1 (Ours) & uMDSAC-v2 (Ours) \\
		\hline
		$\varpi_{1}$ & $\textcolor{blue}{\textbf{237.0354}}$  & $-572.3336$  & $213.6282$  & $\textcolor{red}{\textbf{333.7877}}$  & $205.0277$  & $\textcolor{mydarkgreen}{\textbf{257.1773}}$  & $232.4141$  \\ & $(± 115.214)$  & $(± 142.7074)$  & $(± 135.504)$  & $(± 139.3518)$  & $(± 206.685)$  & $(± 82.4278)$  & $(± 60.9046)$  \\

		\hline
		$\varpi_{2}$ & $284.672$  & $-462.5208$  & $515.9602$  & $\textcolor{red}{\textbf{667.6373}}$  & $389.421$  & $\textcolor{mydarkgreen}{\textbf{569.4012}}$  & $\textcolor{blue}{\textbf{517.1394}}$  \\ & $(± 256.4479)$  & $(± 387.4888)$  & $(± 197.4969)$  & $(± 166.3724)$  & $(± 192.2084)$  & $(± 161.4897)$  & $(± 150.0641)$  \\

		\hline
		$\varpi_{3}$ & $-426.0525$  & $-724.6991$  & $-321.5364$  & $\textcolor{red}{\textbf{-158.5956}}$  & $-375.2957$  & $\textcolor{blue}{\textbf{-298.4371}}$  & $\textcolor{mydarkgreen}{\textbf{-274.2848}}$  \\ & $(± 239.4242)$  & $(± 90.1514)$  & $(± 225.558)$  & $(± 245.3578)$  & $(± 181.755)$  & $(± 274.4661)$  & $(± 297.6253)$  \\

		\hline
		$\varpi_{4}$ & $\textcolor{mydarkgreen}{\textbf{201.0208}}$  & $-682.6824$  & $132.779$  & $\textcolor{red}{\textbf{317.0557}}$  & $89.5324$  & $-16.7846$  & $\textcolor{blue}{\textbf{175.3097}}$  \\ & $(± 256.0806)$  & $(± 145.0523)$  & $(± 343.1106)$  & $(± 207.2087)$  & $(± 316.9227)$  & $(± 269.5543)$  & $(± 228.3241)$  \\

		\hline
		$\varpi_{5}$ & $\textcolor{blue}{\textbf{-597.5513}}$  & $-701.6267$  & $\textcolor{mydarkgreen}{\textbf{-596.962}}$  & $\textcolor{red}{\textbf{-558.1447}}$  & $-628.3319$  & $-604.5063$  & $-603.957$  \\ & $(± 45.6794)$  & $(± 39.0303)$  & $(± 50.9708)$  & $(± 73.3719)$  & $(± 54.4551)$  & $(± 43.8835)$  & $(± 37.0137)$  \\

		\hline
		$\varpi_{6}$ & $\textcolor{red}{\textbf{922.676}}$  & $-340.7216$  & $352.7469$  & $\textcolor{mydarkgreen}{\textbf{507.2193}}$  & $292.1519$  & $284.2891$  & $\textcolor{blue}{\textbf{456.578}}$  \\ & $(± 96.7141)$  & $(± 403.5193)$  & $(± 192.3073)$  & $(± 139.5459)$  & $(± 198.0986)$  & $(± 121.898)$  & $(± 99.1317)$  \\

		\hline
		$\varpi_{7}$ & $903.8725$  & $-330.5345$  & $\textcolor{mydarkgreen}{\textbf{1012.6412}}$  & $\textcolor{red}{\textbf{1054.2995}}$  & $880.0715$  & $907.6697$  & $\textcolor{blue}{\textbf{954.8724}}$  \\ & $(± 126.709)$  & $(± 390.9088)$  & $(± 108.1756)$  & $(± 84.7809)$  & $(± 237.1899)$  & $(± 150.5904)$  & $(± 180.0533)$  \\

		\hline
		$\varpi_{8}$ & $806.1864$  & $-280.062$  & $\textcolor{blue}{\textbf{899.2468}}$  & $\textcolor{red}{\textbf{946.2957}}$  & $749.7205$  & $\textcolor{mydarkgreen}{\textbf{926.858}}$  & $859.4435$  \\ & $(± 236.1734)$  & $(± 381.5728)$  & $(± 108.2814)$  & $(± 45.8901)$  & $(± 263.3619)$  & $(± 205.333)$  & $(± 122.9577)$  \\

		\hline
		$\varpi_{9}$ & $-539.5034$  & $-695.1658$  & $-462.5124$  & $\textcolor{blue}{\textbf{-404.1614}}$  & $\textcolor{mydarkgreen}{\textbf{-396.9162}}$  & $\textcolor{red}{\textbf{-373.2156}}$  & $-421.6303$  \\ & $(± 38.0489)$  & $(± 47.195)$  & $(± 136.3865)$  & $(± 143.4837)$  & $(± 166.6368)$  & $(± 231.4245)$  & $(± 124.5182)$  \\

		\hline
		$\varpi_{10}$ & $\textcolor{red}{\textbf{982.5445}}$  & $-301.3247$  & $387.8474$  & $\textcolor{blue}{\textbf{396.5654}}$  & $280.2142$  & $356.1581$  & $\textcolor{mydarkgreen}{\textbf{409.6402}}$  \\ & $(± 61.8875)$  & $(± 379.1755)$  & $(± 219.8893)$  & $(± 99.1377)$  & $(± 254.0094)$  & $(± 107.1401)$  & $(± 141.7691)$  \\

		\hline
		$\varpi_{11}$ & $-260.2414$  & $-685.7196$  & $\textcolor{blue}{\textbf{7.8834}}$  & $\textcolor{red}{\textbf{171.8334}}$  & $-72.4414$  & $-99.4681$  & $\textcolor{mydarkgreen}{\textbf{39.9494}}$  \\ & $(± 163.2047)$  & $(± 51.3708)$  & $(± 281.7159)$  & $(± 217.1697)$  & $(± 336.7655)$  & $(± 279.5182)$  & $(± 122.2564)$ \\

		\hline
	\end{tabular}
	\end{adjustbox}
\end{table*}

\begin{table*}[t]
  \captionsetup{width=\textwidth, justification=centering}
  \caption{Hopper-v3 12D Sim-to-Sim evaluations CCS scores. The highest score is highlighted in red while the second and third highest scores are highlighted in green and blue respectively (\textcolor{red}{Red}$>$\textcolor{mydarkgreen}{Green}$>$\textcolor{blue}{Blue}). Hence, the column with the most red is the best.}
  \label{tab:hopper_multidim_ccs}
  \centering
  \begin{adjustbox}{width=1\textwidth}
  \begin{tabular}{p{2cm}|c|c|c|c|c|c|c}
		\hline\hline
		$\varpi$ & DRSAC & SIRSA & cMDSAC (Ours) & eMDSAC (Ours) & uMD-SIRSA (Ours) & uMDSAC-v1 (Ours) & uMDSAC-v2 (Ours) \\
		\hline
		$\varpi_{1}$ & $1300.6304$  & $1326.8353$  & $2233.1447$  & $2058.6257$  & $\textcolor{blue}{\textbf{2468.2994}}$  & $\textcolor{mydarkgreen}{\textbf{2580.6852}}$  & $\textcolor{red}{\textbf{2681.5032}}$  \\ & $(± 457.8133)$  & $(± 698.1123)$  & $(± 189.8573)$  & $(± 268.2997)$  & $(± 322.1358)$  & $(± 128.0213)$  & $(± 128.444)$  \\

		\hline
		$\varpi_{2}$ & $1345.892$  & $1125.3599$  & $\textcolor{red}{\textbf{2493.7272}}$  & $2268.5828$  & $1978.1856$  & $\textcolor{mydarkgreen}{\textbf{2482.2225}}$  & $\textcolor{blue}{\textbf{2459.7566}}$  \\ & $(± 693.8424)$  & $(± 876.33)$  & $(± 397.5463)$  & $(± 438.5787)$  & $(± 822.971)$  & $(± 235.4931)$  & $(± 379.8202)$  \\

		\hline
		$\varpi_{3}$ & $2177.0686$  & $1506.6745$  & $\textcolor{red}{\textbf{3572.7845}}$  & $\textcolor{mydarkgreen}{\textbf{3476.1291}}$  & $2781.858$  & $3450.5034$  & $\textcolor{blue}{\textbf{3464.9857}}$  \\ & $(± 1060.074)$  & $(± 843.582)$  & $(± 64.6847)$  & $(± 63.4483)$  & $(± 858.5226)$  & $(± 168.2517)$  & $(± 146.5297)$  \\

		\hline
		$\varpi_{4}$ & $954.2426$  & $1183.8283$  & $\textcolor{mydarkgreen}{\textbf{3023.9668}}$  & $2809.5845$  & $2847.892$  & $\textcolor{red}{\textbf{3070.8964}}$  & $\textcolor{blue}{\textbf{2880.8178}}$  \\ & $(± 407.9529)$  & $(± 812.1805)$  & $(± 240.5186)$  & $(± 227.1655)$  & $(± 472.5372)$  & $(± 120.9817)$  & $(± 276.9204)$  \\

		\hline
		$\varpi_{5}$ & $1969.0966$  & $1053.7671$  & $\textcolor{red}{\textbf{2950.3496}}$  & $\textcolor{blue}{\textbf{2839.4337}}$  & $2739.3824$  & $2765.6128$  & $\textcolor{mydarkgreen}{\textbf{2907.9963}}$  \\ & $(± 786.1205)$  & $(± 713.8866)$  & $(± 45.6208)$  & $(± 110.9394)$  & $(± 209.1572)$  & $(± 263.8676)$  & $(± 69.2827)$  \\

		\hline
		$\varpi_{6}$ & $849.0959$  & $742.7881$  & $\textcolor{red}{\textbf{2991.0581}}$  & $2544.0004$  & $\textcolor{blue}{\textbf{2745.8706}}$  & $2726.9436$  & $\textcolor{mydarkgreen}{\textbf{2778.7025}}$  \\ & $(± 281.2457)$  & $(± 554.0967)$  & $(± 128.0052)$  & $(± 552.4728)$  & $(± 399.196)$  & $(± 486.1404)$  & $(± 300.0653)$  \\

		\hline
		$\varpi_{7}$ & $462.7684$  & $452.3087$  & $2611.3835$  & $\textcolor{mydarkgreen}{\textbf{2649.8378}}$  & $\textcolor{blue}{\textbf{2646.3645}}$  & $2598.4224$  & $\textcolor{red}{\textbf{2871.2423}}$  \\ & $(± 84.9125)$  & $(± 264.9766)$  & $(± 710.9714)$  & $(± 305.6031)$  & $(± 547.7926)$  & $(± 446.2572)$  & $(± 178.4836)$  \\

		\hline
		$\varpi_{8}$ & $728.5938$  & $898.181$  & $\textcolor{mydarkgreen}{\textbf{3012.5108}}$  & $2720.0152$  & $2816.668$  & $\textcolor{blue}{\textbf{2973.8199}}$  & $\textcolor{red}{\textbf{3034.2115}}$  \\ & $(± 188.815)$  & $(± 796.3476)$  & $(± 250.6376)$  & $(± 588.5513)$  & $(± 343.0552)$  & $(± 238.948)$  & $(± 102.7548)$  \\

		\hline
		$\varpi_{9}$ & $2001.2405$  & $1689.0671$  & $\textcolor{red}{\textbf{3092.1574}}$  & $\textcolor{blue}{\textbf{2884.2181}}$  & $2538.7188$  & $2579.3928$  & $\textcolor{mydarkgreen}{\textbf{2888.6953}}$  \\ & $(± 863.0336)$  & $(± 953.1295)$  & $(± 180.9248)$  & $(± 299.0357)$  & $(± 352.4499)$  & $(± 434.6877)$  & $(± 446.0168)$  \\

		\hline
		$\varpi_{10}$ & $989.605$  & $597.2106$  & $\textcolor{red}{\textbf{2789.0977}}$  & $\textcolor{blue}{\textbf{2591.5504}}$  & $\textcolor{mydarkgreen}{\textbf{2605.6394}}$  & $2437.3772$  & $2587.2779$  \\ & $(± 487.157)$  & $(± 272.8945)$  & $(± 37.3462)$  & $(± 300.336)$  & $(± 373.7277)$  & $(± 433.7174)$  & $(± 330.2287)$  \\

		\hline
		$\varpi_{11}$ & $1883.0439$  & $1230.3173$  & $\textcolor{mydarkgreen}{\textbf{2921.6126}}$  & $\textcolor{blue}{\textbf{2881.1589}}$  & $\textcolor{red}{\textbf{2933.299}}$  & $2822.3036$  & $2807.4377$  \\ & $(± 663.5933)$  & $(± 942.8375)$  & $(± 84.5593)$  & $(± 75.2379)$  & $(± 82.3575)$  & $(± 63.0276)$ & $(± 103.7656)$ \\

		\hline
	\end{tabular}
	\end{adjustbox}
\end{table*}

\begin{table*}[t]
  \captionsetup{width=\textwidth, justification=centering}
  \caption{Walker2d-v3 21D Sim-to-Sim evaluations CCS scores. The highest score is highlighted in red while the second and third highest scores are highlighted in green and blue respectively (\textcolor{red}{Red}$>$\textcolor{mydarkgreen}{Green}$>$\textcolor{blue}{Blue}). Hence, the column with the most red is the best.}
  \label{tab:walker_multidim_ccs}
  \centering
  \begin{adjustbox}{width=1\textwidth}
  \begin{tabular}{p{2cm}|c|c|c|c|c|c|c}
		\hline\hline
		$\varpi$ & DRSAC & SIRSA & cMDSAC (Ours) & eMDSAC (Ours) & uMD-SIRSA (Ours) & uMDSAC-v1 (Ours) & uMDSAC-v2 (Ours) \\
		\hline
		$\varpi_{1}$ & $3433.2934$  & $1323.8685$  & $\textcolor{blue}{\textbf{3931.7936}}$  & $3448.4635$  & $\textcolor{red}{\textbf{4192.7058}}$  & $\textcolor{mydarkgreen}{\textbf{4178.0038}}$  & $3924.3896$  \\ & $(± 696.177)$  & $(± 786.0416)$  & $(± 681.8913)$  & $(± 691.4435)$  & $(± 420.9694)$  & $(± 525.399)$  & $(± 383.3146)$  \\

		\hline
		$\varpi_{2}$ & $2201.4192$  & $716.6437$  & $\textcolor{blue}{\textbf{3397.8297}}$  & $3396.2943$  & $\textcolor{mydarkgreen}{\textbf{3675.035}}$  & $\textcolor{red}{\textbf{3817.5924}}$  & $3335.5168$  \\ & $(± 944.3087)$  & $(± 819.1268)$  & $(± 902.2641)$  & $(± 520.0866)$  & $(± 322.672)$  & $(± 410.595)$  & $(± 779.6465)$  \\

		\hline
		$\varpi_{3}$ & $3332.7673$  & $2060.9423$  & $4190.078$  & $4238.1199$  & $\textcolor{red}{\textbf{4694.5352}}$  & $\textcolor{mydarkgreen}{\textbf{4583.2537}}$  & $\textcolor{blue}{\textbf{4526.2967}}$  \\ & $(± 778.2211)$  & $(± 1196.6469)$  & $(± 616.8122)$  & $(± 483.4928)$  & $(± 411.8515)$  & $(± 829.8579)$  & $(± 421.4979)$  \\

		\hline
		$\varpi_{4}$ & $3350.1185$  & $1622.4805$  & $3639.7043$  & $3486.416$  & $\textcolor{mydarkgreen}{\textbf{3800.3933}}$  & $\textcolor{red}{\textbf{3961.1034}}$  & $\textcolor{blue}{\textbf{3667.1687}}$  \\ & $(± 326.7588)$  & $(± 807.2248)$  & $(± 424.5692)$  & $(± 575.7388)$  & $(± 326.8727)$  & $(± 246.2397)$  & $(± 266.9335)$  \\

		\hline
		$\varpi_{5}$ & $3838.1673$  & $1424.2942$  & $4405.0414$  & $4262.6117$  & $\textcolor{red}{\textbf{4684.5495}}$  & $\textcolor{mydarkgreen}{\textbf{4632.2886}}$  & $\textcolor{blue}{\textbf{4534.743}}$  \\ & $(± 538.1616)$  & $(± 886.0849)$  & $(± 474.171)$  & $(± 532.8634)$  & $(± 307.9186)$  & $(± 473.3707)$  & $(± 417.7406)$  \\

		\hline
		$\varpi_{6}$ & $3622.2182$  & $1321.3894$  & $4261.8153$  & $3940.8412$  & $\textcolor{mydarkgreen}{\textbf{4340.6352}}$  & $\textcolor{red}{\textbf{4469.6424}}$  & $\textcolor{blue}{\textbf{4338.2361}}$  \\ & $(± 850.5584)$  & $(± 1088.031)$  & $(± 541.7246)$  & $(± 426.58)$  & $(± 539.4903)$  & $(± 401.1468)$  & $(± 448.9267)$  \\

		\hline
		$\varpi_{7}$ & $3317.6661$  & $1257.6275$  & $\textcolor{mydarkgreen}{\textbf{3722.679}}$  & $3467.5896$  & $3465.669$  & $\textcolor{red}{\textbf{3875.7516}}$  & $\textcolor{blue}{\textbf{3509.6657}}$  \\ & $(± 728.1113)$  & $(± 859.8451)$  & $(± 421.5296)$  & $(± 262.044)$  & $(± 630.1653)$  & $(± 388.3618)$  & $(± 485.8133)$  \\

		\hline
		$\varpi_{8}$ & $3484.2513$  & $2446.086$  & $4093.9695$  & $3722.9688$  & $\textcolor{blue}{\textbf{4176.9696}}$  & $\textcolor{red}{\textbf{4386.95}}$  & $\textcolor{mydarkgreen}{\textbf{4311.5814}}$  \\ & $(± 400.2656)$  & $(± 620.1616)$  & $(± 436.7499)$  & $(± 578.7529)$  & $(± 427.1366)$  & $(± 382.7648)$  & $(± 456.7172)$  \\

		\hline
		$\varpi_{9}$ & $3452.932$  & $1503.3069$  & $\textcolor{blue}{\textbf{3894.6345}}$  & $3803.1132$  & $3799.5894$  & $\textcolor{red}{\textbf{4205.3187}}$  & $\textcolor{mydarkgreen}{\textbf{4065.3038}}$  \\ & $(± 832.8624)$  & $(± 1050.7717)$  & $(± 544.2426)$  & $(± 211.6056)$  & $(± 775.5775)$  & $(± 307.5163)$  & $(± 402.2682)$  \\

		\hline
		$\varpi_{10}$ & $3919.5806$  & $1487.2424$  & $\textcolor{blue}{\textbf{4300.4085}}$  & $3979.9616$  & $\textcolor{mydarkgreen}{\textbf{4322.6407}}$  & $\textcolor{red}{\textbf{4425.8382}}$  & $4218.6358$  \\ & $(± 493.7754)$  & $(± 1063.2785)$  & $(± 378.4825)$  & $(± 441.8303)$  & $(± 346.7213)$  & $(± 340.8703)$  & $(± 332.6603)$  \\

		\hline
		$\varpi_{11}$ & $3617.4551$  & $1287.2819$  & $\textcolor{mydarkgreen}{\textbf{4331.3734}}$  & $3829.1445$  & $4215.6263$  & $\textcolor{blue}{\textbf{4294.0949}}$  & $\textcolor{red}{\textbf{4339.5526}}$  \\ & $(± 1014.1674)$  & $(± 1083.6038)$  & $(± 537.0895)$  & $(± 547.3069)$  & $(± 532.4654)$  & $(± 580.0009)$ & $(± 563.7209)$ \\

		\hline
	\end{tabular}
	\end{adjustbox}
\end{table*}

\subsection{Sim-to-real transfer}

\subsubsection{Setup:}

\begin{figure}[tb]
     \centering
     \captionsetup{justification=centering}
     \includegraphics[keepaspectratio=true,width=\linewidth]{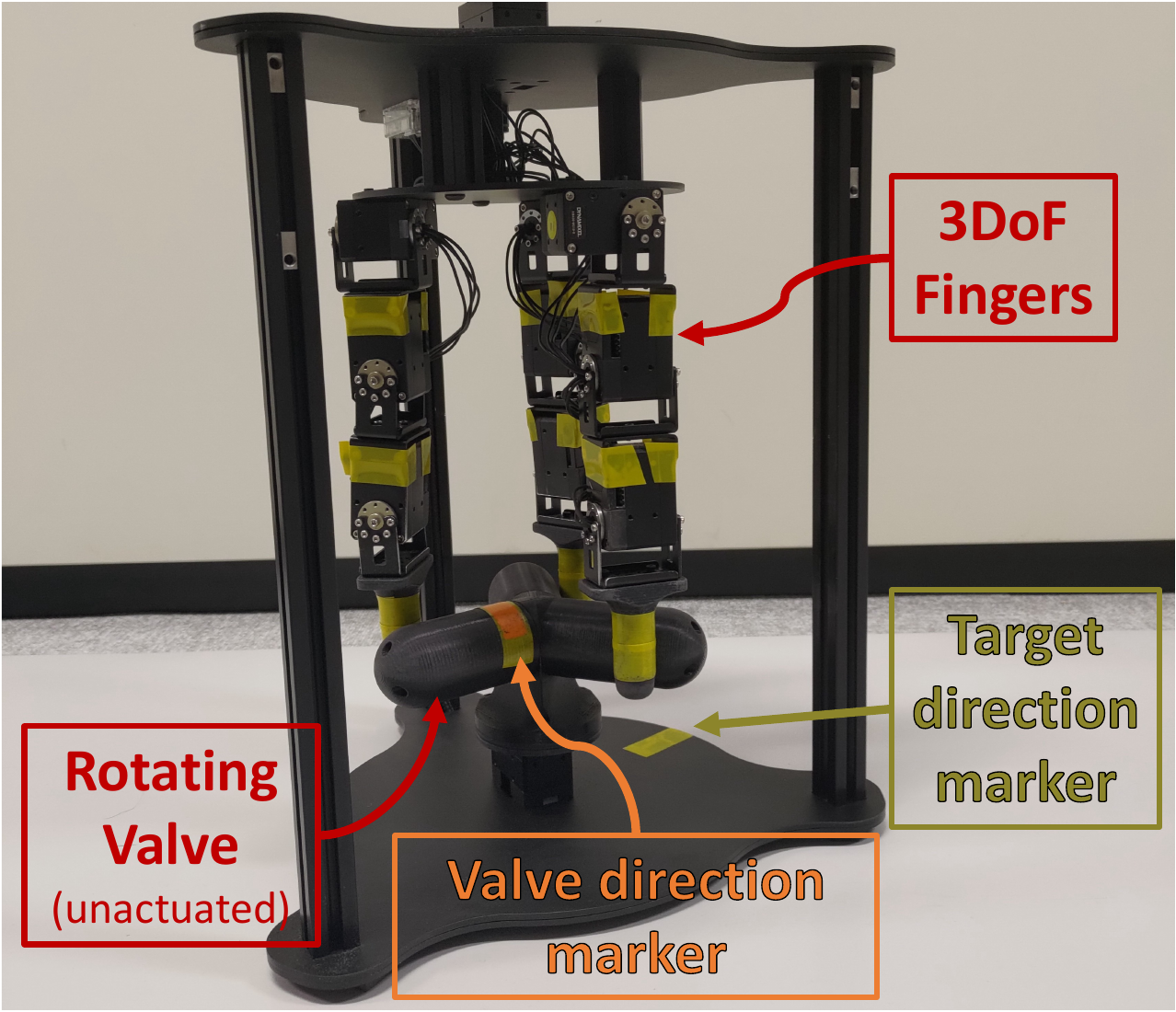}
     \caption{D'Claw robot used in the sim-to-real experiment}
     \label{fig:dclaw}
\end{figure}

To further verify the efficiency of the MDSAC algorithms, and the advantage of using an uncertainty-aware policy for sim-to-real transfer, we conducted a real robot experiment using the D'Claw robot depicted in Fig.~\ref{fig:dclaw}.
D'Claw is a platform introduced by project-ROBEL (RObotics BEnchmarks for Learning by~\cite{ahn2020robel}) for studying and benchmarking dexterous manipulation. It is a nine degrees of freedom (DoFs) platform that consists of three identical fingers mounted symmetrically on a base.

In the experiments, we focus on the transferability of the agents trained on the DClawTurnFixed-v0 task (both 2-dimensional and high-dimensional randomization models), which consists in rotating a passive DoF (the object - or valve - located on the middle of the base in Fig.~\ref{fig:dclaw}) from a fixed starting position to a fixed target angle, while respecting some safety constraints captured by the reward function, with a fixed episode length of $40$ steps. Specifically, the task consists in turning the object from the angle $0.0$ to the target angle $\pi$ while minimizing the velocity and staying as close as possible to the nominal position.

 The actions' dimension is $9$, corresponding to the position of the fingers' joints and the observation space is given by the angular position $\theta$ of the fingers' nine joints, the previous action, the current angular position of the object represented by the cosine and sine values, and finally the error between the current position and the target position, for a total dimension of $21$.

\subsubsection{Results:}

%
\begin{table*}
\captionsetup{justification=centering}
\centering
    \begin{minipage}[c]{\linewidth}\centering
        \begin{tabular}{|c|c|c|c|c|}
			\hline
			\multirow{3}{*}{\textbf{Algo}} & \multicolumn{4}{c|}{\textbf{Scores}} \\
			\cline{2-5}
			 & \multicolumn{2}{c|}{Fixed $\varpi$ (Full uncertainty)} & \multicolumn{2}{c|}{Adaptive $\varpi$ (OSI)} \\
			 \cline{2-5}
			 & 2d & High-Dim & 2d & High-Dim \\
			\hline
			DRSAC & $\textcolor{red}{\textbf{-397.4923}}$ & $-626.2161$ & $\textcolor{mydarkgreen}{\textbf{-397.4923}}$ & $\textcolor{blue}{\textbf{-626.2161}}$ \\ & $(± 481.3695)$ & $(± 226.7301)$ & $(± 481.3695)$ & $(± 226.7301)$ \\
			\cline{1-5}
			SIRSA & $-697.3132$ & $-695.3794$ & $-697.3132$ & $-695.3794$ \\ & $(± 146.4365)$ & $(± 254.4036)$ & $(± 146.4365)$ & $(± 254.4036)$ \\
			\cline{1-5}
			cMDSAC (Ours) & $\textcolor{blue}{\textbf{-455.8342}}$ & $\textcolor{red}{\textbf{-461.3371}}$ & $\textcolor{red}{\textbf{-326.5449}}$ & $-655.2146$ \\ & $(± 685.9239)$ & $(± 286.9469)$ & $(± 669.7612)$ & $(± 212.5434)$ \\
			\cline{1-5}
			eMDSAC (Ours) & $-592.406$ & $\textcolor{mydarkgreen}{\textbf{-507.7111}}$ & $-573.2$ & $-710.3957$ \\ & $(± 288.7826)$ & $(± 551.7123)$ & $(± 340.2891)$ & $(± 206.1083)$ \\
			\cline{1-5}
			uMD-SIRSA (Ours) & $-609.5321$ & $-667.3349$ & $-546.3307$ & $-683.8679$ \\ & $(± 299.5206)$ & $(± 295.6178)$ & $(± 410.3494)$ & $(± 264.3838)$ \\
			\cline{1-5}
			uMDSAC-v1 (Ours) & $\textcolor{mydarkgreen}{\textbf{-454.5132}}$ & $\textcolor{blue}{\textbf{-550.0329}}$ & $\textcolor{blue}{\textbf{-485.6712}}$ & $\textcolor{red}{\textbf{-483.6149}}$ \\ & $(± 629.2461)$ & $(± 233.1124)$ & $(± 580.9715)$ & $(± 228.8893)$ \\
			\cline{1-5}
			uMDSAC-v2 (Ours) & $-518.2694$ & $-661.1004$ & $-579.5727$ & $\textcolor{mydarkgreen}{\textbf{-574.8425}}$ \\ & $(± 410.22)$ & $(± 217.7788)$ & $(± 376.5343)$ & $(± 273.2554)$ \\
			\hline
		\end{tabular}
    \end{minipage}
    \caption{Evaluations on the real D'Claw robot using both models trained on the 2-dimensional parameter space (2d column) and the high-dimensional parameter space (High-Dim column), and with or without the use of the OSI during the episode to change the uncertainty. In the fixed uncertainty case, $\varpi$ is set to the full randomization range (i.e. the MDSAC policies operate with full uncertainty). As before, the highest score for each configuration or column is highlighted in red while the second and third highest scores are highlighted in green and blue respectively (\textcolor{red}{Red}$>$\textcolor{mydarkgreen}{Green}$>$\textcolor{blue}{Blue}).}
    \label{tab:real_evals}
\end{table*}

 For each algorithm and each trained model, we run 50 trials on the robot. We first run the MDSAC algorithms without the OSI, using a fixed $\varpi$ input corresponding to the full uncertainty. After these evaluations, we run the algorithms using the OSI, which reduces the uncertainty during each episode. The results of the evaluation is summarized in the Table~\ref{tab:real_evals}.

 As can be seen, in the 2-dimensional case, when the uncertainty is at its fullest and fixed, the performance of the MDSAC algorithms remain inferior to that of \emph{DRSAC}. This however does not hold for the high-dimensional case where \emph{cMDSAC}, \emph{eMDSAC} and \emph{uMD-SIRSA} give a better overall performance.

 Interestingly, even though the use of the OSI improves the performance of the \emph{cMDSAC}, \emph{eMDSAC} and \emph{uMD-SIRSA} algorithms in the 2-dimensional case, we notice instead a drop in performance for the high-dimensional case. In contrast, the \emph{uMDSAC} algorithms show the reverse tendency with the OSI causing a drop in performance in the 2-dimensional parameter case and an increase in the higher dimensional evaluations. A quick look at the performance difference from the \emph{DRSAC} confirms that the high dimensional randomization tended to produce a more conservative policy than the 2-dimensional randomization. Hence, one would expect that the use of the OSI should lead to a particularly sharp increase in performance in the high-dimensional case for all MDSAC algorithms, which is the case only for both \emph{uMDSAC} algorithms.

\begin{figure*}[tb]
    \centering
    \captionsetup{justification=centering}
    \begin{minipage}[c]{\linewidth}
        \centering
        \includegraphics[keepaspectratio=true,width=\linewidth]{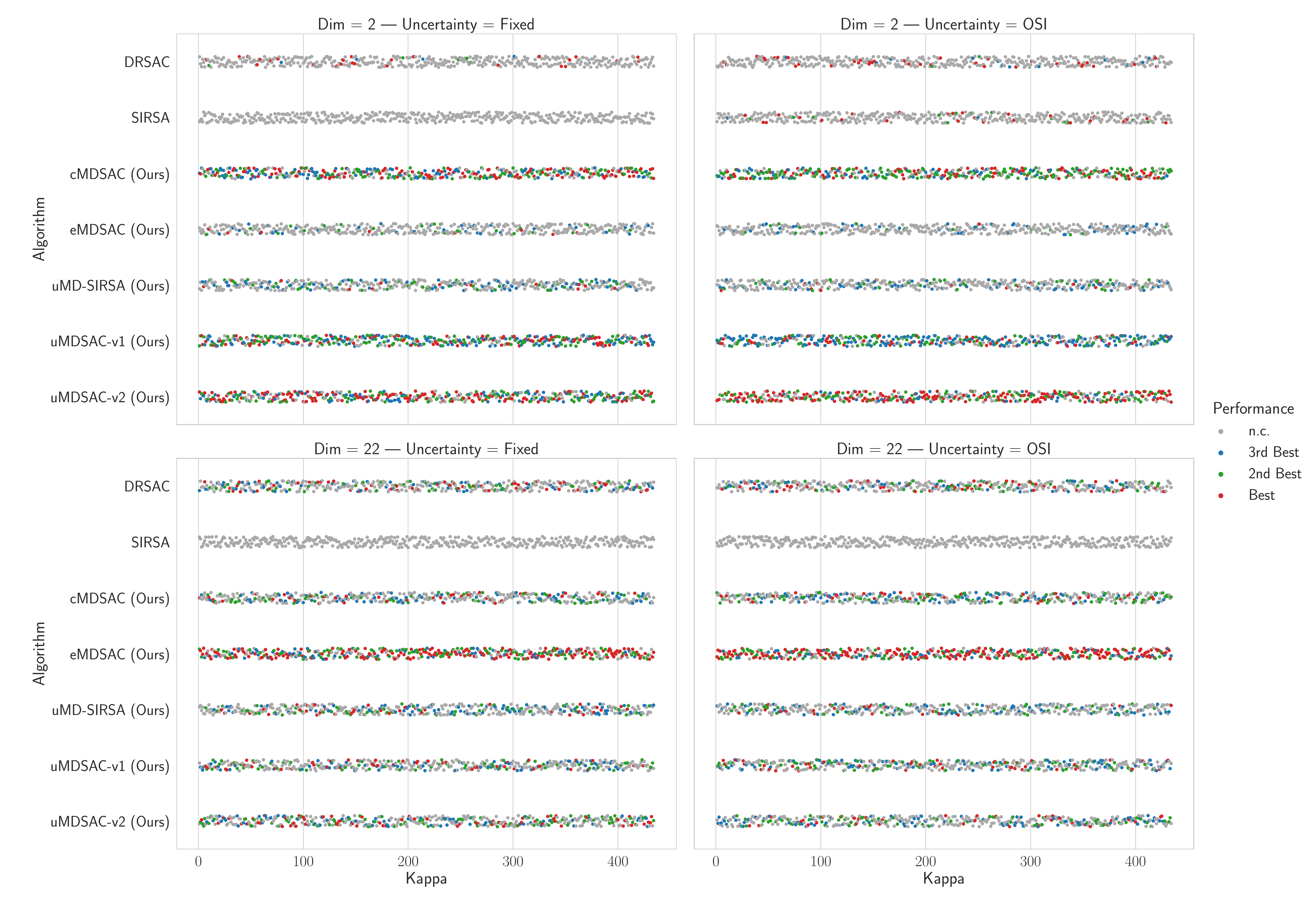}
    \end{minipage}
    \caption{Sim-to-Sim performance on the DClawTurnFixed-v0 simulation on 434 simulation parameters (``Kappa" - $\kappa$) sampled randomly from the full domain range. ``n.c." is short for ``not classified" and refers to values that fall outside the top three. The algorithm with the most red is the overall best.}
    \label{fig:sim2sim_with_osi}
\end{figure*}

 To better understand what is happening, we consider the sim-to-sim transfer results in Fig.~\ref{fig:sim2sim_with_osi} which shows the performance of the various algorithms on 434 simulation parameters sampled randomly from the full randomized domain range. As noticed, the performance is in accord with the CCS results, with \emph{uMDSAC-v2} being the best on the 2-dimensional case and \emph{eMDSAC} being the best on the high-dimensional case, regardless of whether or not the OSI was employed. However, we can also notice that all algorithms have gray points meaning that none of the algorithms evaluated was able to consistently remain in the top three best performance (given by the colors red, green and blue). Table~\ref{tab:sim2sim_evals} shows the exact number of parameters on which each algorithm performed the best, second best and third best, while Table~\ref{tab:sim2sim_eval_tendency} shows the effect of the use of the OSI on the performance compared to simply using the full uncertainty as input to the actors.

 From both tables, we can see that the use of an OSI was beneficial to \emph{cMDSAC} on both of the randomization dimension, but beneficial to \emph{uMDSAC-v2} only on the 2-dimensional case and to \emph{eMDSAC} and \emph{uMDSAC-v1} only on the high dimensional case. Furthermore, the number of environments on which the OSI is beneficial is not overwhelmingly larger than the number of environments where its effect caused a drop in performance. This can explain the mixed results on the sim-to-real transfer of Table~\ref{tab:real_evals} as the probability for an increase or decrease in performance is nearly a fifty-fifty for most of the MDSAC algorithms.
\begin{table*}
\captionsetup{justification=centering}
\centering
    \begin{minipage}[c]{0.5\linewidth}\centering
        \subcaption{Dim $=$ 2 $-$ Uncertainty $=$ Fixed}
        \label{tab:2d_sim_eval_noosi}
        \begin{tabular}{|l|rrrr|}
			\hline
			 & \textcolor{red}{Best} & \textcolor{mydarkgreen}{2nd Best} & \textcolor{blue}{3rd Best} & n.c. \\
			\hline
			DRSAC & 27 & 5 & 4 & 398 \\
			SIRSA & 0 & 0 & 0 & 434 \\
			cMDSAC (Ours) & 131 & 107 & 100 & 96 \\
			eMDSAC (Ours) & 8 & 26 & 33 & 367 \\
			uMD-SIRSA (Ours) & 14 & 52 & 75 & 293 \\
			uMDSAC-v1 (Ours) & 106 & 132 & 133 & 63 \\
			uMDSAC-v2 (Ours) & 148 & 112 & 89 & 85 \\
			\hline
		\end{tabular}
    \end{minipage}%
    \begin{minipage}[c]{0.5\linewidth}\centering
        \subcaption{Dim $=$ 2 $-$ Uncertainty $=$ OSI}
        \label{tab:2d_sim_eval_osi}
        \begin{tabular}{|l|rrrr|}
			\hline
			 & \textcolor{red}{Best} & \textcolor{mydarkgreen}{2nd Best} & \textcolor{blue}{3rd Best} & n.c. \\
			\hline
			DRSAC & 37 & 2 & 12 & 383 \\
			SIRSA & 26 & 13 & 2 & 393 \\
			cMDSAC (Ours) & 110 & 169 & 68 & 87 \\
			eMDSAC (Ours) & 1 & 15 & 44 & 374 \\
			uMD-SIRSA (Ours) & 14 & 41 & 65 & 314 \\
			uMDSAC-v1 (Ours) & 61 & 89 & 196 & 88 \\
			uMDSAC-v2 (Ours) & 185 & 105 & 47 & 97 \\
			\hline
		\end{tabular}
    \end{minipage}
    \begin{minipage}[c]{0.5\linewidth}\centering
        \subcaption{Dim $=$ 22 $-$ Uncertainty $=$ Fixed}
        \label{tab:multid_sim_eval_noosi}
        \begin{tabular}{|l|rrrr|}
			\hline
			 & \textcolor{red}{Best} & \textcolor{mydarkgreen}{2nd Best} & \textcolor{blue}{3rd Best} & n.c. \\
			\hline
			DRSAC & 58 & 52 & 65 & 259 \\
			SIRSA & 0 & 0 & 0 & 434 \\
			cMDSAC (Ours) & 42 & 63 & 78 & 251 \\
			eMDSAC (Ours) & 184 & 105 & 66 & 79 \\
			uMD-SIRSA (Ours) & 37 & 60 & 87 & 250 \\
			uMDSAC-v1 (Ours) & 48 & 64 & 70 & 252 \\
			uMDSAC-v2 (Ours) & 65 & 90 & 68 & 211 \\
			\hline
		\end{tabular}
    \end{minipage}%
    \begin{minipage}[c]{0.5\linewidth}\centering
        \subcaption{Dim $=$ 22 $-$ Uncertainty $=$ OSI}
        \label{tab:multid_sim_eval_osi}
        \begin{tabular}{|l|rrrr|}
			\hline
			 & \textcolor{red}{Best} & \textcolor{mydarkgreen}{2nd Best} & \textcolor{blue}{3rd Best} & n.c. \\
			\hline
			DRSAC & 55 & 56 & 54 & 269 \\
			SIRSA & 0 & 0 & 0 & 434 \\
			cMDSAC (Ours) & 48 & 88 & 83 & 215 \\
			eMDSAC (Ours) & 222 & 96 & 60 & 56 \\
			uMD-SIRSA (Ours) & 24 & 53 & 81 & 276 \\
			uMDSAC-v1 (Ours) & 49 & 73 & 75 & 237 \\
			uMDSAC-v2 (Ours) & 36 & 68 & 81 & 249 \\
			\hline
		\end{tabular}
    \end{minipage}
    \caption{Performance quantification (i.e. number of environments on which each algorithm's performance fall in the four possibilities: best, second best, third best and ``n.c.") of the Sim-to-Sim evaluation on the DClawTurnFixed-v0 simulation on 434 simulation parameters sampled randomly from the full domain range. ``n.c." is short for ``not classified" and refers to values that fall outside the top three (the smaller the number in this column and the larger the number in the ``Best" column, the better the algorithm is).}
    \label{tab:sim2sim_evals}
\end{table*}
\begin{table*}
\captionsetup{justification=centering}
\centering
\begin{tabular}{|l|l|c|c|}
\hline
 Dim. & Algorithm & Number of perf. decrease & Number of perf. increase \\
 \hline
\multirow[c]{6}{*}{2}
 & \textbf{SIRSA} & \textbf{93} & \textbf{341} \\
 & \textbf{cMDSAC (Ours)} & \textbf{199} & \textbf{235} \\
 & eMDSAC (Ours) & 233 & 201 \\
 & uMD-SIRSA (Ours) & 281 & 153 \\
 & uMDSAC-v1 (Ours) & 275 & 159 \\
 & \textbf{uMDSAC-v2 (Ours)} & \textbf{200} & \textbf{234} \\
 \hline
\multirow[c]{6}{*}{22}
 & \textbf{SIRSA} & \textbf{0} & \textbf{434} \\
 & \textbf{cMDSAC (Ours)} & \textbf{190} & \textbf{244} \\
 & \textbf{eMDSAC (Ours)} & \textbf{189} & \textbf{245} \\
 & uMD-SIRSA (Ours) & 230 & 204 \\
 & \textbf{uMDSAC-v1 (Ours)} & \textbf{208} & \textbf{226} \\
 & uMDSAC-v2 (Ours) & 265 & 169 \\
 \hline
\end{tabular}
\caption{Number of simulation paramaters on which the OSI produced an increase or decrease in performance of the Sim-to-Sim evaluation on the DClawTurnFixed-v0 simulation on 434 simulation parameters sampled randomly from the full domain range. A higher number of increase (bold rows) shows that the corresponding algorithm benefits from the use of an OSI.}
    \label{tab:sim2sim_eval_tendency}
\end{table*}
 
 Despite that, \emph{uMDSAC-v1} appears to be overall the best algorithm for this sim-to-real evaluation (this can be seen in Table~\ref{tab:real_evals} by looking at the most colorful row which corresponds to the algorithm that most often ended up on the podium), and in general we can say that the MDSAC algorithms are better algorithms compared to \emph{DRSAC} and \emph{SIRSA} for bridging the transfer gap.

\section{Discussion}

 Although the above simulations and experiments display the advantage of learning the CCS with the MDSAC algorithms in order to solve Multi-Domain problems, we discuss below the limitations and possible extensions to the current work.

\subsection{Different types of domains and tasks}

 As the reader will have noticed, in this work, we have considered continuous states and actions tasks, with continuous and dense reward functions. However, in the broader range of RL applications, there exists different types of domains, ranging from sparse or discrete rewards to image inputs, each coming with its own set of challenges. In principle, the algorithms from the PMOMDP framework are not restricted by the kind of task at hand, since PMOMDP is an extension of the classical MDP and therefore inherits many of the characteristics of typical MDP RL algorithms. Hence, for each type of tasks, the PMOMDP framework can be applied to extend the already existing and specialized algorithms in order to obtain new uncertainty-aware methods specifically suited for the control problem to solve. Nevertheless, we note that PMOMDP is also bequeathed with many of the challenges of conventional MDP RL algorithms, with the added overhead that it must learn a set of policies instead of only one, which can accentuate sample-efficiency-related issues. For example, in tasks with discrete reward signals, issues such as slow convergence are usually encountered, requiring more complex network structures in order to overcome the discrete nature of the reward landscape. In such a case, the necessity for uncertainty-aware networks can be expected to further heighten the level of complexity required by the task.

 Despite that, we believe that the ability to learn a universal policy that can adapt to different target environments while being able to act safely and conservatively during the first stages of interaction and without the need to be retrained each time the level of knowledge on the system is updated represents a formidable advantage that overshadows the previously mentioned overheads.

\subsection{Performance improvements}

 The simulations performed above revealed the advantages and disadvantages of the main proposed algorithms (notably the superiority of the \emph{uMDSAC} algorithmic strategies). However, improvements are still possible. Indeed, given that these policies are intended to be deployed without any additional retraining, a straightforward way to improve their performance is to train for longer time in simulations. This is because the algorithms developed here intend on learning a set of domain randomization policies, and simulators are attractive because they offer sandbox environments where an unlimited amount of data can be safely generated in order to deal with the sample inefficiency of RL algorithms. Since our algorithms learn a set of policies instead of only one, it is natural that they would require a larger number of samples in order to properly cover the entire CCS.
 
 Furthermore, instead of the linear utility function, a scalarization function that accounts for the worst case performance over the uncertainty set may lead to better performing policies. Indeed, it is well known that linear scalarizations techniques, despite their simplicity, are unable to find policies that lie in a non-convex region of the Pareto front due to their reliance on the convex combination. It would therefore be interesting to investigate the diverse utility functions of the MORL litterature~\cite{van2014multi} in order to find which one leads to better sim-to-real transfer performances. In addition, as pointed out in \cite{mehta2020activedr}, the uniform sampling strategy may not be the most efficient one when dealing with domain randomization. Hence, it is of interest to study the integration of the Pseudo-MOMDP framework with DR methods such as active domain randomization.
 
Finally, a third possible extension lies in the production of uncertainty-aware policies with actions specifically designed to reduce the system identification uncertainty. For example, in the SAC case, one can consider adding another term associated with the uncertainty to the reward function:
\begin{align*}
 \begin{split}
 \tilde{r}(s, a, \kappa, \varpi) &= r(s, a, \kappa) - \alpha \ln \pi(a|s, \varpi) \\
 &+ \beta \ln p(\kappa | s, a, s', \varpi)
 \end{split}
\end{align*}
 with the term $\alpha \ln \pi(a|s, \varpi)$ being related to the entropy maximization of the typical SAC algorithm and $\beta \ln p(\kappa | s, a, s', \varpi)$ being a new term that encourages the policy to take actions that reduce the uncertainty (maximize the likelihood of the true parameter $\kappa$) of the probabilistic system identifier $p(\cdot | s, a, s', \varpi)$. We can notice that the efficiency of this mechanism will depend on the accuracy and stability of the stochastic OSI $p(\cdot | s, a, s', \varpi)$. Hence, this problem requires the development of a theoretically and experimentally stable and accurate OSI model. This requirement is also shared by the proposed methods discussed in this paper, since although the PMOMDP algorithms allow the training of uncertainty-aware policies, their efficiency once deployed depends on the accuracy of the system identifier. Hence, a mechanism that ensures the stability and accuracy of the identifier would benefit both the proposed algorithms and potential uncertainty-reducing policies.

\subsection{Handling environments with noisy system processes}

 In this work, we focused on the basic definition of the domain randomization problem where we have uncertainty over the parameters of the simulation model. In that approach, the scalarization assumes that the parameters do not change in the subsequent steps. Indeed, typical DR algorithms (including our proposals) learn the $Q$-value function as:
\begin{align*}
\begin{split}
&Q^{\pi}\left( s, a, \boldsymbol{\omega} \right) = \sum_{i=1}^{n} w_{i} \int\int \mathcal{P}_{i}(s', r | s, a) \left[ r  + \right.\\
&\left.\;\;\;\;\;\;\;\;\;\;\;\;\;\;\;\;\;\;\;\;\;\;\;\;\;\;\;\;\;\;\;\;\;\;\;\;\;\;\;\;\;\;\;\;\;\;\;\;\;\;\;\; \gamma V^{\pi}_{i}(s', \boldsymbol{\omega}) \right] ds' dr
\end{split} \\
\begin{split}
&V^{\pi}_{i}(s', \boldsymbol{\omega}) = \int \int\int \pi(a'|s', \boldsymbol{\omega}) \mathcal{P}_{i}(s'', r' | s', a') \left[ r' + \right.\\
&\left.\;\;\;\;\;\;\;\;\;\;\;\;\;\;\;\;\;\;\;\;\;\;\;\;\;\;\;\;\;\;\;\;\;\;\;\;\;\;\;\;\;\;\;\;\;\;\;\;\;\;\;\; \gamma V^{\pi}_{i}(s'', \boldsymbol{\omega}) \right] ds'' dr' da'
\end{split}
\end{align*}
That is, all subsequent states are assumed to be generated from the same environment and same transition model $\mathcal{P}_{i}$. With $s=s_{0}$, this allows the DR algorithm to associate particular states with specific environments, since certain states may never be reached in some environments.

 However, a different approach is to fully integrate the system uncertainty into the transition model, i.e.:
\begin{align*}
\begin{split}
&Q^{\pi}\left( s, a, \boldsymbol{\omega} \right) = \sum_{i=1}^{n} w_{i} \int\int \mathcal{P}_{i}(s', r | s, a) \big[ r + \\
&\;\;\;\;\;\;\;\;\;\;\;\;\;\;\;\;\;\;\;\;\;\;\;\;\;\;\;\;\;\;\;\;\;\;\;\;\;\;\;\;\;\;\; \gamma \sum_{j=1}^{n} w_{j} V^{\pi}_{j}(s', \boldsymbol{\omega}) \big] ds' dr
\end{split} \\
\begin{split}
&V^{\pi}_{j}(s', \boldsymbol{\omega}) = \int\int\int \pi(a'|s', \boldsymbol{\omega}) \mathcal{P}_{j}(s'', r' | s', a') \big[ r' + \\
&\;\;\;\;\;\;\;\;\;\;\;\;\;\;\;\;\;\;\;\;\;\;\;\;\;\;\;\;\;\;\;\;\;\;\;\;\;\;\;\;\;\;\; \gamma \sum_{k=1}^{n} w_{k} V^{\pi}_{k}(s'', \boldsymbol{\omega}) \big] ds'' dr' da'
\end{split}
\end{align*}
 This is equivalent to assuming that each next state is only generated with probability $w_{i}$ by the model $\mathcal{P}_{i}$. As long as there exists $w_{j} \neq w_{i}$ such that $w_{j} \neq 0$ for $i \neq j$, then the environment is similar to a system with noisy process or random disturbances, captured by the uncertainty $\boldsymbol{\omega} = \left[ w_{1}, \cdots, w_{n} \right]$, $\sum_{i} w_{i} = 1$. In this case, $\boldsymbol{\omega}$ no longer just captures our current uncertainty on the model misspecification, but also the uncertainty related to the random process underpinning the system.
 
 It is obvious that in such a case, a classical DR approach would be even more conservative than in the first case since it would become impossible to make associations between states and parameters. Uncertainty-Aware policies such as those generated by our proposals may therefore be particularly suited for such sim-to-real transfer gaps. It is thus compelling to investigate the application of the PMOMDP to these challenging cases.

\section{Related works}

As mentioned in the introductory section, in order to overcome the conservative tendency of DR techniques, diverse strategies have appeared in the RL literature (Table~\ref{tab:prev_works_pmomdp_comp} summarizes the approaches found in previous works).

\subsection{Adaptive domain randomization with probabilistic system identification:}

 Within this class of methods, we can first mention \cite{chebotar2019simopt} which advocated training policies across a spectrum of simulators, aligning their parameters with real-world data. Their method oscillates between optimizing policies under the DR distribution and refining the distribution by reducing the gap between simulated and real-world trajectories.

 Similarly,~\cite{ramos2019bayessim} employed a Likelihood-Free Inference (LFI) technique to compute the posterior distribution of simulation parameters using real-world trajectories.
Following this work,~\cite{matl2020inferring} applied Ramos et al.'s Bayesian inference process to simulate granular media, estimating factors like friction and restitution coefficients, effectively scaling the Bayesian inference procedure. The same year,~\cite{barcelos2020disco} introduced a method that combines DR, LFI, and policy optimization. They update the controller through nonlinear Model Predictive Control (MPC) and utilize the unscented transform to simulate various domain instances during the control horizon. This algorithm dynamically adjusts uncertainty as the system progresses over time, assigning higher costs to more uncertain trajectories. Their approach was validated on a simulated inverted pendulum swing-up task as well as a real trajectory following task using a wheeled robot.

 On the same LFI technique for probabilistic system identification,~\cite{muratore2022neuraldr} introduced a sequential neural posterior estimation algorithm based on normalizing flows to estimate the conditional posterior across the simulator parameters and illustrated the efficacy of this neural inference method in learning posterior beliefs over contact-rich black-box simulations. Furthermore, they evaluated this approach with policy optimization for an underactuated swing-up and balancing task, showcasing superior performance compared to BayesSim by~\cite{ramos2019bayessim} and Bayesian linear regression.

 \cite{rajeswaran2016epopt} proposed a related approach for robust policy learning by using an ensemble of simulated source domains and a form of adversarial training to learn policies that are robust and generalize to a broad range of possible target domains, including unmodeled effects. Their method involves adapting the probability distribution across source domains within the ensemble using data from the target domain. This adaptation process leverages approximate Bayesian methods to iteratively refine the distribution, gradually improving its approximation.

Unfortunately and as mentioned previously, all these previous methods involve retraining the policy each time the DR uncertainty set is updated by the probabilistic system identification, which limits their applicability to specific target domains.

\subsection{Universal policy networks with point-estimate-based system identification:}

Another class of methods focus on policies conditioned on context variables, representing unobserved physical properties that vary within the simulator and allowing for more flexibility in the control policy adaptation. For example,~\cite{chen2018hardware} suggested learning a policy conditioned on the robot hardware properties, reflecting variations in environment dynamics (e.g., friction and mass) drawn from a fixed simulator distribution. When explicit, this context matches the simulator parameters and when implicit, the mapping between context vectors and simulator environments is learned via policy optimization during training.

 Similarly,~\cite{yu2018policy} proposed training policies conditioned on simulator parameters, optimizing the context vector to enhance performance at test time. This method uses population-based gradient-free search to optimize the context vector for maximizing task performance.
 \cite{rakelly2019efficient} also employed context-conditioned policies, implicitly encoding context into a vector $z$. During training, their method improves policy performance while learning a probabilistic mapping from trajectory data to context vectors. At test time, the learned mapping infers the context vector online. Similarly,~\cite{yu2017uposi} and~\cite{ding2021not} combined Universal Policies with Online System Identification (OSI) for (respectively explicit and implicit) context inference during deployment.

 Unfortunately, these methods rely on deterministic system identification and assumes that the context can be accurately identified when deployed on the real system.

\subsection{Uncertainty-aware universal policy networks:}
 
 Among methods that consider probabilistic system identification and promote uncertainty awareness,~\cite{xie2022robust} formulated a multi-set robustness problem to learn a policy conditioned on the system uncertainty and robust to different perturbation sets. In their method, an OSI is trained in conjunction with the uncertainty-aware universal policy in order to learn to reduce uncertainty where possible given a few interactions, while allowing the policy to still act conservatively with respect to the remaining uncertainty. Their approach relies on the identifiability concept and therefore trains the UPN to be less conservative with respect to easily identifiable parameters. However, even if some parameters can be quickly identified in simulation, the noisiness of real world applications means that these same parameters can remain unidentifiable when the OSI encounters real data.

 Another recent work, worth mentioning is the method proposed by~\cite{semage2022UncAPS}. Indeed, they proposed to extend the work of~\cite{yu2017uposi} by first storing various task-specific policies using Universal Policy Networks (UPN) from simulations that cover a wide range of environmental factors and then, using the Unscented Bayesian optimization method to create strong policies for specific environments by blending related UPN policies in a way that considers both aleatoric and epistemic uncertainty arising from the Bayesian system identification process. Their approach however still results in point estimates of the parameters of the system and the system noise scale is assumed known in advance.

\begin{table*}[tb]
    \captionsetup{width=\textwidth, justification=centering}
    \caption{Succinct summary of previous approaches, and their correspondence within the new PMOMDP framework.}
    \label{tab:prev_works_pmomdp_comp}
    \centering
    \begin{tabular}{|p{2cm}|p{7cm}|p{5cm}|p{2cm}|}
		\hline
		Approach & Optimal policy and critic & Caracteristics & Equivalence within PMOMDP
		\\
		\hline\hline
		DR & \begin{itemize}
			\item $\pi^{*}(\cdot | s) = \argmax{\pi}{ \mathbb{E}_{\pi}[Q(s, a)] }$
			\item $(\mathcal{T}Q)(s, a) = r_{\kappa} + \gamma V^{\pi}(s')$, $\forall \kappa\sim\varpi$
		\end{itemize} & \begin{itemize}
			\item Fixed prior uncertainty level $\varpi$
		\end{itemize} & N/A \\
		\hline
		DR + Probabilistic SI (\cite{ramos2019bayessim}, \cite{rajeswaran2016epopt}, etc.) & \begin{itemize}
			\item Set initial $\varpi$
			\item Solve $\pi^{*}(\cdot | s) = \argmax{\pi}{ \mathbb{E}_{\pi}[Q(s, a)] }$
			\item with $(\mathcal{T}Q)(s, a) = r_{\kappa} + \gamma V^{\pi}(s')$, $\forall \kappa\sim\varpi$
			\item Update from $\varpi$ to $\varpi'$
			\item Solve $\pi^{*}(\cdot | s) = \argmax{\pi}{ \mathbb{E}_{\pi}[Q(s, a)] }$
			\item with $(\mathcal{T}Q)(s, a) = r_{\kappa} + \gamma V^{\pi}(s')$, $\forall \kappa\sim\varpi'$
		\end{itemize} & \begin{itemize}
			\item Adapts to different levels of uncertainty
			\item Retrain policy after uncertainty update
		\end{itemize} & Online Scalarized MORL \\
		\hline
		UP + Deterministic SI (\cite{yu2017uposi, ding2021not}) & \begin{itemize}
			\item $\pi^{*}(\cdot | s, \kappa) = \argmax{\pi}{ \mathbb{E}_{\pi}[Q(s, a, \kappa)] }$
			\item $(\mathcal{T}Q)(s, a, \kappa) = r_{\kappa} + \gamma V^{\pi}(s', \kappa)$, $\forall \kappa\sim\varpi$
		\end{itemize} & \begin{itemize}
			\item Non-conservative policies
			\item No uncertainty-awareness
		\end{itemize} & Conditioned MORL with degenerate probability distributions \\
		\hline
		 Multi-Set Robustness (\cite{xie2022robust}) & \begin{itemize}
			\item $\pi^{*}(\cdot | s, \varpi) = \argmax{\pi}{ \mathbb{E}_{\pi}[ \mathbb{E}_{\kappa\sim\varpi}[ Q(s, a, \kappa)] ] }$
			\item $(\mathcal{T}Q)(s, a, \kappa) = r_{\kappa} + \gamma V^{\pi_{\varpi'}}(s', \kappa)$, $\forall \kappa\sim\varpi$
			\item with $\varpi' = \mathrm{OSI}(s, a, s', \varpi)$
		\end{itemize} & \begin{itemize}
			\item Adapts to different levels of uncertainty
			\item Uncertainty decreased when possible with OSI
			\item Discrete number of uncertainty sets making it dependent on the random seed
		\end{itemize} & N/A \\
        \hline
    \end{tabular}
\end{table*}

\begin{table*}[tb]
    \captionsetup{width=\textwidth, justification=centering}
    \caption{Succinct summary of the algorithms derived in this paper, and their correspondence within the new PMOMDP framework.}
    \label{tab:pmomdp_comp}
    \centering
    \begin{tabular}{|p{2cm}|p{7cm}|p{5cm}|p{2cm}|}
		\hline
		Approach & Optimal policy and critic & Caracteristics & Equivalence within PMOMDP
		\\
		\hline\hline
		\textbf{cMDRL} & \begin{itemize}
			\item $\pi^{*}(\cdot | s, \varpi) = \argmax{\pi}{ \mathbb{E}_{\pi}[ \mathbb{E}_{\kappa\sim\varpi}[ Q(s, a, \kappa, \varpi)] ] }$
			\item $(\mathcal{T}Q)(s, a, \kappa, \varpi) = r_{\kappa} + \gamma V^{\pi_{\varpi}}(s', \kappa, \varpi)$, $\forall \kappa\sim\varpi$
		\end{itemize} & \begin{itemize}
			\item Adapts to different levels of uncertainty
			\item Learns a specific critic $Q_{\varpi}$ for each uncertainty-aware policy $\pi_{\varpi}$
			\item Stable, but not guaranteed to fully acquire the optimal CCS
		\end{itemize} & Conditioned MORL \\
    	\hline
		\textbf{eMDRL} & \begin{itemize}
			\item $\pi^{*}(\cdot | s, \varpi) = \argmax{\pi}{ \mathbb{E}_{\pi}[ \mathbb{E}_{\kappa\sim\varpi}[ Q(s, a, \kappa, \varpi^{*})] ] }$
			\item $(\mathcal{T}Q)(s, a, \kappa, \varpi) = r_{\kappa} + \gamma V^{\pi_{\varpi}}(s', \kappa, \varpi)$, $\forall \kappa\sim\varpi$
			\item with $\varpi^{*} = \argmax{\varpi'}{ \mathbb{E}_{\pi_{\varpi'}}[ \mathbb{E}_{\kappa\sim\varpi}[ Q(s, a, \kappa, \varpi')] ] }$
		\end{itemize} & \begin{itemize}
			\item Adapts to different levels of uncertainty
			\item Learns a specific critic $Q_{\varpi}$ for each uncertainty-aware policy $\pi_{\varpi}$
			\item Can be unstable, but theoretically derived to fully acquire the optimal CCS
		\end{itemize} & Envelope MORL \\
    	\hline
		\textbf{uMDRL} & \begin{itemize}
			\item $\pi^{*}(\cdot | s, \varpi) = \argmax{\pi}{ \mathbb{E}_{\pi}[ \mathbb{E}_{\kappa\sim\varpi}[ Q(s, a, \kappa)] ] }$
			\item $(\mathcal{T}Q)(s, a, \kappa) = r_{\kappa} + \gamma V^{\pi_{\delta_{\kappa}}}(s', \kappa)$, $\forall \kappa\sim\varpi$
		\end{itemize} & \begin{itemize}
			\item Adapts to different levels of uncertainty
			\item Learns a specific critic $Q_{\delta_{\kappa}}$ for each domain-aware policy $\pi_{\delta_{\kappa}}$
			\item Stable and theoretically derived to fully acquire the optimal CCS 
		\end{itemize} & Utopia-based MORL \\
        \hline
    \end{tabular}
\end{table*}

\section{Conclusion}

 In this paper, we tackle the problem of learning efficient policies when there is uncertainty on the target system by introducing a novel reinforcement learning framework from which we derived three main types of algorithms (Conditioned, Envelope-based and Utopia-based) that equip policies with domain uncertainty awareness. This framework, based on the analogy between multi-objective RL and the multi-domain RL problem arising from domain randomization, offers the ability to adapt diverse algorithms from the MORL literature, in order to learn the convex coverage set in the policy optimization of uncertainty-aware policies. These algorithms are able to generate policies that can adapt to different environments and system uncertainties, facilitating efficient transitions from simulated environments to real-world application.
 
 In this work, while we focused on the linear utility function, the rich literature on MORL
offers various other scalarization functions that exploits, captures and reveals different regions of the Pareto front. It is therefore of interest to study, under the PMOMDP framework, the best scalarization function for sim-to-real transfer of uncertainty-aware policies. In addition, as revealed by the experiments, each type of algorithm has its own strength and weaknesses making it impossible for a single one of them to always be the best on every environment or uncertainty subset. A strategy that combines all three types of algorithms (Conditioned, Envelope-based and Utopia-based) into a single algorithm that can interpolate automatically between them is a natural extension to this work worth investigating.

\begin{acks}
This work was supported by JSPS KAKENHI, Grant-in-Aid for Scientific Research (B), Grant Number JP20H04265.
\end{acks}

\bibliographystyle{SageH}
\bibliography{biblio_mdrl}

\begin{thebibliography}{51}
\providecommand{\natexlab}[1]{#1}
\providecommand{\url}[1]{\texttt{#1}}
\providecommand{\urlprefix}{URL }
\expandafter\ifx\csname urlstyle\endcsname\relax
  \providecommand{\doi}[1]{DOI:\discretionary{}{}{}#1}\else
  \providecommand{\doi}{DOI:\discretionary{}{}{}\begingroup
  \urlstyle{rm}\Url}\fi

\bibitem[{Abels et~al.(2019)Abels, Roijers, Lenaerts, Now{\'e} and
  Steckelmacher}]{abels2019cMORL}
Abels A, Roijers D, Lenaerts T, Now{\'e} A and Steckelmacher D (2019) Dynamic
  weights in multi-objective deep reinforcement learning.
\newblock In: \emph{International Conference on Machine Learning}. PMLR, pp.
  11--20.

\bibitem[{Ahn et~al.(2020)Ahn, Zhu, Hartikainen, Ponte, Gupta, Levine and
  Kumar}]{ahn2020robel}
Ahn M, Zhu H, Hartikainen K, Ponte H, Gupta A, Levine S and Kumar V (2020)
  Robel: Robotics benchmarks for learning with low-cost robots.
\newblock In: \emph{Conference on Robot Learning}. PMLR, pp. 1300--1313.

\bibitem[{Andrychowicz et~al.(2020)Andrychowicz, Baker, Chociej, Jozefowicz,
  McGrew, Pachocki, Petron, Plappert, Powell, Ray
  et~al.}]{andrychowicz2020learning}
Andrychowicz OM, Baker B, Chociej M, Jozefowicz R, McGrew B, Pachocki J, Petron
  A, Plappert M, Powell G, Ray A et~al. (2020) Learning dexterous in-hand
  manipulation.
\newblock \emph{The International Journal of Robotics Research} 39(1): 3--20.

\bibitem[{Barcelos et~al.(2020)Barcelos, Oliveira, Possas, Ott and
  Ramos}]{barcelos2020disco}
Barcelos L, Oliveira R, Possas R, Ott L and Ramos F (2020) Disco: Double
  likelihood-free inference stochastic control.
\newblock In: \emph{2020 IEEE International Conference on Robotics and
  Automation (ICRA)}. IEEE, pp. 10969--10975.

\bibitem[{Chebotar et~al.(2019)Chebotar, Handa, Makoviychuk, Macklin, Issac,
  Ratliff and Fox}]{chebotar2019simopt}
Chebotar Y, Handa A, Makoviychuk V, Macklin M, Issac J, Ratliff N and Fox D
  (2019) Closing the sim-to-real loop: Adapting simulation randomization with
  real world experience.
\newblock In: \emph{2019 International Conference on Robotics and Automation
  (ICRA)}. IEEE, pp. 8973--8979.

\bibitem[{Chen et~al.(2018)Chen, Murali and Gupta}]{chen2018hardware}
Chen T, Murali A and Gupta A (2018) Hardware conditioned policies for
  multi-robot transfer learning.
\newblock \emph{Advances in Neural Information Processing Systems} 31.

\bibitem[{Chen et~al.(2021)Chen, Hu, Jin, Li and
  Wang}]{chen2021understandingdr}
Chen X, Hu J, Jin C, Li L and Wang L (2021) Understanding domain randomization
  for sim-to-real transfer.
\newblock \emph{arXiv preprint arXiv:2110.03239} .

\bibitem[{Dehban et~al.(2019)Dehban, Borrego, Figueiredo, Moreno, Bernardino
  and Santos-Victor}]{dehban2019impact}
Dehban A, Borrego J, Figueiredo R, Moreno P, Bernardino A and Santos-Victor J
  (2019) The impact of domain randomization on object detection: A case study
  on parametric shapes and synthetic textures.
\newblock In: \emph{2019 IEEE/RSJ International Conference on Intelligent
  Robots and Systems (IROS)}. IEEE, pp. 2593--2600.

\bibitem[{Derman et~al.(2019)Derman, Mankowitz, Mann and Mannor}]{bayesian_rrl}
Derman E, Mankowitz D, Mann T and Mannor S (2019) A bayesian approach to robust
  reinforcement learning.
\newblock \emph{arXiv preprint arXiv:1905.08188} .

\bibitem[{Ding(2021)}]{ding2021not}
Ding Z (2021) Not only domain randomization: Universal policy with embedding
  system identification.
\newblock \emph{arXiv preprint arXiv:2109.13438} .

\bibitem[{Doersch and Zisserman(2019)}]{doersch2019sim2real}
Doersch C and Zisserman A (2019) Sim2real transfer learning for 3d human pose
  estimation: motion to the rescue.
\newblock \emph{Advances in Neural Information Processing Systems} 32.

\bibitem[{Farahani et~al.(2021)Farahani, Voghoei, Rasheed and
  Arabnia}]{da_review}
Farahani A, Voghoei S, Rasheed K and Arabnia HR (2021) A brief review of domain
  adaptation.
\newblock \emph{Advances in Data Science and Information Engineering} :
  877--894.

\bibitem[{Haarnoja et~al.(2018)Haarnoja, Zhou, Hartikainen, Tucker, Ha, Tan,
  Kumar, Zhu, Gupta, Abbeel et~al.}]{haarnoja2018soft}
Haarnoja T, Zhou A, Hartikainen K, Tucker G, Ha S, Tan J, Kumar V, Zhu H, Gupta
  A, Abbeel P et~al. (2018) Soft actor-critic algorithms and applications.
\newblock \emph{arXiv preprint arXiv:1812.05905} .

\bibitem[{Ilboudo et~al.(2023)Ilboudo, Kobayashi and
  Matsubara}]{ilboudo2023domains}
Ilboudo WEL, Kobayashi T and Matsubara T (2023) Domains as objectives:
  Multi-domain reinforcement learning with convex-coverage set learning for
  domain uncertainty awareness.
\newblock In: \emph{2023 IEEE/RSJ International Conference on Intelligent
  Robots and Systems (IROS)}. IEEE, pp. 5622--5629.

\bibitem[{Ilboudo et~al.(2020)Ilboudo, Kobayashi and
  Sugimoto}]{ilboudo2020robust}
Ilboudo WEL, Kobayashi T and Sugimoto K (2020) Robust stochastic gradient
  descent with student-t distribution based first-order momentum.
\newblock \emph{IEEE Transactions on Neural Networks and Learning Systems}
  33(3): 1324--1337.

\bibitem[{James et~al.(2018)James, Wohlhart, Kalakrishnan, Kalashnikov, Irpan,
  Ibarz, Levine, Hadsell and Bousmalis}]{rcan2019}
James S, Wohlhart P, Kalakrishnan M, Kalashnikov D, Irpan A, Ibarz J, Levine S,
  Hadsell R and Bousmalis K (2018) Sim-to-real via sim-to-sim: Data-efficient
  robotic grasping via randomized-to-canonical adaptation networks.
\newblock \emph{CoRR} abs/1812.07252.
\newblock \urlprefix\url{http://arxiv.org/abs/1812.07252}.

\bibitem[{Lin et~al.(2020)Lin, Thomas, Yang and Ma}]{mba_rl}
Lin Z, Thomas G, Yang G and Ma T (2020) Model-based adversarial
  meta-reinforcement learning.
\newblock \emph{arXiv preprint arXiv:2006.08875} .

\bibitem[{Ljung(2010)}]{si_perspectives2010}
Ljung L (2010) Perspectives on system identification.
\newblock \emph{Annual Reviews in Control} 34(1): 1--12.

\bibitem[{Mankowitz et~al.(2019)Mankowitz, Levine, Jeong, Shi, Kay,
  Abdolmaleki, Springenberg, Mann, Hester and Riedmiller}]{mmissspec_rrl}
Mankowitz DJ, Levine N, Jeong R, Shi Y, Kay J, Abdolmaleki A, Springenberg JT,
  Mann T, Hester T and Riedmiller M (2019) Robust reinforcement learning for
  continuous control with model misspecification.
\newblock \emph{arXiv preprint arXiv:1906.07516} .

\bibitem[{Matl et~al.(2020)Matl, Narang, Bajcsy, Ramos and
  Fox}]{matl2020inferring}
Matl C, Narang Y, Bajcsy R, Ramos F and Fox D (2020) Inferring the material
  properties of granular media for robotic tasks.
\newblock In: \emph{2020 ieee international conference on robotics and
  automation (icra)}. IEEE, pp. 2770--2777.

\bibitem[{Medeiros(2018)}]{medeiros2018unscented}
Medeiros JEGd (2018) Unscented transform framework for quantization modeling in
  data conversion systems .

\bibitem[{Mehta et~al.(2020)Mehta, Diaz, Golemo, Pal and
  Paull}]{mehta2020activedr}
Mehta B, Diaz M, Golemo F, Pal CJ and Paull L (2020) Active domain
  randomization.
\newblock In: \emph{Conference on Robot Learning}. PMLR, pp. 1162--1176.

\bibitem[{Morimoto and Doya(2005)}]{rrl2005}
Morimoto J and Doya K (2005) Robust reinforcement learning.
\newblock \emph{Neural computation} 17(2): 335--359.

\bibitem[{Mossalam et~al.(2016)Mossalam, Assael, Roijers and
  Whiteson}]{mossalam2016sMORL}
Mossalam H, Assael YM, Roijers DM and Whiteson S (2016) Multi-objective deep
  reinforcement learning.
\newblock \emph{arXiv preprint arXiv:1610.02707} .

\bibitem[{Mozian et~al.(2020)Mozian, Higuera, Meger and
  Dudek}]{mozian2020learning}
Mozian M, Higuera JCG, Meger D and Dudek G (2020) Learning domain randomization
  distributions for training robust locomotion policies.
\newblock In: \emph{2020 IEEE/RSJ International Conference on Intelligent
  Robots and Systems (IROS)}. IEEE, pp. 6112--6117.

\bibitem[{Muratore et~al.(2021)Muratore, Eilers, Gienger and
  Peters}]{muratore2021data}
Muratore F, Eilers C, Gienger M and Peters J (2021) Data-efficient domain
  randomization with bayesian optimization.
\newblock \emph{IEEE Robotics and Automation Letters} 6(2): 911--918.

\bibitem[{Muratore et~al.(2022{\natexlab{a}})Muratore, Gruner, Wiese, Belousov,
  Gienger and Peters}]{muratore2022neural}
Muratore F, Gruner T, Wiese F, Belousov B, Gienger M and Peters J
  (2022{\natexlab{a}}) Neural posterior domain randomization.
\newblock In: \emph{Conference on Robot Learning}. PMLR, pp. 1532--1542.

\bibitem[{Muratore et~al.(2022{\natexlab{b}})Muratore, Gruner, Wiese, Belousov,
  Gienger and Peters}]{muratore2022neuraldr}
Muratore F, Gruner T, Wiese F, Belousov B, Gienger M and Peters J
  (2022{\natexlab{b}}) Neural posterior domain randomization.
\newblock In: \emph{Conference on Robot Learning}. PMLR, pp. 1532--1542.

\bibitem[{Natarajan and Tadepalli(2005)}]{natarajan2005dynamicpreference}
Natarajan S and Tadepalli P (2005) Dynamic preferences in multi-criteria
  reinforcement learning.
\newblock In: \emph{Proceedings of the 22nd international conference on Machine
  learning}. pp. 601--608.

\bibitem[{Peng et~al.(2018)Peng, Andrychowicz, Zaremba and
  Abbeel}]{peng2018sim2real}
Peng XB, Andrychowicz M, Zaremba W and Abbeel P (2018) Sim-to-real transfer of
  robotic control with dynamics randomization.
\newblock In: \emph{2018 IEEE international conference on robotics and
  automation (ICRA)}. IEEE, pp. 3803--3810.

\bibitem[{Petrik and Russel(2019)}]{conservative_rrl2019}
Petrik M and Russel RH (2019) Beyond confidence regions: Tight bayesian
  ambiguity sets for robust mdps.
\newblock \emph{arXiv preprint arXiv:1902.07605}
  \urlprefix\url{http://arxiv.org/abs/1902.07605}.

\bibitem[{Pinto et~al.(2017)Pinto, Davidson, Sukthankar and Gupta}]{adv_rrl}
Pinto L, Davidson J, Sukthankar R and Gupta A (2017) Robust adversarial
  reinforcement learning.
\newblock \emph{arXiv preprint arXiv:1703.02702} .

\bibitem[{Rajeswaran et~al.(2016)Rajeswaran, Ghotra, Ravindran and
  Levine}]{rajeswaran2016epopt}
Rajeswaran A, Ghotra S, Ravindran B and Levine S (2016) Epopt: Learning robust
  neural network policies using model ensembles.
\newblock \emph{arXiv preprint arXiv:1610.01283} .

\bibitem[{Rakelly et~al.(2019)Rakelly, Zhou, Finn, Levine and
  Quillen}]{rakelly2019efficient}
Rakelly K, Zhou A, Finn C, Levine S and Quillen D (2019) Efficient off-policy
  meta-reinforcement learning via probabilistic context variables.
\newblock In: \emph{International conference on machine learning}. PMLR, pp.
  5331--5340.

\bibitem[{Ramos et~al.(2019)Ramos, Possas and Fox}]{ramos2019bayessim}
Ramos F, Possas RC and Fox D (2019) Bayessim: adaptive domain randomization via
  probabilistic inference for robotics simulators.
\newblock \emph{arXiv preprint arXiv:1906.01728} .

\bibitem[{Ruiz et~al.(2018)Ruiz, Schulter and Chandraker}]{ruiz2018guidedDR}
Ruiz N, Schulter S and Chandraker M (2018) Learning to simulate.
\newblock \doi{10.48550/ARXIV.1810.02513}.
\newblock \urlprefix\url{https://arxiv.org/abs/1810.02513}.

\bibitem[{Semage et~al.(2022)Semage, Karimpanal, Rana and
  Venkatesh}]{semage2022UncAPS}
Semage BL, Karimpanal TG, Rana S and Venkatesh S (2022) Uncertainty aware
  system identification with universal policies.
\newblock In: \emph{2022 26th International Conference on Pattern Recognition
  (ICPR)}. IEEE, pp. 2321--2327.

\bibitem[{Tesauro et~al.(2007)Tesauro, Das, Chan, Kephart, Levine, Rawson and
  Lefurgy}]{tesauro2007nonlinearscalarization}
Tesauro G, Das R, Chan H, Kephart J, Levine D, Rawson F and Lefurgy C (2007)
  Managing power consumption and performance of computing systems using
  reinforcement learning.
\newblock \emph{Advances in neural information processing systems} 20.

\bibitem[{Tessler et~al.(2019)Tessler, Efroni and Mannor}]{action_rrl}
Tessler C, Efroni Y and Mannor S (2019) Action robust reinforcement learning
  and applications in continuous control.
\newblock \emph{arXiv preprint arXiv:1901.09184} .

\bibitem[{Tobin et~al.(2017)Tobin, Fong, Ray, Schneider, Zaremba and
  Abbeel}]{dr2017Tobin}
Tobin J, Fong R, Ray A, Schneider J, Zaremba W and Abbeel P (2017) Domain
  randomization for transferring deep neural networks from simulation to the
  real world.
\newblock \emph{arXiv preprint arXiv:1703.06907}
  \doi{10.48550/ARXIV.1703.06907}.
\newblock \urlprefix\url{https://arxiv.org/abs/1703.06907}.

\bibitem[{Todorov et~al.(2012)Todorov, Erez and Tassa}]{todorov2012mujoco}
Todorov E, Erez T and Tassa Y (2012) Mujoco: A physics engine for model-based
  control.
\newblock In: \emph{2012 IEEE/RSJ International Conference on Intelligent
  Robots and Systems}. IEEE, pp. 5026--5033.
\newblock \doi{10.1109/IROS.2012.6386109}.

\bibitem[{Uhlmann(1995)}]{uhlmann1995dynamic}
Uhlmann J (1995) \emph{Dynamic map building and localization: new theoretical
  foundations.}
\newblock PhD Thesis, University of Oxford.

\bibitem[{Van~Moffaert et~al.(2013)Van~Moffaert, Drugan and
  Now{\'e}}]{van2013scalarized}
Van~Moffaert K, Drugan MM and Now{\'e} A (2013) Scalarized multi-objective
  reinforcement learning: Novel design techniques.
\newblock In: \emph{2013 IEEE symposium on adaptive dynamic programming and
  reinforcement learning (ADPRL)}. IEEE, pp. 191--199.

\bibitem[{Van~Moffaert and Now{\'e}(2014)}]{van2014multi}
Van~Moffaert K and Now{\'e} A (2014) Multi-objective reinforcement learning
  using sets of pareto dominating policies.
\newblock \emph{The Journal of Machine Learning Research} 15(1): 3483--3512.

\bibitem[{Volpi et~al.(2021)Volpi, Larlus and Rogez}]{volpi2021continual}
Volpi R, Larlus D and Rogez G (2021) Continual adaptation of visual
  representations via domain randomization and meta-learning.
\newblock In: \emph{Proceedings of the IEEE/CVF Conference on Computer Vision
  and Pattern Recognition}. pp. 4443--4453.

\bibitem[{Weng(2019)}]{weng2019DR}
Weng L (2019) Domain randomization for sim2real transfer.
\newblock \emph{lilianweng.github.io}
  \urlprefix\url{https://lilianweng.github.io/posts/2019-05-05-domain-randomization/}.

\bibitem[{Xie et~al.(2022)Xie, Sodhani, Finn, Pineau and Zhang}]{xie2022robust}
Xie A, Sodhani S, Finn C, Pineau J and Zhang A (2022) Robust policy learning
  over multiple uncertainty sets.
\newblock In: \emph{International Conference on Machine Learning}. PMLR, pp.
  24414--24429.

\bibitem[{Yang et~al.(2019)Yang, Sun and Narasimhan}]{yang2019generalizedMORL}
Yang R, Sun X and Narasimhan K (2019) A generalized algorithm for
  multi-objective reinforcement learning and policy adaptation.
\newblock \emph{arXiv preprint arXiv:1908.08342}
  \urlprefix\url{http://arxiv.org/abs/1908.08342}.

\bibitem[{Yang and Nguyen(2021)}]{yang2021recurrent}
Yang Z and Nguyen H (2021) Recurrent off-policy baselines for memory-based
  continuous control.
\newblock \emph{arXiv preprint arXiv:2110.12628} .

\bibitem[{Yu et~al.(2019)Yu, Liu and Turk}]{yu2018policy}
Yu W, Liu CK and Turk G (2019) Policy transfer with strategy optimization.
\newblock In: \emph{International Conference on Learning Representations}.
\newblock \urlprefix\url{https://openreview.net/forum?id=H1g6osRcFQ}.

\bibitem[{Yu et~al.(2017)Yu, Tan, Liu and Turk}]{yu2017uposi}
Yu W, Tan J, Liu CK and Turk G (2017) Preparing for the unknown: Learning a
  universal policy with online system identification.
\newblock \emph{arXiv preprint arXiv:1702.02453} .

\end{thebibliography}

\onecolumn

\clearpage
\section{Appendix: Pseudocodes} \label{apdx:pseudocodes}

\begin{minipage}{\textwidth}
\begin{algorithm}[H]
\caption{Common skeleton pseudo-code for all algorithms. The details per algorithm for each element $Q$, $\chi$, OSI, $y_{i}$, $L_{ij}$, $J$, $L$ and $L_{\chi}$ are given in Table~\ref{tab:algo_summary}}
\algcomment{$\varpi_{t+1} = \mathrm{OSI}(s_{t}, a_{t}, s_{t+1}, \varpi_{t}) = [ \underset{j=1,\cdots, k}{\mathrm{mean}}\left( f_{\Xi_{j}}(s_{t}, a_{t}, s_{t+1}, \varpi_{t}) \right), \underset{j=1,\cdots, k}{\mathrm{std}}\left( f_{\Xi_{j}}(s_{t}, a_{t}, s_{t+1}, \varpi_{t}) \right) ]$}
\begin{algorithmic}[1]
\State Define uncertainty set using the full domain uncertainty $\varpi_{\mathrm{full}}$ and uncertainty space $\Omega$
\State \textcolor{Red}{(For SIRSA algo)} Sample $B$ uncertainty subsets from $\Omega$ to build $\mathcal{D} = \{ \varpi_{b} \}_{b=1}^{B}$
\State Randomly initialize actor network $\pi_{\theta}$ and critic networks $Q_{\phi_1}$ and $Q_{\phi_2}$
\State Initialize target networks $\bar{Q}_{\bar{\phi}_1}$ and $\bar{Q}_{\bar{\phi}_2}$, with $\bar{\phi}_i = \phi_i$, $i\in\{ 1, 2 \}$
\State \textcolor{Red}{(For eMDSAC and uMDSAC-v2 algos)} \textcolor{blue}{Initialize solver network $\chi_{\psi}$}
\State \textcolor{Red}{(For SIRSA and MDSAC algos)} \textcolor{blue}{Randomly initialize OSI ensemble networks $\{ f_{\Xi_i} \}_{i=1}^{k}$ and} \textcolor{Red}{(for MDSAC algos)} \textcolor{blue}{dynamics network $D_{\psi}$}
\State Initialize experience replay memory $R$
\State Initialize steps counter $c \gets 0$
\State Initialize total steps number $M$ and number of initial warm-up steps $w$ (steps without training)
\While{$c < M$}
	\State Receive initial observation state $s_1$
	\State \textcolor{Red}{(For MDSAC algos)} \textcolor{blue}{Sample $\varpi$ from $\Omega$ and sample environment parameter $\kappa\sim\varpi$}
	\State \textcolor{Red}{(For SIRSA algo)} \textcolor{blue}{Sample $\varpi$ from $\mathcal{D}$ and sample environment parameter $\kappa\sim\varpi$}
	\State \textcolor{Red}{Set $\varpi_{1} = \varpi$ }
	\For{$t = 1$ to $T$}
		\If{$c \leq w$}
			\State Select action $a_t$ randomly
		\ElsIf{$c > w$}
			\State Select action $a_t \sim \pi_{\theta}(s_{t})$ according to current policy
		\EndIf
		\State Execute action $a_t$ and observe reward $r_t$ and new state $s_{t+1}$
		\State Store transition $\left( s_t, a_t, r_t, s_{t+1}, \textcolor{blue}{\kappa, \varpi} \right)$ in $R$
		\If{$c > w$}
			\State Sample a random minibatch of $N$ transitions $\left( s_i, a_i, r_i, s_{i+1}, \kappa_{i}, \varpi_{i} \right)$ from $R$
			\State Compute TD target $y_i$
			\State Update both critics by minimizing the loss $L_{ij}$, \quad $j\in\{ 1, 2 \}$
			\State Update actor by minimizing the loss $J$
			\State \textcolor{Red}{(For eMDSAC and uMDSAC-v2 algos)} \textcolor{blue}{Update solver by minimizing the loss $L_{\chi}$}
			\State \textcolor{Red}{(For SIRSA and MDSAC algos)} \textcolor{blue}{Update OSI ensemble and} \textcolor{Red}{(For MDSAC algos)} \textcolor{blue}{dynamics networks by minimizing the loss $L$}
			\State Update the target networks: $\bar{\phi}_{j} = \tau \phi_{j} + (1-\tau) \bar{\phi}_{j}$, \quad $j\in\{ 1, 2 \}$
		\EndIf
		\State \textcolor{Red}{(For SIRSA and MDSAC algos)} \textcolor{blue}{$\varpi_{t+1} = \mathrm{OSI}(s_{t}, a_{t}, s_{t+1}, \varpi_{t})$}
		\State $c \gets c + 1$
	\EndFor
\EndWhile
\end{algorithmic}
\end{algorithm}
\end{minipage}

\begin{landscape}
\begin{table*}[t]
  \captionsetup{width=\textwidth, justification=centering}
  \caption{Summary of algorithms involved in the experiments.}
  \label{tab:algo_summary}
  \centering
  \begin{adjustbox}{width=1.5\textwidth}
  \begin{tabular}{p{2.5cm}|c|c|c|c|c|c|c}
		\hline\hline
		 & DRSAC & SIRSA & cMDSAC (Ours) & eMDSAC (Ours) & uMD-SIRSA (Ours) & uMDSAC-v1 (Ours) & uMDSAC-v2 (Ours) \\
		\hline
		Q-Networks & $Q(s, a)$  & $Q(s, a, \kappa)$  & \multicolumn{2}{c}{$Q(s, a, \kappa, \varpi)$} & \multicolumn{3}{|c}{$Q(s, a, \kappa)$}  \\ &  &  &\multicolumn{2}{c}{} & \multicolumn{3}{|c}{} \\
		
		\hline
		Solver Network & $-$ & $-$ & $-$ & $\varpi^{*} = \chi(s, \varpi)$ & $-$ & - & $\varpi^{*} = \chi(s, \kappa)$  \\ &  &  &  &  &  &  &  \\

		\hline
		OSI-related Networks & $-$  & $\{ f_i(s, a, s', \varpi) \}_{i=1}^{k}$  & \multicolumn{5}{c}{$\{ f_i(s, a, s', \varpi) \}_{i=1}^{k}$} \\ &  &  & \multicolumn{5}{c}{$D(s'|s, a, \kappa)$}  \\ &  &  & \multicolumn{5}{c}{} \\

		\hline
		TD Target $y_{i} = $ & $r_i + \gamma \biggl( \underset{j=1,2}{ \min } \bar{Q}_{j}\left( s_{i+1}, a_{i+1} \right) \biggr)$ & $r_i + \gamma \biggl( \underset{j=1,2}{ \min } \bar{Q}_{j}\left( s_{i+1}, a_{i+1}, \textcolor{Red}{\kappa_{i}} \right) \biggr)$  & \multicolumn{2}{c}{$r_i + \gamma \biggl( \underset{j=1,2}{ \min } \bar{Q}_{j}\left( s_{i+1}, a_{i+1}, \textcolor{Red}{\kappa_{i}, \varpi_{i}} \right) \biggr)$} & \multicolumn{3}{|c}{$r_i + \gamma \biggl( \underset{j=1,2}{ \min } \bar{Q}_{j}\left( s_{i+1}, a_{i+1}, \textcolor{Red}{\kappa_{i}} \right) \biggr)$}  \\
		& $a_{i+1} \sim \pi(\cdot|s_{i+1})$ & $a_{i+1} \sim \pi(\cdot|s_{i+1}, \textcolor{Red}{ \tilde{\varpi}_{i+1} })$ & $a_{i+1} \sim \pi(\cdot|s_{i+1}, \textcolor{Red}{\varpi_{i}})$  & $a_{i+1} \sim \pi(\cdot|s_{i+1}, \textcolor{Red}{ \varpi^{*} })$ & $a_{i+1} \sim \pi(\cdot|s_{i+1}, \textcolor{Red}{ \tilde{\varpi}_{i+1} })$ & $a_{i+1} \sim \pi(\cdot|s_{i+1}, \textcolor{Red}{\delta_{\kappa_i}})$  & $a_{i+1} \sim \pi(\cdot|s_{i+1}, \textcolor{Red}{ \varpi^{*} })$  \\
		&  & \textcolor{Red}{ $\tilde{\varpi}_{i+1} = \mathrm{OSI}(s_{i}, a_{i}, s_{i+1}, \varpi_{i})$ } &  & \textcolor{Red}{ $\varpi^{*} = \chi(s_{i+1}, \varpi_{i})$ } & \textcolor{Red}{ $\tilde{\varpi}_{i+1} = \mathrm{OSI}(s_{i}, a_{i}, s_{i+1}, \varpi_{i})$ } &  & \textcolor{Red}{ $\varpi^{*} = \chi(s_{i+1}, \kappa_{i})$ }  \\ &  &  &  &  &  &  &  \\

		\hline
		Critic Loss $L_{ij} = $ & $\frac{1}{N} \sum\limits_{i=1}^N \left( y_i - Q_{j}(s_i, a_i) \right)^2$  & $\frac{1}{N} \sum\limits_{i=1}^N \left( y_i - Q_{j}(s_i, a_i, \textcolor{Red}{\kappa_{i}}) \right)^2$ & \multicolumn{2}{c}{$\frac{1}{N} \sum\limits_{i=1}^N \left( y_i - Q_{j}(s_i, a_i, \textcolor{Red}{\kappa_{i}, \varpi_{i}}) \right)^2$} & \multicolumn{3}{|c}{$\frac{1}{N} \sum\limits_{i=1}^N \left( y_i - Q_{j}(s_i, a_i, \textcolor{Red}{\kappa_{i}}) \right)^2$} \\ & $j\in\{ 1, 2 \}$  & $j\in\{ 1, 2 \}$  & \multicolumn{2}{c}{$j\in\{ 1, 2 \}$} & \multicolumn{3}{|c}{$j\in\{ 1, 2 \}$}  \\ &  &  & \multicolumn{2}{c}{} & \multicolumn{3}{|c}{} \\

		\hline
		Actor Loss $J = $ & $\frac{1}{N} \sum\limits_{i=1}^N - \biggl( \underset{j=1,2}{ \min } Q_{j}\left( s_{i}, \tilde{a}_{i} \right)$ & $\frac{1}{N} \sum\limits_{i=1}^N - \biggl( \textcolor{blue}{ \mathbb{E}_{\kappa\sim\textcolor{Red}{ \tilde{\varpi}_{i+1} }} \left[ \underset{j=1,2}{ \min } Q_{j}\left( s_{i}, \tilde{a}_{i}, \textcolor{Red}{ \kappa } \right) \right] }$  & \multicolumn{2}{c}{$\frac{1}{N} \sum\limits_{i=1}^N - \biggl( \textcolor{blue}{ \mathbb{E}_{\kappa\sim\varpi_{i}} \left[ \underset{j=1,2}{ \min } Q_{j}\left( s_{i}, \tilde{a}_{i}, \textcolor{Red}{ \kappa, \varpi_{i} } \right) \right] }$} & \multicolumn{3}{|c}{$\frac{1}{N} \sum\limits_{i=1}^N - \biggl( \textcolor{blue}{ \mathbb{E}_{\kappa\sim\varpi_{i}} \left[ \underset{j=1,2}{ \min } Q_{j}\left( s_{i}, \tilde{a}_{i}, \textcolor{Red}{ \kappa } \right) \right] }$} \\ 
		& $\;\;\;\;\;\;\;\;\;\;\;\;\;\;\;\; - \alpha \ln \pi(\tilde{a}_{i} | s_{i}) \biggr)$ & $\;\;\;\;\;\;\;\;\;\;\;\;\;\;\;\; - \alpha \ln \pi(\tilde{a}_{i} | s_{i}, \textcolor{red}{\tilde{\varpi}_{i+1}}) \biggr)$ & \multicolumn{2}{c}{$\;\;\;\;\;\;\;\;\;\;\;\;\;\;\;\; - \alpha \ln \pi(\tilde{a}_{i} | s_{i}, \textcolor{blue}{\varpi_{i}}) \biggr)$} & \multicolumn{3}{|c}{$ \;\;\;\;\;\;\;\;\;\;\;\;\;\;\;\; - \alpha \ln \pi(\tilde{a}_{i} | s_{i}, \textcolor{blue}{\varpi_{i}}) \biggr)$} \\
		& $\tilde{a}_{i} \sim \pi(\cdot|s_{i})$  & $\tilde{a}_{i} \sim \pi(\cdot|s_{i}, \textcolor{red}{\tilde{\varpi}_{i+1}})$  & \multicolumn{2}{c}{$\tilde{a}_{i} \sim \pi(\cdot|s_{i}, \textcolor{blue}{\varpi_{i}})$} & \multicolumn{3}{|c}{$\tilde{a}_{i} \sim \pi(\cdot|s_{i}, \textcolor{blue}{\varpi_{i}})$}  \\ &  & \textcolor{Red}{ $\tilde{\varpi}_{i+1} = \mathrm{OSI}(s_{i}, a_{i}, s_{i+1}, \varpi_{i})$ } &  \multicolumn{2}{c}{} & \multicolumn{3}{|c}{} \\ &  &  & \multicolumn{2}{c}{} & \multicolumn{3}{|c}{} \\

		\hline
		OSI Loss $L = $ & $-$ & $\frac{1}{N} \sum\limits_{i=1}^N \left[ \left( \kappa_{i} - f_{j}(s_{i}, a_{i}, s_{i+1}, \varpi_{i}) \right)^2 \right]_{j\sim\mathcal{U}(\{ 1, \cdots, k \})}$ & \multicolumn{5}{c}{$\frac{1}{N} \sum\limits_{i=1}^N \biggl( \left[ - \ln D(s_{i+1} | s_{i}, a_{i}, \kappa_{ji}) \right]_{j\sim\mathcal{U}(\{ 1, \cdots, k \})} + \ln \varpi_{i+1} - \ln \varpi_{i} \biggr)$}  \\ & & & \multicolumn{5}{c}{$\kappa_{ji} = f_{j}(s_{i}, a_{i}, s_{i+1}, \varpi_{i})$}  \\ &  &  & \multicolumn{5}{c}{$\varpi_{i+1} = [ \underset{j}{\mathrm{mean}}\left( \kappa_{ji} \right), \underset{j}{\mathrm{std}}\left( \kappa_{ji} \right) ]$} \\ & & $j\in\{ 1, \cdots, k \}$ & \multicolumn{5}{c}{$j\in\{ 1, \cdots, k \}$}  \\ &  &  & \multicolumn{5}{c}{} \\

		\hline
		Solver Loss $L_{\chi}$ & $-$ & $-$ & $-$ & $\frac{1}{N} \sum\limits_{i=1}^N - \biggl( \textcolor{blue}{ \mathbb{E}_{\kappa\sim\varpi_{i}} \left[ \underset{j=1,2}{ \min } Q_{j}\left( s_{i}, \tilde{a}_{i}, \textcolor{Red}{ \kappa, \varpi_{i} } \right) \right] }$ & $-$ & $-$ & $\frac{1}{N} \sum\limits_{i=1}^N - \biggl( \textcolor{blue}{ \underset{j=1,2}{ \min } Q_{j}\left( s_{i}, \tilde{a}_{i}, \textcolor{Red}{ \kappa_{i} } \right) }$ \\ & & & & $\;\;\;\;\;\;\;\;\;\;\;\;\;\;\;\; - \alpha \ln \pi(\tilde{a}_{i} | s_{i}, \textcolor{Red}{ \varpi^{*} }) \biggr)$ & & & $\;\;\;\;\;\;\;\;\;\;\;\;\;\;\;\; - \alpha \ln \pi(\tilde{a}_{i} | s_{i}, \textcolor{Red}{ \varpi^{*} }) \biggr)$  \\ 
		&  &  &  & $\tilde{a}_{i} \sim \pi(\cdot|s_{i}, \textcolor{Red}{ \varpi^{*} })$ &  &  & $\tilde{a}_{i} \sim \pi(\cdot|s_{i}, \textcolor{Red}{ \varpi^{*} })$  \\ 
		&  &  &  & \textcolor{Red}{ $\varpi^{*} = \chi(s_{i}, \varpi_{i})$ } &  &  & \textcolor{Red}{ $\varpi^{*} = \chi(s_{i}, \kappa_{i})$ }  \\ &  &  &  &  &  &  &  \\

		\hline
	\end{tabular}
	\end{adjustbox}
\end{table*}
\end{landscape}

\newpage
\section{Appendix: Unscented Transform} \label{apdx:UT_details}

 The Unscented Transform (UT), proposed by \cite{uhlmann1995dynamic}, is a method that uses a discrete set of sigma-points to capture the effects of nonlinear mappings on the statistical properties of a random variable. This approach is grounded in the idea that approximating a probability distribution is simpler than approximating a general nonlinear function or transformation and can be formally justified by considering the expectation of the Maclaurin series (see \cite{medeiros2018unscented}).
 
 Mathematically, the Unscented Transform can be defined as:
\begin{definition}[($\alpha$-th order Unscented Tranform)]
 The $\alpha$-th order Unscented Tranform of a continuous random variable $\textbf{x}$ with probability density function $p_{x}(\textbf{x})$ is a set of $n$ sigma-points and weights pairs, $\{ S_{i}, w_{i} \}$, characterizing a discrete probability distribution $F_{x}(S)$, such that:
\begin{align}
\mathbb{E}_{S\sim F_{x}} \left[ S^{k} \right] = \sum\limits_{i=1}^{n} S_{i}^{k} w_{i} = \int_{-\infty}^{+\infty} \textbf{x}^{k} p_{x}(\textbf{x}) d\textbf{x} = \mathbb{E}_{\textbf{x}\sim p_{x}} \left[ \textbf{x}^{k} \right] = m_{k}
\label{eq:ut_moments}
\end{align}
for $k = 0, 1, 2, \cdots, \alpha$, that is, both the discrete and the continuous distributions have the same moments up to the $\alpha$-th order.
\label{def:ut_definition}
\end{definition}

 Crucially, the UT is not unique and there exists many pairs $\{ S_{i}, w_{i} \}$ satisfying Definition \ref{def:ut_definition}. In this work, we automatically obtain $S_{i}$ by solving the linear form of equation \ref{eq:ut_moments} given by:
\begin{align}
\begin{bmatrix}
1 & 1 & \cdots & 1 \\
S_{1} & S_{2} & \cdots & S_{n} \\
S_{1}^{2} & S_{2}^{2} & \cdots & S_{n}^{2} \\
\cdot & \cdot & \cdots & \cdot \\
\cdot & \cdot & \cdots & \cdot \\
\cdot & \cdot & \cdots & \cdot \\
S_{1}^{\alpha} & S_{2}^{\alpha} & \cdots & S_{n}^{\alpha} \\
\end{bmatrix} 
\begin{bmatrix}
w_{1} \\
w_{2} \\
\cdot \\
\cdot \\
\cdot \\
w_{\alpha} \\
\end{bmatrix} &=
\begin{bmatrix}
1 \\
m_{1} \\
m_{2} \\
\cdot \\
\cdot \\
\cdot \\
m_{\alpha} \\
\end{bmatrix}
\end{align}
We set $n = 2d+1$, where $d$ is the dimension of the randomized parameter space and, since $\varpi$ is taken to be the continuous uniform distribution, we fix $w_{i} = 1/n$, $\forall i= 1, \cdots, n$. For the uniform distribution $\mathcal{U}([a, b])$, the $k$-th order moment is given by:
\begin{align*}
m_{k} = \frac{b^{k+1} - a^{k+1}}{(k+1)(b - a)}
\end{align*}

Using $\alpha = \max (10, d)$, we solve for $a=0$ and $b=1$ and then, given $\varpi = [\mu, \sigma]$, the corresponding sigma-points are easily obtained as $S^{(\varpi)}_{i} = \mu + 2 \sqrt{3} \sigma S_{i}$.

\begin{figure*}[t]
    \centering
    \captionsetup{justification=centering}
    \begin{minipage}[c]{\linewidth}
        \centering
        \includegraphics[keepaspectratio=true,width=\linewidth]{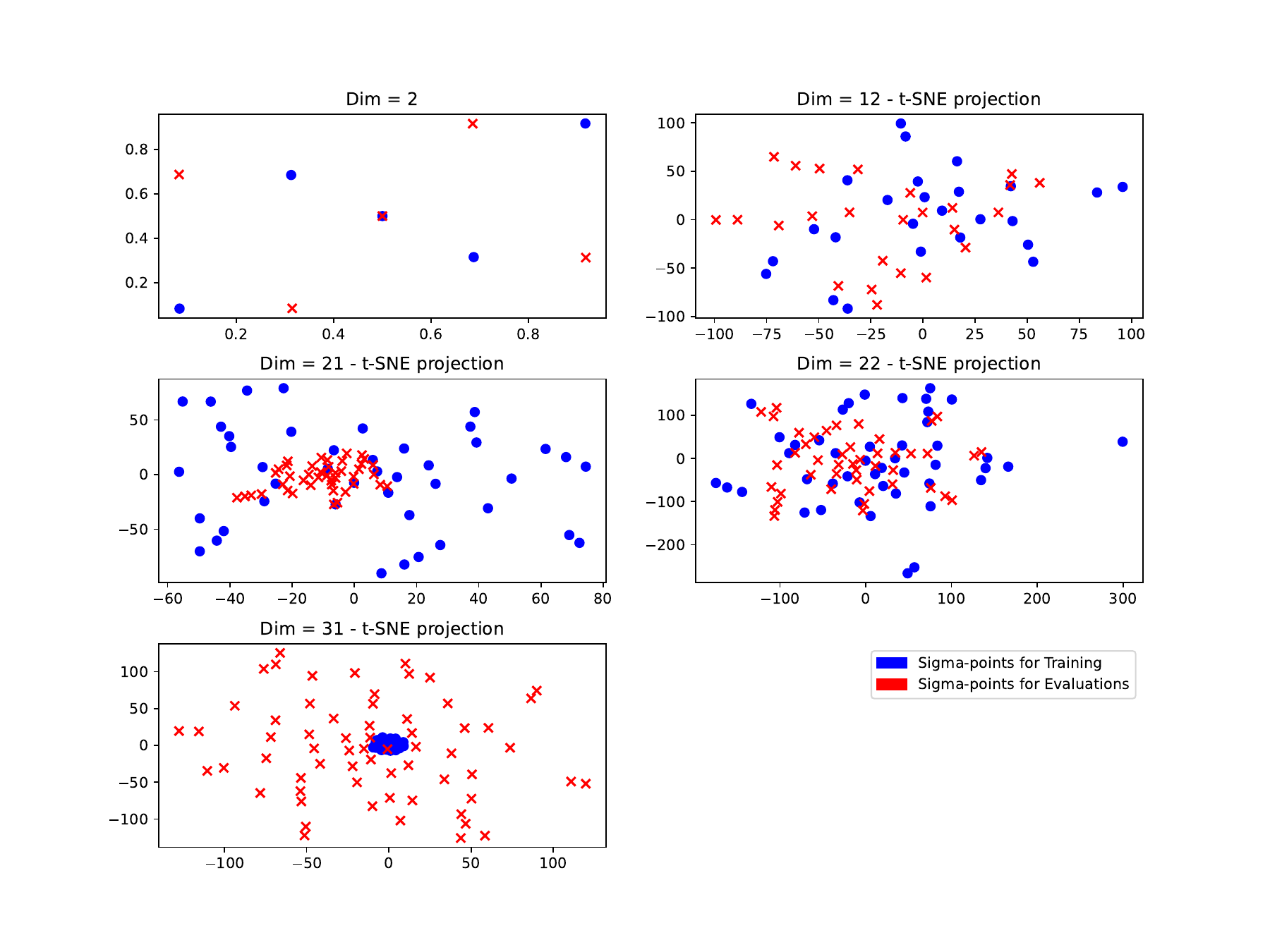}
    \end{minipage}
    \caption{Sigma-points used in the experiments. For the higher dimension ($d>2$), we use the \emph{sklearn} library's t-SNE to allow the visualization of both sets of sigma-points (for training and for evaluations) in a 2d plot.}
    \label{fig:ut_sigmapoints_viz}
\end{figure*}

In our experiments, for each dimension $d$ involved, we train two different sets of sigma-points with the first set used for training and the second used for the evaluations. Figure~\ref{fig:ut_sigmapoints_viz} gives the visualization of both sets of sigma-points for each dimensionality, while Table~\ref{tab:ut_2d_sigmapoints} gives the values for the 2-dimensional case (Note that the exact values used for the other dimensions can be found on the github link https://github.com/Mahoumaru/MDRL).
\begin{table*}[h]
		\captionsetup{width=\textwidth, justification=centering}
        \caption{Obtained sigma-points for the 2-dimensional case}
        \label{tab:ut_2d_sigmapoints}
        \centering
        \begin{tabular}{|l|cc|}
			\hline
			 & \textcolor{blue}{For training} & \textcolor{red}{For evaluations} \\
			\hline
			$S_{1}$ & [0.68726563, 0.3149943] &  [0.31483606, 0.08374587] \\
			$S_{2}$ & [0.31273094, 0.68499416] & [0.49997485, 0.49999967] \\
			$S_{3}$ & [0.0837525, 0.08276018] & [0.08281405, 0.6872253] \\
			$S_{4}$ & [0.91625065, 0.9172508] & [0.9171803, 0.31277505] \\
			$S_{5}$ & [0.5, 0.5] & [0.68519473, 0.91625404] \\
			\hline
		\end{tabular}
\end{table*}

\end{document}